\documentclass[lettersize,journal]{IEEEtran}
\usepackage{times}
\usepackage{cite}
\usepackage{amsmath,amssymb,amsfonts}
\usepackage{amsthm}
\usepackage{bm}
\usepackage{mathtools}
\usepackage{algorithm}
\usepackage{algpseudocode}
\usepackage[T1]{fontenc}
\usepackage{graphicx}
\usepackage{textcomp}
\usepackage[table,dvipsnames]{xcolor}
\usepackage{booktabs}
\usepackage{multirow}
\usepackage{array}
\usepackage{makecell}
\usepackage{colortbl}
\usepackage{threeparttable}
\usepackage{pifont}
\usepackage{balance}
\usepackage{tabularx}
\usepackage[caption=false,font=footnotesize]{subfig}
\usepackage{capt-of}
\usepackage{tikz}
\usepackage{pgfplots}
\pgfplotsset{compat=1.18}
\usetikzlibrary{arrows.meta,positioning,calc,fit,backgrounds,shapes.geometric,shapes.misc}
\usepgfplotslibrary{groupplots,fillbetween}
\usepackage{url}
\newcommand{\cmark}{\textcolor{cgreen}{\ding{51}}}

\usepackage[breaklinks=true,letterpaper=true,colorlinks,bookmarks=false]{hyperref}
\usepackage{microtype}

\definecolor{cgreen}{RGB}{31,140,90}
\definecolor{cblue}{RGB}{36,84,158}
\definecolor{corange}{RGB}{214,121,30}
\definecolor{cred}{RGB}{196,53,53}
\definecolor{cpurple}{RGB}{128,64,160}
\definecolor{cteal}{RGB}{0,128,128}
\definecolor{cgray}{RGB}{120,120,120}
\definecolor{rcsbg}{RGB}{255,247,232}
\definecolor{acpbg}{RGB}{233,243,252}
\definecolor{boxgray}{RGB}{244,244,244}

\newcommand{\ie}{\textit{i.e.}}

\newcommand{\cag}{\mathrm{GT\text{-}CAG}}
\newcommand{\cagb}{\mathrm{CAG}_{\mathrm{bnd}}}
\newcommand{\trust}{\tau_{\mathrm{trust}}}

\usepackage[capitalize]{cleveref}
\crefname{section}{Sec.}{Secs.}
\Crefname{section}{Section}{Sections}
\Crefname{table}{Table}{Tables}

\newcommand{\best}[1]{\textbf{#1}}

\newcommand{\up}{\,$\uparrow$}
\newcommand{\dn}{\,$\downarrow$}
\newcommand{\sdv}[1]{{\scriptsize$\pm$#1}}
\newcommand{\Fb}{$F_{\beta}$}
\newcommand{\Fm}{$F_{\beta}^{m}$}
\newcommand{\Fmax}{$F_{\beta}^{\max}$}
\newcommand{\Ep}{$E_{\phi}$}
\newcommand{\Sa}{$S_{\alpha}$}
\newcommand{\MAE}{$M$}

\ifCLASSOPTIONcompsoc
  \usepackage[nocompress]{cite}
\else
  \usepackage{cite}
\fi

\ifCLASSINFOpdf
\else
\fi

\begin{document}
\bstctlcite{IEEEexample:BSTcontrol}

\title{Supervision-Adaptive Curriculum and Guarded Anti-Curriculum Learning for Dense Prediction}

\author{Chunming~He, Dingming Zhang, Fengyang~Xiao, ~Zhiwen~Cao,~and~Sina~Farsiu,~\IEEEmembership{Fellow,~IEEE}
\IEEEcompsocitemizethanks{This work was supported in part by an unrestricted grant from Research to Prevent Blindness to Duke University. \IEEEcompsocthanksitem Corresponding authors: Fengyang Xiao and Sina Farsiu (e-mail: fengyang.xiao@duke.edu, sina.farsiu@duke.edu). 
\IEEEcompsocthanksitem Chunming He, Dingming Zhang, Fengyang Xiao, and Sina Farsiu are with the Department of Biomedical Engineering, Duke University, Durham, NC 27708 USA (e-mail: chunming.he@duke.edu).
\IEEEcompsocthanksitem Zhiwen Cao is with Adobe Inc., San Jose, CA, USA.
\IEEEcompsocthanksitem A preliminary version of this work appeared at ICML 2026. 
}}

\markboth{Submitted to IEEE Transactions on Pattern Analysis and Machine Intelligence}%
{He \MakeLowercase{\textit{et al.}}: Supervision-Adaptive Curriculum and Guarded Anti-Curriculum Learning for Dense Prediction}

\IEEEaftertitletext{%
  \vspace{-10mm}
  \begin{minipage}{\textwidth}
    \normalfont\normalsize
    \centering
    \includegraphics[width=\linewidth]{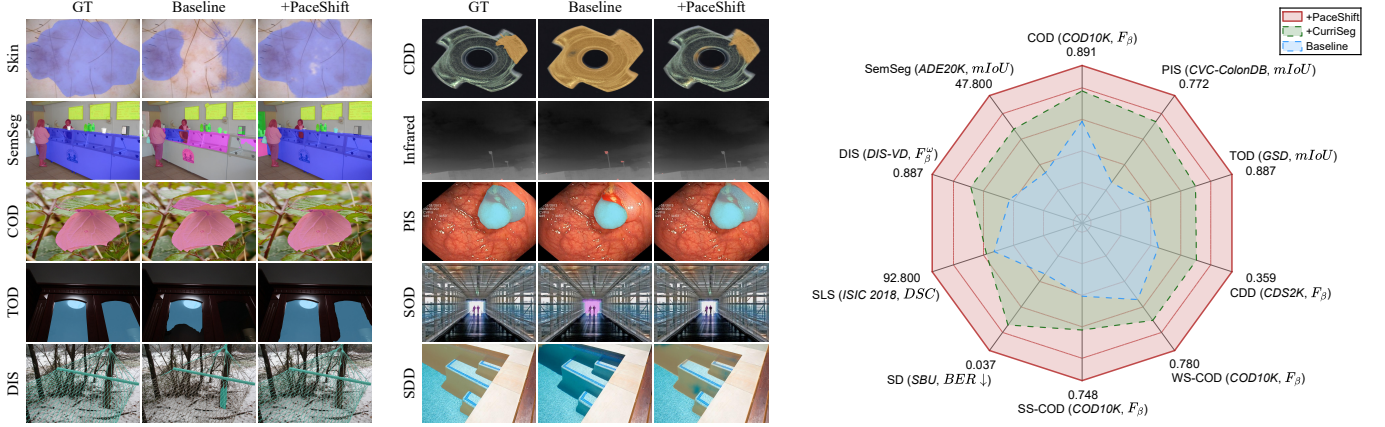}\vspace{-4mm}
\captionof{figure}{PaceShift across dense prediction tasks. \textbf{Left}: Ground-truth masks and predictions from the baseline and its PaceShift-trained counterpart, overlaid on the inputs. \textbf{Right}: Radar chart comparing the baseline, CurriSeg, and PaceShift, with each axis labeled ``Task (Dataset, Metric)''. Each axis is independently scaled so that the outer ring represents the best performance. For BER, lower values are mapped farther outward.
PaceShift adds no inference-time parameters.}
    \label{fig:teaser}
  \end{minipage}
  \vspace{2mm}
}

\maketitle

\begin{abstract}
Curriculum learning orders samples by difficulty, but difficulty alone does not establish supervision reliability for dense prediction. Easy samples can expose shortcuts, hard samples can be informative, and noisy or ambiguous samples can mislead training. Building on our conference method CurriSeg, we present PaceShift, a stabilize-then-challenge paradigm combining supervision-adaptive curriculum selection with guarded spectral post-training. A learning-dynamics view motivates separating supervision reliability from temporal update value and applying spectral challenges after stabilization. The first phase estimates reliability from available supervision, augmentation consistency, model disagreement, and prediction certainty. Temporal statistics guide sample weighting, while pixel-uncertainty weights reduce the early influence of uncertain regions. The second phase ranks predictions from multiple spectral views using supervision agreement, robustness, reference consistency, and penalties. A pairwise preference objective measured against the frozen Phase-I reference, together with an anchor, refines the prediction. The framework supports full, weak, semi-supervised, and model-derived supervision. Experiments across context-entangled content segmentation (CECS), broader dense-prediction tasks, and foundation-model and cross-modal adaptations show accuracy gains. Targeted diagnostics indicate improved pooled calibration and corruption robustness in the evaluated settings. PaceShift preserves the deployed architecture and adds no inference computation. Code will be publicly released.
\end{abstract}

\IEEEpeerreviewmaketitle

\setlength{\abovedisplayskip}{2pt}
\setlength{\belowdisplayskip}{2pt}
\section{Introduction}
\IEEEPARstart{D}{ense} prediction assigns a structured label to every pixel and underpins perception across medical, autonomous, and industrial applications~\cite{fan2020camouflaged,zhao2024spider,he2023strategic,xiao2024survey,he2026ride}. Most progress has been architectural: stronger backbones~\cite{wang2022pvt}, dedicated detectors with multi-scale aggregation~\cite{fan2021concealed,bui2024meganet,pang2024zoomnext}, attention~\cite{He2023Camouflaged}, and frequency-aware modules~\cite{sun2024frequency}.
We instead argue that dense prediction is limited not only by network capacity but also by \emph{training-signal reliability} and the order in which it is consumed (Fig.~\ref{fig:teaser}).

\begin{figure*}[t]
\setlength{\abovecaptionskip}{0cm}
\setlength{\belowcaptionskip}{-0.3cm}
\centering
\includegraphics[width=\linewidth]{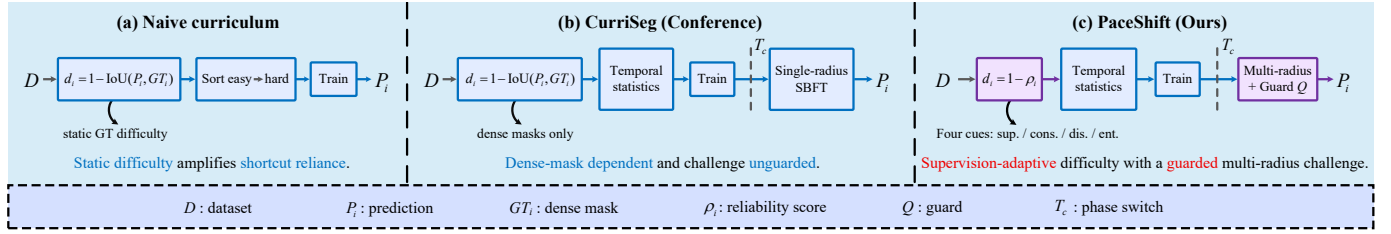}\vspace{-1mm}
\caption{Curriculum paradigms for dense prediction. A naive curriculum orders samples easy-to-hard by a static ground-truth difficulty and can strengthen reliance on fragile texture cues. CurriSeg adds temporal statistics and a stabilize-then-challenge schedule, but its score is defined by dense-mask error and its single-radius challenge is unguarded. PaceShift keeps the same overall two-phase training skeleton and changes two core components: a four-cue supervision-adaptive score $\rho_i$ and a guarded multi-radius spectral challenge during post-training, without changing the deployed model architecture.}
\label{fig:structure}\vspace{-6mm}
\end{figure*}

The reliability problem is sharpest in context-entangled content segmentation (CECS), where objects share intrinsic visual patterns with their background, as in camouflaged object detection~\cite{fan2021concealed}, transparent object detection~\cite{he2025run}, and concealed defect detection~\cite{xiao2024survey,fan2023advances}. 
We also include polyp segmentation as a CECS-related task with low boundary contrast, local ambiguity, and annotation uncertainty.
Existing methods address the resulting feature ambiguity mainly by enhancing representations~\cite{sun2024frequency,he2026nested}. Yet a more basic bottleneck is the learning dynamics that arise when the data distribution is entangled. We make this concrete with three observations that, while most visible in CECS, recur across dense prediction.

\looseness=-1
Difficulty and learning value are not the same quantity. Curriculum learning~\cite{bengio2009curriculum} typically orders samples from easy to hard, but in CECS, apparent easiness can arise from residual texture contrast that reveals the object. Prioritizing samples with low loss or error may therefore encourage reliance on fragile high-frequency cues instead of shape and context. This concern is supported by our experiments: under a matched budget, naive easy-to-hard ordering performs no better than random ordering, while conventional schedules can reduce accuracy. Frequency-intervention results further show greater sensitivity to high-frequency perturbations after naive easy-first training.

\looseness=-1
Hard samples pose a complementary challenge: high prediction error can reflect either informative visual difficulty or unreliable supervision. A well-concealed object may be difficult yet informative, whereas weak annotations, imperfect ground-truth quality~\cite{xiao2026beyond}, label drift, or intrinsically ambiguous regions can produce high errors without equally useful training signals. Since a static difficulty score cannot reliably distinguish these cases, difficulty-based selection may discard informative samples while retaining unreliable ones, increasing the risk of overfitting.

\looseness=-1
Confidence, finally, is not reliability. Uncertain boundary and low-contrast regions can dominate early gradients, while high confidence need not imply correctness. Directly rewarding confidence can therefore reinforce errors and degrade calibration.

Together, these three observations suggest that reliable training must distinguish supervision quality from temporal learning value, regulate uncertain pixels, and challenge shortcut reliance only after initial stabilization is reached.

\noindent\textbf{From CurriSeg to PaceShift.} Our conference work CurriSeg~\cite{he2026refining}, published at ICML 2026, introduced a two-phase stabilize-then-challenge framework for CECS. It combined dense-mask difficulty trajectories, temporal sample weighting, pixel uncertainty weighting, and single-radius spectral-blindness fine-tuning (SBFT). However, its difficulty score required dense ground-truth masks and could not be applied uniformly across weak, semi-supervised, and model-derived supervision. Moreover, its single-radius spectral stage lacked a mechanism for assessing the reliability of alternative predictions. To address these limitations, we develop PaceShift (Fig.~\ref{fig:structure}), which retains the two-phase design while introducing supervision-adaptive curriculum selection and guarded multi-radius post-training with reference-relative preference learning. These changes extend the training framework without modifying the deployed architecture or increasing inference cost.

\looseness=-1
\noindent\textbf{Scope.} CECS serves as our primary setting for studying supervision reliability and texture shortcuts together. Experiments on broader dense-prediction tasks evaluate the transfer of the full training framework, while mechanism-specific evidence is established primarily in CECS and unreliable-supervision settings.

This paper makes four contributions:

\noindent(1) \textbf{A learning-dynamics view of dense prediction (Sec.~\ref{sec:dynamics}).} We explain why naive easy-first curricula can amplify fragile shortcuts and why the stabilize-then-challenge framework succeeds, separating supervision reliability from temporal update value and treating SBFT as a post-stabilization frequency bottleneck whose order matters.

\noindent(2) \textbf{Supervision-adaptive curriculum selection (Sec.~\ref{sec:sarcs}).} We replace ground-truth difficulty with a reliability score combining available supervision, augmentation consistency, model disagreement, and prediction uncertainty, yielding one curriculum under full, weak, semi-supervised, and model-derived supervision while reducing confirmation bias.

\noindent(3) \textbf{Guarded anti-curriculum post-training (Sec.~\ref{sec:sbftp}).} SBFT-P (SBFT with preference-guarded post-training) combines multi-radius spectral challenges with a supervision-adaptive preference guard based on correctness, robustness, and consistency, with uncertainty as a weak penalty and not as the sole selection rule, improving robustness while reducing overconfidence in the evaluated settings and recovering the original SBFT as a special case under full supervision.

\noindent(4) \textbf{Systematic validation and diagnostics (Sec.~\ref{sec:exp}).} We evaluate PaceShift across multiple dense-prediction families, supervision regimes, architectures, data-centric baselines, and foundation-model and cross-modal settings. Targeted experiments examine training dynamics, supervision reliability, candidate selection, corruption robustness, and calibration without additional inference computation.

Beyond the methodological changes, this journal extension substantially expands the evaluation of CurriSeg~\cite{he2026refining}. We extend the broader dense-prediction evaluation to clinical, dichotomous, and salient segmentation; add vision-language and reference-guided evaluations; expand data-centric comparisons; and report paired-run statistical tests. Targeted analyses of supervision reliability, temporal update value, guard selection, phase ordering, frequency sensitivity, corruption robustness, and calibration further examine the underlying mechanisms.

\section{Related Work}
\noindent\textbf{Context-entangled and dense prediction segmentation.} CECS spans camouflaged object detection~\cite{he2023strategic,he2026unfoldcrf}, transparent object detection~\cite{mei2020don}, and concealed defect detection~\cite{fan2023advances}, with polyp segmentation~\cite{dong2023polyp} as a closely related low-contrast task. Progress has been mostly architectural: encoder-decoder backbones with multi-scale aggregation~\cite{fan2021concealed}, attention-based refinement~\cite{pang2024zoomnext}, edge- or uncertainty-aware branches~\cite{He2023Camouflaged,yang2021uncertainty}, transformer context modeling~\cite{yin2024camoformer}, and frequency-aware modules~\cite{sun2024frequency}.
These works strengthen representation but keep a standard supervised routine with shuffled mini-batches, so the roles of sample difficulty, label ambiguity, and texture shortcuts in the learning dynamics remain underexplored. Reversible unfolding~\cite{he2025run} and nested unfolding~\cite{he2026nested} designs have begun to inject inductive structure into the decoder, yet the optimization schedule itself is rarely treated as an explicit design variable.

\looseness=-1
\noindent\textbf{Curriculum learning and self-paced learning.}
Curriculum learning orders training from easy to hard to improve generalization~\cite{bengio2009curriculum,soviany2022curriculum}, with variants in self-paced learning~\cite{kumar2010self}, teacher-student curricula~\cite{matiisen2019teacher,saglietti2022analytical}, and loss- or hardness-based ranking~\cite{pentina2015curriculum,soviany2020curriculum}. These methods assume that easy samples give reliable supervision and that difficulty tracks learning utility. Both break in CECS: easy samples may carry spurious texture that promotes shortcuts, and hard samples can be at once informative and ambiguous, consistent with analyses showing that the benefit of ordering depends sharply on data structure~\cite{saglietti2022analytical}. 
Most curricula also ignore the pixel-level uncertainty of dense prediction, and almost none deliberately \emph{increase} difficulty after a stable solution is reached. 
PaceShift instead separates reliability from temporal update value, extends the curriculum to pixels, and introduces a post-stabilization anti-curriculum.

\noindent\textbf{Data-centric learning: training dynamics and data efficiency.}
Training-dynamics methods characterize examples through evolving predictions, losses, margins, or gradients, including forgetting events~\cite{toneva2018empirical,he2026hard}, EL2N and GraNd~\cite{paul2021deep}, dataset cartography~\cite{swayamdipta2020dataset}, and AUM~\cite{pleiss2020identifying}. Dyn-Unc~\cite{he2024large} and TDDS~\cite{zhang2024spanning} further exploit temporal variability, while InfoBatch~\cite{qin2024infobatch} and SeTa~\cite{zhou2025scale} dynamically remove low-value samples for efficiency. ADELE~\cite{liu2022adaptive} instead uses early-learning dynamics to correct noisy labels. Most of these formulations assume sample-level labels and do not directly address sparse, pixel-level, or model-derived supervision. PaceShift builds on this line of work but derives the trajectory from supervision-adaptive dense reliability, couples it with pixel-level uncertainty, and follows stabilization with guarded post-training. Its aim is accuracy, robustness, and calibration under spatially and supervision-dependent unreliability, not pruning or label correction.

\looseness=-1
\noindent\textbf{Learning under incomplete supervision.}
Weakly and semi-supervised segmentation learn from sparse annotations or unlabeled data through consistency regularization and pseudo-labeling~\cite{tarvainen2017mean,sohn2020fixmatch}.
A recurring failure is confirmation bias: training on the model's own confident predictions reinforces existing errors~\cite{arazo2020pseudo}. In CECS this is acute, since confidence and correctness diverge in concealed regions, so a confidence threshold both keeps wrong pseudo-labels and discards hard but correct ones. 
PaceShift instead combines confidence, disagreement, consistency, and an external foundation-model cue for reliability estimation, whereas SEE~\cite{he2025segment} generates changing pseudo-labels from a frozen SAM and SCALER~\cite{he2025scaler} jointly adapts SAM and the segmenter.

\noindent\textbf{Shortcut bias, texture bias, and frequency-domain learning.}
Networks trained by SGD favor the simplest features that separate the data even when more robust ones exist, studied as simplicity bias~\cite{shah2020pitfalls,pezeshki2021gradient} and, for vision, texture bias~\cite{hermann2020origins}. This differs from the spectral bias of network functions, which learn the low-frequency components of the input-output map first~\cite{rahaman2019spectral}. The two concern different frequencies, image versus function, and conflating them yields contradictory predictions about a low-pass operation. We therefore frame SBFT as suppressing a high-frequency \emph{image} shortcut to counteract texture bias without invoking function-space spectral bias.

\noindent\textbf{Foundation-model adaptation and post-training.}
Segment-Anything-style models~\cite{kirillov2023segment} are adapted to specialized domains through adapters, prompt modules, or low-rank updates~\cite{chen2023sam,hu2022lora}. Related segmentation systems condition on text or image prompts~\cite{luddecke2022image,zhang2024evf}, open-vocabulary alignment~\cite{cho2024cat}, or reference images~\cite{zhang2025referring}. Such conditioning introduces another source of unreliability, namely the alignment between the conditioning signal and the predicted mask, which PaceShift treats as an additional reliability cue without adding a new mechanism. Separately, DPO~\cite{rafailov2023direct} learns from preferred and dispreferred outputs without fitting a separate explicit reward model. We adopt a related pairwise objective for dense predictions and define preferences through correctness, robustness, and consistency on dense masks.

\begin{figure*}[t]
\setlength{\abovecaptionskip}{0cm}
\setlength{\belowcaptionskip}{-0.3cm}
\centering
\includegraphics[width=\linewidth]{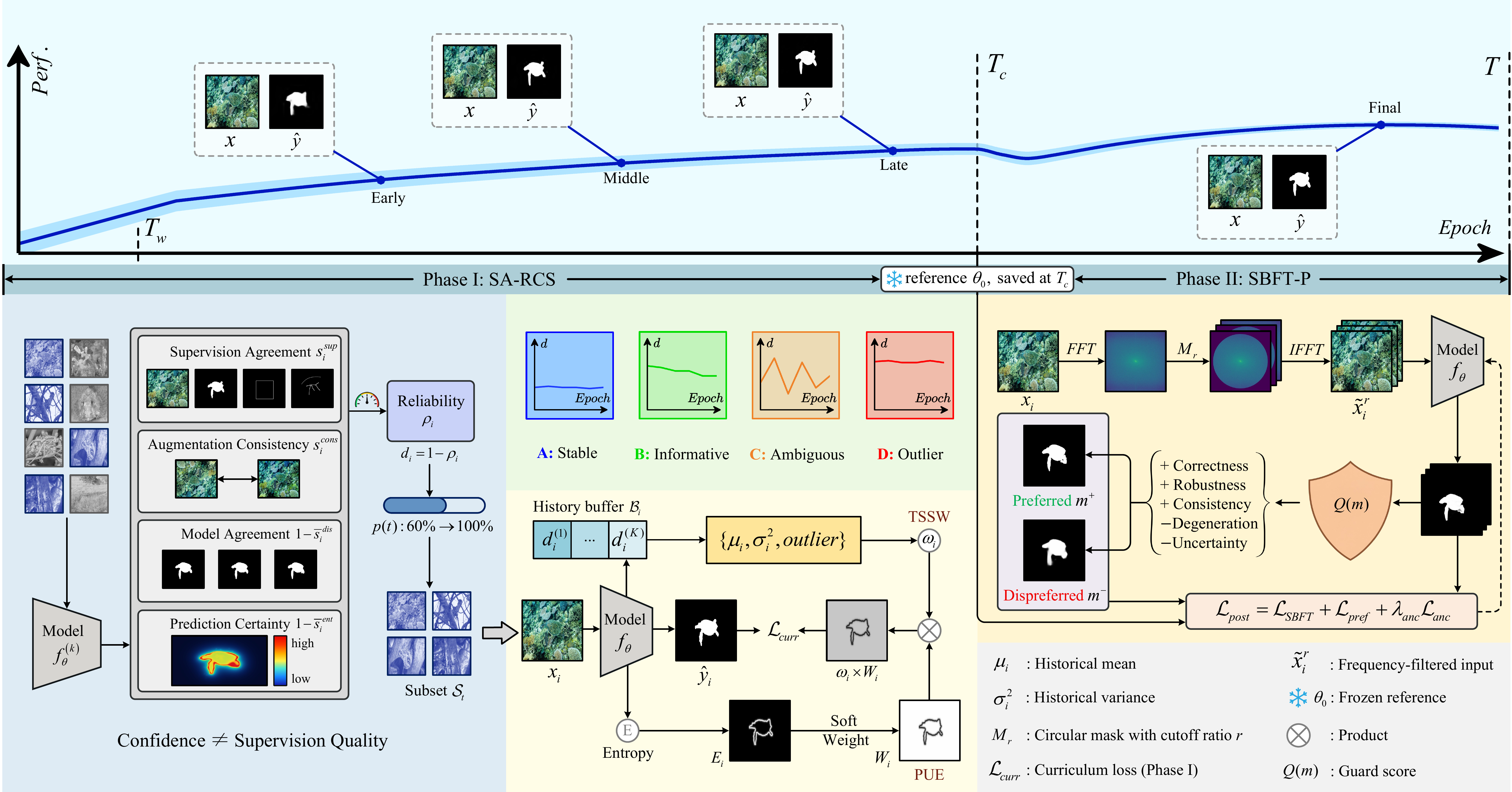}
\caption{Framework of PaceShift. \textbf{Top} (schematic): performance rises in Phase~I, is perturbed by the spectral challenge at $T_c$, and settles higher in Phase~II. \textbf{Phase~I (SA-RCS)}: four reliability cues form a supervision-adaptive difficulty $d_i=1-\rho_i$. After a warm-up on all samples, a percentile curriculum selects $\mathcal{S}_t$, and TSSW and PUE reweight the loss at sample and pixel level. \textbf{Phase~II (SBFT-P)}: a multi-radius challenge yields candidates that a guard scores into a preferred and dispreferred mask, refined against the reference $\theta_0$. No inference parameters are added.}
\label{fig:framework} \vspace{-6mm}
\end{figure*}

\section{Method}\label{sec:method}
\subsection{Problem Formulation and Overview}
Let $\mathcal{D}=\{(x_i,s_i)\}_{i=1}^{N}$ be a dense prediction dataset, where $x_i\in\mathbb{R}^{H_s\times W_s\times 3}$ is an image and $s_i$ is its supervision. In full supervision, $s_i$ is a dense mask $y_i\in\{0,1\}^{H_s\times W_s}$. In weak supervision, $s_i$ is a sparse cue such as a scribble, point, or box. In semi-supervision, $s_i$ is a dense mask for a small labeled subset $\mathcal{D}_L$ and empty for the unlabeled subset $\mathcal{D}_U$. A model $f_\theta$ predicts a per-pixel logit map $\hat{y}_i=f_\theta(x_i)$. The conference framework assumed $s_i=y_i$ for all $i$. PaceShift keeps the same network and the same stabilize-then-challenge schedule, but reorganizes training around the reliability of $s_i$ so that the schedule operates for any of the regimes above.

As shown in Fig.~\ref{fig:framework}, PaceShift runs in two phases over $T$ epochs. Phase~I, supervision-adaptive robust curriculum selection (SA-RCS, epochs $1$ to $T_c$), builds a stable representation by progressively selecting reliable yet informative samples and by damping uncertain pixels. Phase~II, guarded anti-curriculum promotion (SBFT-P, epochs $T_c+1$ to $T$), deliberately raises difficulty through a multi-radius spectral challenge and refines the model through a guarded preference objective. 

\looseness=-1
The two phases can be viewed as a time-varying objective: Phase~I minimizes $\sum_{i\in\mathcal{S}_t}\omega_i\,\mathcal{L}_{\mathrm{seg}}(f_\theta(x_i),t_i,W_i)$ over the growing reliable subset $\mathcal{S}_t$, where $\mathcal{L}_{\mathrm{seg}}$ is a weighted segmentation loss, $t_i$ is the supervision-adaptive target, and $\omega_i$ and $W_i$ are the sample and pixel weights, while Phase~II minimizes the post-training objective of \cref{eq:lpost}. The schedule hence moves from maximizing reliable information intake to minimizing reliance on nuisance frequencies, and both phases use the same supervision-evidence and source-trust construction, for reliable intake in Phase~I and guarded selection in Phase~II.

\subsection{A Learning-Dynamics View}\label{sec:dynamics}
We first lay out the learning dynamics motivating the design.

\subsubsection{Easy samples amplify a high-frequency shortcut} In a context-entangled scene, an object is concealed when its appearance matches the background. In CECS, one recurring source of apparent easiness is residual local high-frequency texture contrast that separates the object from its background. A model trained on these first can minimize the loss by keying on that texture difference, a feature that is cheap to fit and locally discriminative but does not transfer to well-concealed cases. This is an instance of simplicity and texture bias~\cite{shah2020pitfalls}: gradient descent prefers the simplest separating feature even when a more robust shape or context feature exists. This is \emph{not} the classical spectral bias of network functions~\cite{rahaman2019spectral,zhi2020frequency}, which concerns the order in which frequencies of the input-output map are learned. Here the relevant frequency is in image space, and the pathology is that early easy-first training increases reliance on a high-frequency image cue. This motivates the two phases: keep shortcut-prone easy samples from dominating early training, then suppress the high-frequency channel once a sufficiently stable representation exists.

\subsubsection{Temporal dynamics reveal update value}
A single difficulty value cannot tell an informative hard sample from a noisy one, while the training trajectory can. For sample $i$ let $d_i^{(k)}$ denote its difficulty at checkpoint $k$. Two patterns are used as heuristics. A high variance of $\{d_i^{(k)}\}_k$ reflects strong temporal variation, which is common near decision boundaries or under intrinsic ambiguity. A high mean with low variance reflects persistent failure, which can arise from outliers or label errors that the model cannot resolve, and the empirical rule down-weights it. 
This reading aligns with example-difficulty and forgetting analyses~\cite{toneva2018empirical,paul2021deep} and with Dyn-Unc and TDDS~\cite{he2024large,zhang2024spanning}, which likewise exploit training dynamics instead of a single value. We import neither score directly: we model the mean, the variance, and the persistent-failure pattern of high mean and low variance as separate, individually ablatable terms on the supervision-adaptive difficulty of Sec.~\ref{sec:sarcs}, because the noise-ambiguity dilemma of CECS requires distinguishing all three under any supervision regime.
This is what motivates the parameter-free mid-variance preference and the outlier penalty $\omega^{\mathrm{out}}$ in Sec.~\ref{sec:sarcs}: intermediate temporal variation serves as a heuristic for potentially informative hardness, whereas persistent stability and persistent failure are down-weighted.

\subsubsection{SBFT is a post-stabilization frequency bottleneck}
Split the input into a low-frequency component $x^{\mathrm{lo}}$ and a high-frequency component $x^{\mathrm{hi}}$. A representation useful for dense prediction should retain the information needed for the target without relying on the nuisance channel $x^{\mathrm{hi}}$. SBFT attenuates $x^{\mathrm{hi}}$ while requiring consistent prediction through the supervised or guarded objective, a frequency-level bottleneck inspired by, but not formally an instance of, the information bottleneck~\cite{tishby2015deep}. The phase order is not interchangeable: if the bottleneck is imposed before a stable representation exists, the model is deprived of the cues it needs to bootstrap structure, and applying the challenge directly after warm-up falls below standard training on every reported metric (Table~\ref{tab:phaseorder}).

\subsection{Supervision-Adaptive Robust Curriculum Selection}\label{sec:sarcs}
SA-RCS targets the noise-ambiguity dilemma by monitoring the temporal behavior of samples and the spatial uncertainty of pixels, and it does so without assuming a complete mask. It has four parts: a warm-up curriculum, a supervision-adaptive difficulty, a temporal sample weighting, and a pixel uncertainty weighting, combined in a single loss.

\subsubsection{Supervision-adaptive difficulty}
The difficulty defined in the conference version is the per-sample segmentation error of a periodically saved checkpoint $f_\theta^{(k)}$,
\begin{equation}
d_i^{\mathrm{full}} = 1 - \mathrm{IoU}\!\big(f_\theta^{(k)}(x_i),\, y_i\big),
\label{eq:dfull}
\end{equation}
which requires the dense mask $y_i$. For binary segmentation, IoU-based difficulty, reliability, and guard scores use the thresholded prediction $\mathbf{1}[\sigma(f_\theta(x))\ge 0.5]$. We replace it with a difficulty $d_i = 1-\rho_i$ derived from a reliability score $\rho_i\in[0,1]$ that aggregates the supervision signals that are actually available for the current training sample. We use four normalized cues, each in $[0,1]$ and oriented so that larger means more reliable:
\begin{itemize}
\item supervision agreement $s_i^{\mathrm{sup}}$: for full supervision the IoU between $f_\theta^{(k)}(x_i)$ and $y_i$, for weak supervision the satisfied ratio of the available sparse cue (scribble overlap, point coverage, or box containment), with the point-supervised setting averaging point-cue satisfaction and agreement with a frozen SAM3 proposal, for unlabeled samples the agreement with an exponential-moving-average teacher or a foundation-model mask proposal.
\item augmentation consistency $s_i^{\mathrm{cons}}$: one minus the mean disagreement between predictions on two stochastic augmentations of $x_i$, where each prediction is inverse-warped to the original image coordinates and the disagreement is measured by $1-\mathrm{IoU}$ over the aligned masks.
\item model agreement $1-\bar{s}_i^{\mathrm{dis}}$: one minus the disagreement across Monte-Carlo dropout predictions, or among ensemble predictions when an ensemble is used, an internal model-uncertainty cue, distinct from the external proposal that provides mask-level evidence for $s_i^{\mathrm{sup}}$ where used.
\item prediction certainty $1-\bar{s}_i^{\mathrm{ent}}$: one minus the image-level average of normalized pixel entropy over the probability map, an image-level cue for $\rho_i$ that is distinct from the pixel-level entropy weight $W_{h,w}$ in \cref{eq:pixelw}.
\end{itemize}
The reliability is their convex combination,
\begin{equation}
\begin{aligned}
\rho_i = {}& a^{\mathrm{sup}}\, s_i^{\mathrm{sup}} + a^{\mathrm{cons}}\, s_i^{\mathrm{cons}} + a^{\mathrm{dis}}\,\big(1-\bar{s}_i^{\mathrm{dis}}\big) \\
           &  + a^{\mathrm{ent}}\,\big(1-\bar{s}_i^{\mathrm{ent}}\big), 
\end{aligned}
\label{eq:rho}
\end{equation}
with $a^{\mathrm{sup}}+a^{\mathrm{cons}}+a^{\mathrm{dis}}+a^{\mathrm{ent}}=1$, and the difficulty used everywhere below is $d_i = 1-\rho_i$. Instead of setting the four weights per regime, we derive all of them from a single source-specific \emph{supervision-trust} level $\trust\in[0,1]$ that reflects how verifiable the supervision source is: $a^{\mathrm{sup}}=\trust$, and the residual mass is shared by the auxiliary cues through a fixed global template, $(a^{\mathrm{cons}},a^{\mathrm{dis}},a^{\mathrm{ent}})=(1-\trust)\,\pi$ with $\pi=(0.4,0.4,0.2)$, so that consistency and disagreement remain the main auxiliary signals and entropy a weak one, consistent with the design point below. $\trust$ takes one of three levels according to the primary supervision source (Table~\ref{tab:guardweights}): $\trust=0.7$ for a dense human mask, which concentrates mass on $s^{\mathrm{sup}}$ and recovers \cref{eq:dfull} in the limit $\trust\!\to\!1$, $\trust=0.4$ for a sparse human cue, and $\trust=0.3$ for a model-derived signal, where the agreement term is itself uncertain.

\begin{table}[t]
\setlength{\abovecaptionskip}{0cm} 
\setlength{\belowcaptionskip}{-0.2cm}
\centering
\caption{Supervision-trust presets. In \cref{eq:rho}, $a^{\mathrm{sup}}{=}\trust$ and $(a^{\mathrm{cons}},a^{\mathrm{dis}},a^{\mathrm{ent}})=(1-\trust)\,\pi$ with the fixed template $\pi=(0.4,0.4,0.2)$, and the guard in \cref{eq:guard} reuses $\trust$ with the fixed penalty scale $\varepsilon$. Point supervision averages cue satisfaction with frozen-proposal agreement without changing the preset.}
\label{tab:guardweights}
\setlength{\tabcolsep}{3pt}
\renewcommand{\arraystretch}{1.25}
\footnotesize
\begin{tabularx}{\columnwidth}{
>{\raggedright\arraybackslash\hsize=1.5\hsize}X
>{\centering\arraybackslash\hsize=0.7\hsize}X
>{\raggedright\arraybackslash\hsize=0.8\hsize}X}
\toprule
Supervision sources & \mbox{Trust $\trust$} & $Q_{\mathrm{sup}}$ source\\
\midrule
Dense human mask (full; semi, labeled part) & 0.7 & GT mask\\
Sparse human cue (scribble, point, box) & 0.4 & weak-cue evidence\\
Model-derived supervision (EMA teacher, CAM-derived pseudo-mask) & 0.3 & teacher / pseudo-mask agree\\
\bottomrule
\end{tabularx}\vspace{-5mm}
\end{table}
Reliability is not confidence. Here $\rho_i$ measures the evidence-conditioned reliability of the current training signal, not annotation correctness alone: $\trust$ acts as a source-trust prior, while the sample-dependent cues measure how strongly the available evidence supports learning from sample $i$ at the current stage. The entropy term alone would select the samples the model is already sure about and would close the self-training feedback loop. Combining it with disagreement and consistency, and anchoring $s^{\mathrm{sup}}$ on an external foundation-model proposal that is independent of the current model parameters where such a proposal is used, selects samples that are stable and still informative. This construction helps reduce confirmation bias in self-training.
The trust level $\trust$ is the only regime-level quantity and the template $\pi$ is global, so a new regime requires choosing one trust level, with no search over weights. The presets perform close to validation-tuned values and clearly better than uniform cue weights (Table~\ref{tab:trust}), so the gains are not the product of per-dataset search.

\subsubsection{Warm-up curriculum}
Following the conference~\cite{he2026refining}, we expand the active subset along a percentile schedule after a short warm-up. For the first $T_w$ epochs all samples are used without filtering, so the model acquires a basic representation before difficulty-based selection begins. Let $p(t)$ be the selection percentile at epoch $t$ and $\mathrm{Per}(\{d_n\}_{n=1}^{N},p(t))$ the corresponding threshold. The active subset is
\begin{equation}
\begin{aligned}
\mathcal{S}_t &= \mathcal{D}, \qquad t\le T_w,\\[2pt]
\mathcal{S}_t &= \big\{\, i \mid d_i \le \mathrm{Per}(\{d_n\}_{n=1}^{N},\, p(t)) \,\big\}, \quad T_w<t\le T_c,\\[2pt]
p(t)&=p_{\min}+(1-p_{\min})\,\tfrac{t-T_w-1}{T_c-T_w-1},
\end{aligned}
\label{eq:select}
\end{equation}
with $p_{\min}=0.6$ and $T_w=10$. Training starts from high-reliability samples, selected by the reliability score instead of difficulty alone, and progressively admits lower-reliability, harder samples as the percentile $p(t)$ grows.
 
\subsubsection{Temporal statistics-based sample weighting}
Within $\mathcal{S}_t$, samples still differ in learning value. To capture this, we compute the difficulty at refresh checkpoints. Let $f_\theta^{(k)}$ denote checkpoints saved at epochs $kH<T_c$, $k=1,\ldots,K$, where $K=\left\lfloor\frac{T_c-1}{H}\right\rfloor$ ($K{=}5$ at the defaults), and let $d_i^{(k)}=1-\rho_i^{(k)}$, with $\rho_i^{(k)}$ the reliability of \cref{eq:rho} evaluated under $f_\theta^{(k)}$. The history $\mathcal{B}_i=\{d_i^{(1)},\dots,d_i^{(k)}\}$ stores these values, and
\begin{equation}
\mu_i=\frac{1}{|\mathcal{B}_i|}\sum_{k}d_i^{(k)},\qquad
\sigma_i^{2}=\frac{1}{|\mathcal{B}_i|}\sum_{k}\big(d_i^{(k)}-\mu_i\big)^2,
\label{eq:tstat}
\end{equation}
so the statistics aggregate distinct training states, not repeated copies of one checkpoint. Until two checkpoints exist we set $\omega_i=1$. 
As argued in Sec.~\ref{sec:dynamics}, moderate variance is used as a heuristic for informative hardness, very high variance is common for ambiguous or unstable samples, and a high mean with low variance is treated as persistent failure. After min-max normalizing to $\tilde{\mu}_i$ and $\tilde{\sigma}_i^2$, we map the statistics to weights,
\begin{equation}
\omega_i^{\mu}=1-\tilde{\mu}_i,\qquad
\omega_i^{\sigma}=4\,\tilde{\sigma}_i^{2}\big(1-\tilde{\sigma}_i^{2}\big),
\label{eq:wmuwsig}
\end{equation}
a parameter-free mid-variance preference that peaks at the midpoint of the normalized range and suppresses both already-stable and highly unstable samples. This parameter-free parabolic variance weight replaces the Gaussian variance weight of CurriSeg~\cite{he2026refining}. The final sample weight:
\begin{equation}
\omega_i \!=\! w_{\min}+(1-w_{\min})\,\omega_i^{\mu}\,\omega_i^{\sigma}\,\omega_i^{\mathrm{out}}, \ 
\omega_i^{\mathrm{out}} \!=\! 1-\tilde{\mu}_i\,(1-\tilde{\sigma}_i^{2}),
\label{eq:omega}
\end{equation}
with $w_{\min}=0.1$ a floor that maintains gradient flow and is shared with the pixel weighting in~\cref{eq:pixelw}. 
We use intermediate temporal variation as a heuristic for potentially informative hardness, while the high-mean, low-variance factor suppresses persistent failures. Section~\ref{sec:exp:mech} tests this heuristic directly against update alignment. Under incomplete supervision the same statistics operate on $d_i=1-\rho_i$ without modification.

\subsubsection{Pixel uncertainty weighting (PUE)}
Boundary and low-contrast pixels are harder than homogeneous ones and their gradients can dominate early optimization. Unlike pseudo-label filtering~\cite{he2025segment}, we down-weight uncertain pixels on the predicted mask itself without hiding them. For a logit $\hat{y}_{h,w}$ with probability $p_{h,w}=\sigma(\hat{y}_{h,w})$, the normalized entropy is
\begin{equation}
E_{h,w}=-p_{h,w}\log_2 p_{h,w}-(1-p_{h,w})\log_2(1-p_{h,w}),
\label{eq:entropy}
\end{equation}
which peaks at $p_{h,w}=0.5$ and vanishes at certainty. The soft pixel weight at epoch $t$ is formulated as:
\begin{equation}
W_{h,w}(t)=w_{\min}+(1-w_{\min})\,\big(1-\beta(t)\,E_{h,w}\big),
\label{eq:pixelw}
\end{equation}
with $w_{\min}$ the shared non-zero floor of~\cref{eq:omega} and $\beta(t)=\max(0,\,1-t/T_c)$ a curriculum-aware coefficient. The weight $W$ is detached from the graph, so \cref{eq:pixelw} reweights the loss without back-propagating through the entropy-derived weights. Early on, high-entropy pixels are strongly down-weighted, and as $\beta(t)$ decays to zero the pixel weights return to one and remain uniform throughout Phase~II, while the floor keeps gradient flow from all pixels.

\subsubsection{Unified curriculum loss}
For a sample $i\in\mathcal{S}_t$ the loss combines the two granularities,
\begin{equation}
\mathcal{L}_{\mathrm{curr}}=\omega_i\,\mathcal{L}^{w}_{\mathrm{seg}}(\hat{y}_i,t_i,W_i),
\label{eq:lcurr}
\end{equation}
where $\mathcal{L}^{w}_{\mathrm{seg}}$ is the regime-appropriate weighted segmentation loss, a weighted BCE plus IoU under full supervision, the corresponding scribble, point, or box loss under weak supervision, and an entropy-gated pseudo-label loss for unlabeled samples, applied to the matching supervision-adaptive target $t_i$. Pixels without any supervision contribute zero to $\mathcal{L}^{w}_{\mathrm{seg}}$. Their predictions still enter the consistency and entropy cues of \cref{eq:rho}, so the curriculum remains well defined across regimes.

The same formulation applies across supervision sources within the same framework by re-instantiating $s_i^{\mathrm{sup}}$, $t_i$, and $Q_{\mathrm{sup}}$: dense supervision uses ground-truth masks, sparse supervision uses cue evidence with the corresponding weak loss, and unlabeled samples use available EMA-teacher and frozen-proposal evidence, and when both are present, their agreement is averaged and disagreement pixels are down-weighted. Trust follows the primary supervision source, with the supervision-specific definitions given in the Supp.

\subsubsection{Beyond binary masks}
Multi-class and instance segmentation need no new component: the binary entropy becomes a normalized multi-class entropy, agreement is measured by class-averaged IoU or by Hungarian-matched mask IoU with an object-count term, and the degeneration penalty penalizes collapse to background, onto a single class, or in object count. The trust and loss coefficients $\trust$, $\varepsilon$, and $\lambda_{\mathrm{anc}}$ are shared, with fixed output-specific collapse checks detailed in the Supp., and \cref{eq:tstat,eq:omega,eq:select} transfer unchanged: they operate on the scalar $d_i=1-\rho_i$ and never see the output format.

\subsubsection{Cross-modal conditioning}
Conditioning on a language prompt or a reference image adds one axis of unreliability: the mask may disagree with the conditioning even when the model is internally confident. We instantiate the existing cues rather
than introducing new ones, leaving $\rho_i$ and $Q(m)$ unchanged and $s_i^{\mathrm{sup}}$ mask-level. Language conditioning uses prompt-paraphrase consistency, the agreement across semantically equivalent rephrasings, and vision-text disagreement, the gap to a vision-only prediction. Reference conditioning analogously uses reference-swap consistency and vision-reference disagreement. Conditioned predictions provide reliability evidence, not fixed training labels, while the trust level continues to follow the primary supervision source (Table~\ref{tab:guardweights}).

\subsection{Guarded Anti-Curriculum Promotion}\label{sec:sbftp}
Once SA-RCS stabilizes at epoch $T_c$, we raise difficulty on purpose. The conference SBFT applied a single low-pass filter and fine-tuned against the mask. We extend it into a guarded post-training stage, SBFT-P, that creates a spectral challenge of several strengths, scores the resulting candidate predictions with a supervision-adaptive guard, and refines the model with a guarded preference objective. The guard is what keeps the stage from reducing to confidence maximization.

\subsubsection{Multi-radius spectral challenge}
We low-pass the input in the frequency domain with a circular mask,
\begin{equation}\hspace{-3mm}
\begin{aligned}
\tilde{x}_i^{1.00}&=x_i,\quad
\tilde{x}_i^{r}=\mathcal{F}^{-1}\!\big(\mathcal{F}(x_i)\odot M_r\big)
\ \ \text{for }r<1,\\[2pt]
M_r(u,v)&=\mathbf{1}\!\Big[(u-c_u)^2\!+\!(v-c_v)^2 \!\le\!
\big(\tfrac{r\,\min(H_s,W_s)}{2}\big)^2\Big],
\end{aligned}
\label{eq:filter}
\end{equation}
where $\mathcal{F}$ and $\mathcal{F}^{-1}$ denote the centered Fourier transform and its inverse, $(c_u,c_v)$ is the frequency origin, and $\mathbf{1}[\cdot]$ is the indicator. We use a small set of radii $r\in\{0.90,0.95,1.00\}$. The identity view $r=1.00$ is the clean prediction by definition, while smaller radii progressively remove the high-frequency texture channel. Passing each view through the segmenter gives detached candidate masks $m_i^{r}=\mathrm{sg}\big[\sigma\big(f_\theta(\tilde{x}_i^{r})\big)\big]$, where $\mathrm{sg}[\cdot]$ denotes stop-gradient. The thresholded masks $\mathbf{1}[m_i^{r}\!\ge\!0.5]$ enter the overlap and area checks of the guard, and the soft probabilities enter the entropy term and the preference objective.
The three radii expose how predictions change as high-frequency content is progressively removed, which the guard turns into a robustness signal.

\subsubsection{Supervision-adaptive guard}
We score each candidate mask $m$ with the following evaluation function:
\begin{equation}
\begin{aligned}
Q(m)={}&\trust\,Q_{\mathrm{sup}}(m)\\
&+\tfrac{1-\trust}{2}
\big[Q_{\mathrm{rob}}(m)+Q_{\mathrm{cons}}(m)\big]\\
&-\varepsilon
\big[Q_{\mathrm{deg}}(m)+Q_{\mathrm{unc}}(m)\big],
\end{aligned}
\label{eq:guard}
\end{equation}
{\looseness=-1
whose terms are: a correctness term $Q_{\mathrm{sup}}$, which uses the same supervision source as $s_i^{\mathrm{sup}}$ in \cref{eq:rho} (GT overlap, weak-cue satisfaction, or teacher/proposal agreement) but is evaluated on each candidate mask, a robustness term $Q_{\mathrm{rob}}$, the mean agreement of $m$ with the other candidates across radii, 
a consistency term $Q_{\mathrm{cons}}$, the agreement of $m$ with the clean-view prediction of a frozen reference, by default the Phase-I checkpoint $\theta_0$, which anchors consistency to the Phase-I model instead of an evolving clean prediction,
a degeneration penalty $Q_{\mathrm{deg}}$, which discourages all-background masks, area collapse, and over-smoothing, and an uncertainty term $Q_{\mathrm{unc}}$, the mean entropy. Positive weights sum to one and shift from correctness toward robustness and consistency as supervision becomes less verifiable, while $\varepsilon=0.1$ is a fixed penalty scale shared across regimes. Detached candidates prevent gradients through self-generated targets. The preferred and dispreferred masks are}
\begin{equation}\hspace{-3mm}
r_i^{+}\!=\!\arg\max_{r} Q(m_i^{r}), r_i^{-}\!=\!\arg\min_{r} Q(m_i^{r}), m_i^{\pm}\!=\!m_i^{r_i^{\pm}}.
\label{eq:prefsel}
\end{equation}
Because $Q$ is dominated by accuracy, stability under spectral perturbation, and non-degeneracy, with confidence entering only as a weak penalty, selection favors predictions supported by multiple cues, not confidence alone, and \cref{tab:calib} reports that this does not inflate confidence.

\subsubsection{Guarded preference objective}
We adopt a pairwise preference objective in the spirit of direct preference optimization~\cite{rafailov2023direct}, adapted to dense masks. Let
\begin{equation}
\begin{split}
\ell_\theta(m\mid x)
= &\tfrac{1}{|\Omega|}\!\sum_{h,w}\!\big[
m_{h,w}\log p_{h,w} \\
&+ (1-m_{h,w})\log(1-p_{h,w})
\big]
\end{split}
\label{eq:loglik}
\end{equation}
be the pixel-averaged log-likelihood of mask $m$ under the model, with $p=\sigma(f_\theta(x))$, and let $\theta_0$ be the reference saved at the end of SA-RCS. Writing $\Delta\ell_\theta(m)=\ell_\theta(m)-\ell_{\theta_0}(m)$ for the log-likelihood margin of the model against the reference, the preference loss prefers $m^{+}$ over $m^{-}$,
\begin{equation}
\mathcal{L}_{\mathrm{pref}}=-\log\sigma\!\big(\Delta\ell_\theta(m^{+})-\Delta\ell_\theta(m^{-})\big),
\label{eq:pref}
\end{equation}
evaluated on the clean input $x_i$, not on a challenge view, so post-training transfers the preferred mask back onto the deployed inference path.
Since $\ell_\theta$ is affine in the target mask, the preference margin has the equivalent logit form
\begin{equation}
\Delta\ell_\theta(m^{+})-\Delta\ell_\theta(m^{-})
=\frac{1}{|\Omega|}\sum_{j\in\Omega}
(m^{+}_{j}-m^{-}_{j})\Delta z_j,
\label{eq:preflogit}
\end{equation}
where $\Delta z_j=z_{\theta,j}-z_{\theta_0,j}$ denotes the clean-view logit difference, with $z_\theta=f_\theta(x_i)$ and $\Omega$ the pixel set. The objective favors higher relative logits where $m_j^{+}>m_j^{-}$ and lower ones where $m_j^{+}<m_j^{-}$. A clean-view anchor keeps the model close to the frozen reference and limits preference-driven drift,
\begin{equation}
\mathcal{L}_{\mathrm{anc}}=\mathrm{KL}\!\big(p_{\theta_0}(x_i)\,\|\,p_\theta(x_i)\big),
\label{eq:anchor}
\end{equation}
written with the frozen reference $\theta_0$ as the teacher distribution, the standard reference-distillation direction. The conference SBFT term is retained,
\begin{equation}
\mathcal{L}_{\mathrm{SBFT}}\!=\!\mathcal{L}^{w}_{\mathrm{BCE}}(\tilde{p}_i,t_i)\!+\!\mathcal{L}^{w}_{\mathrm{IoU}}(\tilde{p}_i,t_i),\ \tilde{p}_i\!=\!\sigma\!\big(f_\theta(\tilde{x}_i^{0.95})\big),
\label{eq:sbft}
\end{equation}
where $t_i$ is the supervision-adaptive target (the dense mask, or the guarded preferred mask $m_i^{+}$ when no dense mask exists). The superscript $w$ follows the pixel-weighted loss convention of Phase~I, and because $\beta=0$ in Phase~II, $W_{h,w}=1$ for all pixels. The post-training loss is formulated as follows: 
\begin{equation}
\mathcal{L}_{\mathrm{post}}=\mathcal{L}_{\mathrm{SBFT}}+\mathcal{L}_{\mathrm{pref}}+\lambda_{\mathrm{anc}}\,\mathcal{L}_{\mathrm{anc}},
\label{eq:lpost}
\end{equation}
\looseness=-1
where the preference and temperature scales are fixed to one and $\lambda_{\mathrm{anc}}\!=\!0.1$ is a global anchor strength reused across all datasets and regimes. Under full supervision, dropping the preference and anchor terms, \ie, keeping only $\mathcal{L}_{\mathrm{SBFT}}$ with the single radius $r=0.95$, recovers the conference SBFT objective. Without dense masks the target $t_i$ is the guarded $m_i^{+}$, which the conference formulation did not define. The ablations in \cref{tab:postabl} isolate the contribution of each added term.

\begin{table*}[t]
\setlength{\abovecaptionskip}{0cm} 
\centering
\caption{Camouflaged object detection. ``+'' and ``++'' denote training the baseline under CurriSeg~\cite{he2026refining} and PaceShift. $\Delta$: mean relative improvement over the baseline across all metrics and datasets, with the sign reversed for lower-is-better metrics so that positive values indicate improvement. \best{Bold}: best within each backbone block. $^{\dagger}$: our DINO~V2-L retraining of VSCode under the protocol of this table. $^{\ddagger}$: evaluated by us from the authors' released model under the protocol of this table.}
\label{tab:cod}
\resizebox{\textwidth}{!}{
\setlength{\tabcolsep}{1.4mm}
\begin{tabular}{@{}ll *{16}{c} c}
\toprule
\multirow{2}{*}{Methods} & \multirow{2}{*}{Backbone}
 & \multicolumn{4}{c}{CHAMELEON} & \multicolumn{4}{c}{CAMO} & \multicolumn{4}{c}{COD10K} & \multicolumn{4}{c}{NC4K} & \multirow{2}{*}{$\Delta$(\%)}\\
\cmidrule(lr){3-6}\cmidrule(lr){7-10}\cmidrule(lr){11-14}\cmidrule(lr){15-18}
 & & $M$\dn & \Fb\up & \Ep\up & \Sa\up & $M$\dn & \Fb\up & \Ep\up & \Sa\up & $M$\dn & \Fb\up & \Ep\up & \Sa\up & $M$\dn & \Fb\up & \Ep\up & \Sa\up & \\
\midrule
MGL-R~\cite{zhai2021mutual} & ResNet50 & 0.031 & 0.825 & 0.917 & 0.891 & 0.088 & 0.738 & 0.812 & 0.775 & 0.035 & 0.680 & 0.851 & 0.814 & 0.053 & 0.778 & 0.867 & 0.833 & ---\\
PreyNet~\cite{zhang2022preynet} & ResNet50 & 0.027 & 0.844 & 0.948 & 0.895 & 0.077 & 0.763 & 0.854 & 0.790 & 0.034 & 0.715 & 0.894 & 0.813 & 0.047 & 0.798 & 0.887 & 0.838 & ---\\
FEDER~\cite{He2023Camouflaged} & ResNet50 & 0.028 & 0.850 & 0.944 & 0.892 & 0.070 & 0.775 & 0.870 & 0.802 & 0.032 & 0.715 & 0.892 & 0.810 & 0.046 & 0.808 & 0.900 & 0.842 & ---\\
FEDER+~\cite{he2026refining} & ResNet50 & 0.026 & 0.858 & 0.952 & 0.898 & 0.068 & 0.790 & 0.881 & 0.807 & 0.030 & 0.736 & 0.910 & 0.818 & 0.043 & 0.825 & 0.912 & 0.850 & 2.46 $\uparrow$   \\
\rowcolor{boxgray}
FEDER++ (Ours) & ResNet50 & 0.025 & 0.863 & 0.957 & 0.898 & 0.066 & 0.797 & 0.889 & 0.810 & 0.029 & 0.742 & 0.917 & 0.820 & 0.042 & 0.830 & 0.916 & 0.851 & 3.59 $\uparrow$\\
FSEL~\cite{sun2024frequency} & ResNet50 & 0.029 & 0.847 & 0.941 & 0.893 & 0.069 & 0.779 & 0.881 & 0.816 & 0.032 & 0.722 & 0.891 & 0.822 & 0.045 & 0.807 & 0.901 & 0.847 & ---\\
FSEL+~\cite{he2026refining} &  ResNet50 & 0.028 & 0.856 & 0.949 & 0.898 & 0.067 & 0.792 & {0.889} & {0.819} & 0.030 & 0.742 & 0.909 & 0.831 & 0.042 & 0.823 & 0.919 & 0.855 & 2.22 $\uparrow$  \\  
\rowcolor{boxgray} FSEL++ (Ours) & ResNet50 & 0.027 & 0.863 & 0.953 & 0.900 & 0.066 & 0.795 & 0.894 & 0.824 & 0.029 & 0.746 & 0.916 & 0.833 & 0.041 & 0.831 & 0.927 & 0.857 & 3.28 $\uparrow$\\
RUN~\cite{he2025run} & ResNet50 & 0.027 & 0.855 & 0.952 & 0.895 & 0.070 & 0.781 & 0.868 & 0.806 & 0.030 & 0.747 & 0.903 & 0.827 & 0.042 & 0.824 & 0.908 & 0.851 & ---\\
RUN+~\cite{he2026refining} & ResNet50 & {0.025} & {0.863} &  {0.960} &  {0.900} & 0.068 &  {0.801} & 0.879 & 0.813 &  {0.029} &  {0.765} &  {0.917} &  {0.836} &  {0.040} &  {0.835} &  {0.921} & {0.858} & 2.13 $\uparrow$  \\
\rowcolor{boxgray} RUN++ (Ours) & ResNet50 & \textbf{0.023} & 0.870 & 0.967 & 0.902 & \textbf{0.066} & 0.808 & 0.886 & 0.819 & 0.027 & 0.775 & 0.926 & 0.838 & \textbf{0.038} & 0.841 & \textbf{0.930} & 0.862 & 4.04 $\uparrow$\\
NUN~\cite{he2026nested} & ResNet50 & {0.026} & {0.862} & {0.955} & {0.899} &  {0.069} &  {0.788} & {0.890} & {0.818} & {0.029} & {0.759} & {0.908} & {0.831} & {0.042} & {0.828} & {0.913} & {0.855} & --- \\
NUN+~\cite{he2026refining} & ResNet50 & 0.025 & 0.868 & 0.960 & 0.903 & 0.068 & 0.797 & 0.898 & 0.824 & 0.028 & 0.765 & 0.915 & 0.835 & 0.041 & 0.833 & 0.917 & 0.858 & 1.19 $\uparrow$ \\
\rowcolor{boxgray}  NUN++ (Ours) & ResNet50 & \textbf{0.023} & \textbf{0.881} & \textbf{0.972} & \textbf{0.908} & \textbf{0.066} & \textbf{0.809} & \textbf{0.910} & \textbf{0.830} & \textbf{0.026} & \textbf{0.778} & \textbf{0.927} & \textbf{0.841} & 0.039 & \textbf{0.842} & 0.922 & \textbf{0.866} & 3.41 $\uparrow$ \\
\midrule
BSA-Net~\cite{zhu2022can} & Res2Net50 & 0.027 & 0.851 & 0.946 & 0.895 & 0.079 & 0.768 & 0.851 & 0.796 & 0.034 & 0.723 & 0.891 & 0.818 & 0.048 & 0.805 & 0.897 & 0.841 & ---\\
RUN~\cite{he2025run} & Res2Net50 & 0.024 & 0.879 & 0.956 & 0.907 & 0.066 & 0.815 & 0.905 & 0.843 & 0.028 & 0.764 & 0.914 & 0.849 & 0.041 & 0.830 & 0.917 & 0.859 & ---\\
RUN+~\cite{he2026refining} & Res2Net50 &  {0.023} &  {0.891} &  {0.963} &  {0.911} & {0.065} &  {0.820} & {0.912} & {0.845} & {0.026} & {0.785} &  {0.933} &  {0.857} &  {0.038} &  {0.852} &  {0.932} &  {0.870} & 2.23 $\uparrow$  \\
\rowcolor{boxgray}  RUN++ (Ours) & Res2Net50 &0.022 & \textbf{0.896} & 0.968 & 0.913 & 0.064 & 0.830 & 0.922 & 0.849 & 0.025 & 0.792 & \textbf{0.938} & 0.861 & 0.037 & 0.859 & 0.936 & \textbf{0.873} & 3.44 $\uparrow$\\
NUN~\cite{he2026nested}  & Res2Net50 & {0.023} & {0.883} & {0.959} &  {0.910} &  {0.066} & {0.822} & {0.909} & {0.846} &  {0.026} & {0.781} & {0.920} &  {0.852} & {0.040} & {0.836} & {0.922} & {0.864} & ---\\
NUN+~\cite{he2026refining} & Res2Net50  & 0.022 & 0.889 & 0.967 & 0.915 & 0.064 & 0.827 & 0.915 & 0.849 & 0.025 & 0.789 & 0.927 & 0.856 & 0.038 & 0.843 & 0.928 & 0.867 & 1.50 $\uparrow$ \\
\rowcolor{boxgray}   NUN++ (Ours) & Res2Net50  & \textbf{0.021} & {0.895} & \textbf{0.973} & \textbf{0.919} & \textbf{0.062} & \textbf{0.834} & \textbf{0.923} & \textbf{0.853} & \textbf{0.023} & \textbf{0.797} & {0.934} & \textbf{0.862} & \textbf{0.035} & \textbf{0.860} & \textbf{0.938} & {0.872} & 3.54 $\uparrow$ \\
\midrule
CamoDiff~\cite{sun2025conditional} & PVT~V2 & 0.022 & 0.868 & 0.952 & 0.908 & 0.042 & 0.853 & 0.936 & 0.878 & 0.019 & 0.815 & 0.943 & 0.883 & 0.028 & 0.858 & 0.942 & 0.895 & ---\\
RUN~\cite{he2025run}  & PVT~V2 & 0.021 & 0.877 & 0.958 & 0.916 & 0.045 & 0.861 & 0.934 & 0.877 & 0.021 & 0.810 & 0.941 & 0.878 & 0.030 & 0.868 & 0.940 & 0.892 & ---\\
RUN+~\cite{he2026refining} & PVT~V2  & {0.019} & {0.893} & {0.971} &  {0.922} & {0.042} & {0.879} & {0.952} & {0.885} & {0.018} & {0.828} & {0.957} &  {0.886} & {0.026} & {0.889} & {0.958} & {0.903} & 3.94 $\uparrow$   \\
\rowcolor{boxgray}
RUN++ (Ours) & PVT~V2 & \textbf{0.018} & 0.899 & 0.975 & 0.925 & 0.041 & 0.885 & 0.958 & 0.888 & 0.017 & 0.836 & \textbf{0.963} & 0.889 & 0.025 & 0.896 & \textbf{0.967} & \textbf{0.908} & 5.34 $\uparrow$\\ 
NUN~\cite{he2026nested}  & PVT~V2 &  {0.021} &  {0.888} &  {0.963} &  {0.918} &  {0.042} &  {0.868} &  {0.939} &  {0.885} &  {0.018} & {0.839} & {0.948} & {0.890} & {0.027} & {0.880} & {0.947} & {0.897} &  ---\\
NUN+~\cite{he2026refining} & PVT~V2  & 0.020 & 0.897 & 0.970 & 0.925 & 0.040 & 0.879 & 0.951 & 0.892 & 0.017 & 0.845 & 0.952 & 0.891 & 0.026 & 0.885 & 0.951 & 0.898 & 1.69 $\uparrow$ \\
\rowcolor{boxgray}   NUN++ (Ours) & PVT~V2  & \textbf{0.018} & \textbf{0.908} & \textbf{0.980} & \textbf{0.930} & \textbf{0.038} & \textbf{0.887} & \textbf{0.968} & \textbf{0.899} & \textbf{0.016} & \textbf{0.857} & {0.962} & \textbf{0.894} & \textbf{0.024} & \textbf{0.897} & 0.960 & 0.905 & 4.16 $\uparrow$ \\
\midrule
VSCode$^{\dagger}$~\cite{luo2024vscode} & DINO V2-L & 0.020 & 0.882 & 0.954 & 0.915 & 0.043 & 0.860 & 0.931 & 0.882 & 0.020 & 0.818 & 0.935 & 0.880 & 0.030 & 0.867 & 0.940 & 0.895 &  --- \\
C3Net$^{\ddagger}$~\cite{jan2025c3net} & DINO V2-L & 0.016 & 0.905 & 0.968 & 0.927 & 0.031 & 0.892 & 0.953 & 0.904 & 0.016 & 0.852 & 0.961 & 0.898 & 0.024 & 0.894 & 0.958 & 0.913 &  --- \\
NUN~\cite{he2026nested} & {{DINO V2-L}} & {0.015} & {0.918} & {0.975} & {0.931} & {0.028} & {0.908} & {0.965} & {0.911} & {0.014} & {0.880} & {0.972} & {0.905} & {0.020} & {0.913} & {0.967} & {0.920} &  --- \\ 
NUN+~\cite{he2026refining} & DINO V2-L  & \textbf{0.014} & 0.922 & 0.978 & 0.932 & 0.027 & 0.912 & 0.970 & 0.913 & 0.013 & 0.886 & 0.976 & 0.908 & 0.018 & 0.919 & 0.973 & 0.923 & 2.03 $\uparrow$ \\
\rowcolor{boxgray} NUN++ (Ours) & DINO V2-L  & \textbf{0.014} & \textbf{0.929} & \textbf{0.982} & \textbf{0.935} & \textbf{0.026} & \textbf{0.918} & \textbf{0.977} & \textbf{0.916} & \textbf{0.012} & \textbf{0.891} & \textbf{0.982} & \textbf{0.910} & \textbf{0.016} & \textbf{0.927} & \textbf{0.978} & \textbf{0.925} & 3.71 $\uparrow$ \\
\bottomrule
\end{tabular}}\vspace{-5mm}
\end{table*}

\subsection{Overall Training Pipeline}
PaceShift leaves the architecture unchanged and adds no inference cost, with modest overhead incurred only during training and the full procedure listed in the Supp.

\noindent\textbf{Overhead.}
Modest training overhead comes only from the periodic augmented and stochastic forward predictions that refresh the Phase-I reliability cues, and from the spectral-candidate and frozen-reference evaluations of Phase~II. The temporal history stores one scalar per refresh checkpoint per sample in memory, and the reduced active subset in Phase~I offsets part of the extra computation (Table~\ref{tab:overhead}).

\noindent\textbf{Hyperparameter footprint.}
All reliability and guard weights derive from the trust level $\trust$ of Table~\ref{tab:guardweights}. The mid-variance map is parameter-free, the preference and temperature scales are both fixed to one, $\beta(t)$ follows the prescribed schedule, and $T_c$ admits an adaptive trigger with comparable performance (\cref{tab:phaseorder}). The remaining global constants $p_{\min}$, $w_{\min}$, $\varepsilon$, and $\lambda_{\mathrm{anc}}$ are fixed once and reused unchanged across all tasks, backbones, and regimes, while $H$ and the phase boundaries follow the common budget-scaling rule in the Supp. Sensitivity is reported in the Supp., and no per-dataset tuning is used.

\section{Experiments}\label{sec:exp}
\subsection{Experimental Setup}\label{sec:exp:setup}
\noindent\textbf{Tasks, datasets, and metrics.} 
We group three core CECS tasks and the related polyp-segmentation task into the CECS suite: camouflaged object detection (COD) on \textit{CAMO}~\cite{le2019anabranch}, \textit{COD10K}~\cite{fan2020camouflaged}, \textit{CHAMELEON}~\cite{skurowski2018animal}, and \textit{NC4K}~\cite{lv2021simultaneously},
polyp image segmentation (PIS) trained on Kvasir~\cite{jha2020kvasir} and ClinicDB~\cite{bernal2015wm} and evaluated on \textit{CVC-ColonDB}~\cite{bernal2012towards} and \textit{ETIS}~\cite{silva2014toward}, transparent object detection (TOD) on \textit{GDD}~\cite{mei2020don} and \textit{GSD}~\cite{lin2021rich}, and zero-shot concealed defect detection (CDD) on \textit{CDS2K}~\cite{fan2023advances}. 
We further test broader generality across infrared small-target detection (\textit{IRSTD-1k}~\cite{zhang2022isnet}), shadow detection (\textit{SBU}~\cite{vicente2016large}), medical lesion and fluid segmentation (\textit{ISIC 2018}~\cite{codella2019skin} and \textit{RETOUCH}~\cite{bogunovic2019retouch}), dichotomous and salient segmentation (\textit{DIS-VD}~\cite{qin2022highly} and \textit{DUTS-TE}~\cite{wang2017learning}), and semantic (\textit{ADE20K}~\cite{zhou2017scene}) and instance (\textit{COCO}~\cite{lin2014microsoft}) segmentation, and supervision generality under point and semi-supervised settings, foundation-model adaptation, and language- and reference-conditioned segmentation. For COD, we report mean absolute error ($M$), adaptive F-measure (\Fb)~\cite{achanta2009frequency}, mean E-measure (\Ep)~\cite{fan2018enhanced}, and structure measure (\Sa)~\cite{fan2017structure}. PIS uses mean Dice, mIoU, and \Sa, TOD uses mIoU, \Fmax, and $M$, and CDD adds the mean F-measure (\Fm) to the COD metrics. Other tasks use their standard metrics.
We assess calibration using pixel-pooled ECE~\cite{guo2017calibration}. We also report a ground-truth-stratified confidence--accuracy gap ($\cag$, the mean of the gaps computed on ground-truth foreground and background pixels), a boundary-restricted gap ($\cagb$), and WConf, the mean confidence on incorrectly predicted pixels. These three are complementary diagnostics of where confidence departs from accuracy, not measures of global calibration, since a calibrated predictor can have a nonzero stratified gap.
Please see details in the Supp.

\begin{table}[t]
\setlength{\abovecaptionskip}{0cm}
\centering
\caption{Training cost under full supervision (ResNet50, batch size $2$, single RTX~4090). Vanilla baselines follow official schedules (e.g., FEDER: $100$ epochs), while CurriSeg and PaceShift use $70$ epochs. Relative Time/Epoch is wall-clock time per epoch normalized by the corresponding vanilla baseline. Peak memory uses \texttt{max\_memory\_allocated}, with SBFT-P views processed sequentially and the frozen reference resident in Phase~II. $\Delta$ is the mean relative accuracy gain from Table~\ref{tab:cod}.}
\label{tab:overhead}
\resizebox{\linewidth}{!}{
\setlength{\tabcolsep}{1.3mm}
\begin{tabular}{@{}l cc cc}
\toprule
{Methods} & Training Time (h) & Relative Time/Epoch & {GPU Memory (G)} & {$\Delta$(\%)}\\
\midrule
{FEDER}\;  & 9.62  & 1.00$\times$  & 1.53 & --\\
FEDER+   & 6.84 & 1.02$\times$ & 1.62 & 2.46\,$\uparrow$\\
\rowcolor{boxgray}
FEDER++ & 7.95 & 1.18$\times$ & 1.81 & 3.59\,$\uparrow$\\
\midrule
{FSEL}\;  & 11.54 & 1.00$\times$ & 2.83 & --\\
FSEL+ & 5.96 & 1.03$\times$ & 2.92 & 2.22\,$\uparrow$\\
\rowcolor{boxgray}
FSEL++ & 6.93 & 1.20$\times$ & 3.15 & 3.28\,$\uparrow$\\
\midrule
{RUN}\;  & 12.64 & 1.00$\times$ & 3.66 & --\\
RUN+ & 8.32 & 1.03$\times$ & 3.75 & 2.13\,$\uparrow$\\
\rowcolor{boxgray}
RUN++ & 9.68 & 1.20$\times$ & 4.02 & 4.04\,$\uparrow$\\
\bottomrule
\end{tabular}}\vspace{-5mm}
\end{table}

\noindent\textbf{Implementation.} PaceShift is implemented in PyTorch with Adam, $(\beta_1,\beta_2)\!=\!(0.9,0.999)$. 
All epoch-scale constants share a single dynamics horizon $H\!=\!10$: an $H$-epoch warm-up without filtering, a difficulty checkpoint refreshed every $H$ epochs, and temporal statistics computed over the Phase-I refresh checkpoints (five at the defaults).
The curriculum phase covers epochs 1 to $T_c\!=\!60$ and the anti-curriculum post-training phase $T_c+1$ to $T\!=\!70$, 
and the percentile schedule grows from $p_{\min}\!=\!0.6$ to $1.0$. Standard augmentations (random flip, crop, and photometric jitter) are shared by all baselines and by their PaceShift counterparts, so comparisons are controlled. Inputs are resized to $352\times352$ for the canonical CECS experiments, while cross-task experiments preserve each host's official resolution, optimizer, learning rate, batch size, augmentation, and total training budget, with PaceShift phase boundaries scaled proportionally with that budget (see Supp.).

\subsection{Comparative Evaluation on CECS}\label{sec:exp:cecs}
\begin{figure*}[t]
\centering
\includegraphics[width=\textwidth]{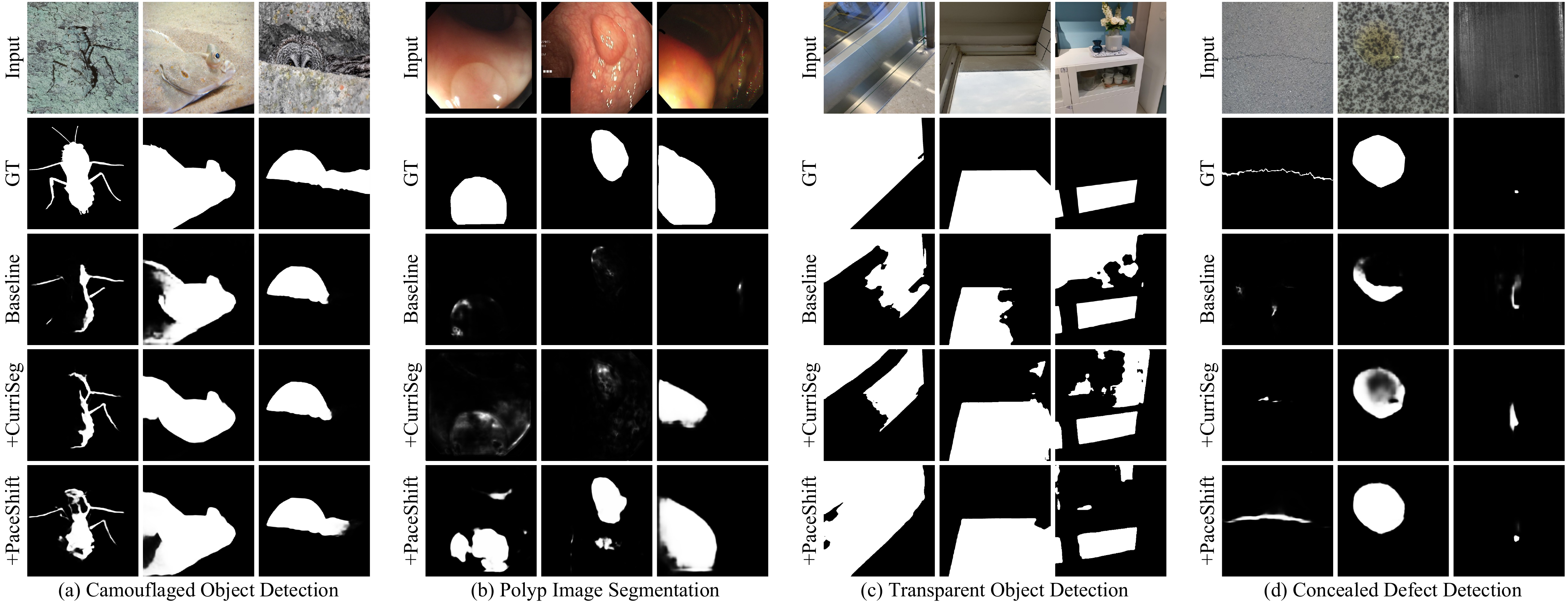}\vspace{-3mm}
\caption{Visual comparisons on COD, PIS, TOD, and CDD using the RUN baseline and its CurriSeg- and PaceShift-trained variants. PaceShift better recovers concealed structures and boundaries, with fewer false positives in texture-rich backgrounds.}\vspace{-5mm}
\label{fig:qual}
\end{figure*}

\noindent\textbf{Camouflaged object detection.} Table~\ref{tab:cod} inserts CurriSeg and PaceShift into four detectors across four backbones and compares them with graph-based, attention-based, transformer, and diffusion detectors. PaceShift improves or matches every detector on every dataset, with mean relative gains of $3.3\%$--$5.3\%$, and NUN++ with DINO~V2-L achieves the strongest overall performance. Fig.~\ref{fig:qual} shows more faithful masks and fewer texture-driven false positives in highly concealed scenes.
Table~\ref{tab:overhead} reports training cost under each recipe's actual schedule. PaceShift increases average per-epoch cost by about $20\%$, with most overhead concentrated in the ten Phase-II epochs. Its reported total wall-clock time is nevertheless lower because the curriculum recipes use $70$ epochs whereas the vanilla baselines follow their longer official schedules. For example, FEDER uses $100$ epochs, while FEDER+ and FEDER++ use $70$ and achieve higher accuracy. Inference cost is unchanged.
Table~\ref{tab:sig} reports five paired runs on COD10K. Both CurriSeg ($+$) and PaceShift ($++$) improve the vanilla baseline at $p<0.001$ on all three metrics. On the same paired runs, PaceShift further improves over CurriSeg on \Fb\ and \Ep\ for all three detectors ($p<0.01$), whereas the smaller \Sa\ gains ($0.002$) are not statistically significant.

\begin{table}[t]
\setlength{\abovecaptionskip}{0cm}
\centering
\caption{Statistical significance on COD10K (ResNet50), mean$\pm$std over five paired runs with shared seeds. Two-sided paired $t$-tests report the maximum $p$-value over \Fb, \Ep, and \Sa. Tests of $++$ against $+$ are reported in the text.
}
\label{tab:sig}
\resizebox{\linewidth}{!}{
\setlength{\tabcolsep}{1.85mm}
\begin{tabular}{@{}l ccc c}
\toprule
Methods & \Fb\up & \Ep\up & \Sa\up & max. $p$ (vs.\ base) \\
\midrule
FEDER           & 0.715\,$\pm$.004 & 0.892\,$\pm$.003 & 0.810\,$\pm$.002 & --- \\
FEDER+ & 0.736\,$\pm$.003 & 0.910\,$\pm$.003 & 0.818\,$\pm$.002 & $<$0.001 \\
\rowcolor{boxgray}
FEDER++ & \best{0.742}\,$\pm$.003 & \best{0.917}\,$\pm$.002 & \best{0.820}\,$\pm$.001 & $<$0.001 \\
\midrule
FSEL            & 0.722\,$\pm$.004 & 0.891\,$\pm$.003 & 0.822\,$\pm$.002 & ---\\
FSEL+ & 0.742\,$\pm$.003 & 0.909\,$\pm$.002 & 0.831\,$\pm$.001 & $<$0.001 \\
\rowcolor{boxgray}
FSEL++ & \best{0.746}\,$\pm$.003 & \best{0.916}\,$\pm$.002 & \best{0.833}\,$\pm$.001 & $<$0.001 \\
\midrule
RUN             & 0.747\,$\pm$.004 & 0.903\,$\pm$.003 & 0.827\,$\pm$.002 & ---\\
RUN+ & 0.765\,$\pm$.003 & 0.917\,$\pm$.002 & 0.836\,$\pm$.001 & $<$0.001\\
\rowcolor{boxgray}
RUN++ & \best{0.775}\,$\pm$.002 & \best{0.926}\,$\pm$.002 & \best{0.838}\,$\pm$.001 & $<$0.001 \\
\bottomrule
\end{tabular}}\vspace{-4mm}
\end{table}

\begin{table}[t]
\setlength{\abovecaptionskip}{0cm} 
\centering
\caption{Polyp image segmentation (PVT~V2 backbone).}
\label{tab:pis}
\resizebox{\linewidth}{!}{
\setlength{\tabcolsep}{1.2mm}
\begin{tabular}{@{}lcccccc c}
\toprule
\multirow{2}{*}{Methods} & \multicolumn{3}{c}{CVC-ColonDB} & \multicolumn{3}{c}{ETIS} & PIS\\
\cmidrule(lr){2-4}\cmidrule(lr){5-7}
 & mDice\up & mIoU\up & \Sa\up & mDice\up & mIoU\up & \Sa\up & $\Delta$(\%)\\
\midrule
PolypPVT~\cite{dong2023polyp} & 0.808 & 0.727 & 0.865 & 0.787 & 0.706 & 0.871 & ---\\
CoInNet~\cite{jain2023coinnet}  & 0.797 & 0.729 & 0.875 & 0.759 & 0.690 & 0.859 & ---\\
LSSNet~\cite{wang2024lssnet}   & 0.820 & 0.741 & 0.867 & 0.779 & 0.701 & 0.867 & ---\\
MEGANet~\cite{bui2024meganet}  & 0.793 & 0.714 & 0.854 & 0.739 & 0.665 & 0.836 & ---\\
MEGANet+~\cite{he2026refining} & 0.815 & 0.736 & 0.866 & 0.755 & 0.682 & 0.848 & 2.24 $\uparrow$   \\
\rowcolor{boxgray}
MEGANet++ (Ours) &0.821 & 0.743 & 0.869 & 0.765 & 0.688 & 0.851 & 3.02 $\uparrow$\\
RUN~\cite{he2025run}      & 0.822 & 0.742 & 0.880 & 0.788 & 0.709 & 0.878 & ---\\
RUN+~\cite{he2026refining} & {0.845} & {0.763} &  {0.893} &  {0.806} &  {0.720} &  {0.890} & 2.05 $\uparrow$ \\
\rowcolor{boxgray}
RUN++ (Ours)     & \textbf{0.852} & \textbf{0.772} & \textbf{0.896} & \textbf{0.815} & \textbf{0.728} & \textbf{0.894} & 2.91 $\uparrow$\\
\bottomrule
\end{tabular}}\vspace{-5mm}
\end{table}

\begin{table}[t]
\setlength{\abovecaptionskip}{0cm} 
\centering
\caption{Transparent object detection (PVT~V2 backbone).}
\label{tab:tod}
\resizebox{\linewidth}{!}{
\setlength{\tabcolsep}{1.5mm}
\begin{tabular}{@{}lcccccc c}
\toprule
\multirow{2}{*}{Methods} & \multicolumn{3}{c}{GDD} & \multicolumn{3}{c}{GSD} & TOD\\
\cmidrule(lr){2-4}\cmidrule(lr){5-7}
 & mIoU\up & \Fmax\up & $M$\dn & mIoU\up & \Fmax\up & $M$\dn & $\Delta$(\%)\\
\midrule
EBLNet~\cite{he2021enhanced}      & 0.870 & 0.922 & 0.064 & 0.817 & 0.878 & 0.059 & ---\\
IEBAF~\cite{han2023internal}       & 0.887 & 0.944 & 0.056 & 0.861 & 0.926 & 0.049 & ---\\
GhostingNet~\cite{yan2024ghostingnet} & 0.893 & 0.943 & 0.054 & 0.838 & 0.904 & 0.055 & ---\\
RFENet~\cite{fan2023rfenet} & 0.886 & 0.938 & 0.057 & 0.865 & 0.931 & 0.048 & ---\\
RFENet+~\cite{he2026refining} &  0.893 & 0.952 & 0.052 & 0.875 & 0.939 & 0.045 & 3.22 $\uparrow$   \\
\rowcolor{boxgray}
RFENet++ (Ours)     & 0.898     & 0.965     & 0.050     & 0.878     & 0.942     & 0.043  & 4.94 $\uparrow$\\
RUN~\cite{he2025run}         & 0.895 & 0.952 & 0.051 & 0.866 & 0.938 & 0.043 & ---\\
RUN+~\cite{he2026refining} &  {0.902} &  {0.966} & {0.047} &  {0.878} &  {0.945} & {0.039} & 3.59 $\uparrow$ \\
\rowcolor{boxgray}
RUN++ (Ours)        & \textbf{0.906}     & \textbf{0.979}     & \textbf{0.045}     & \textbf{0.887}     & \textbf{0.958}     & \textbf{0.037}    & 5.72 $\uparrow$\\
\bottomrule
\end{tabular}}\vspace{-5mm}
\end{table}

\begin{table}[t]
\setlength{\abovecaptionskip}{0cm} 
\centering
\caption{Concealed defect detection on CDS2K.}
\label{tab:cdd}
\resizebox{\linewidth}{!}{
\setlength{\tabcolsep}{2.5mm}
\begin{tabular}{@{}lccccc c}
\toprule
Methods & \Sa\up & $M$\dn & \Ep\up & \Fb\up & \Fm\up & $\Delta$(\%)\\
\midrule
SINet V2~\cite{fan2021concealed}  & 0.551 & 0.102 & 0.567 & 0.223 & 0.248 & ---\\
HitNet~\cite{hu2022high}    & 0.563 & 0.118 & 0.564 & 0.298 & 0.298 & ---\\
OAFormer~\cite{yang2023oaformer}  & 0.541 & 0.121 & 0.535 & 0.216 & 0.239 & ---\\
CamoFormer~\cite{yin2024camoformer} & 0.589 & 0.100 & 0.588 & 0.330 & 0.329 & ---\\
CamoFormer+~\cite{he2026refining} & 0.599 & 0.091 & 0.603 & 0.345 & 0.346 & 4.59 $\uparrow$  \\
\rowcolor{boxgray}
CamoFormer++ (Ours) & \textbf{0.607} & 0.088 & 0.617 & \textbf{0.359} & \textbf{0.361}  & 7.70 $\uparrow$\\
RUN~\cite{he2025run}  & 0.590 & 0.068 & 0.595 & 0.298 & 0.299 & ---\\
RUN+~\cite{he2026refining}   &   {0.598} & {0.061} & {0.612} &  {0.309} &  {0.310} & 4.38 $\uparrow$     \\
\rowcolor{boxgray}
RUN++ (Ours)      & \textbf{0.607} & \textbf{0.058} & \textbf{0.626} & 0.323 & 0.324  & 7.91 $\uparrow$\\
\bottomrule
\end{tabular}}\vspace{-5mm}
\end{table}

\begin{table}[t]
\setlength{\abovecaptionskip}{0cm}
\centering
\caption{Generality across architectures on COD10K. ``+'' and ``++'' denote CurriSeg~\cite{he2026refining} and PaceShift. Base is the baseline's \Sa, and the two $\Delta$ columns report the mean relative gain of each recipe over its baseline across four metrics ($M$, \Fb, \Ep, \Sa). For UGTR, the $+$ cell reports the standard CurriSeg recipe in \cite{he2026refining}.}
\label{tab:archgen}
\resizebox{\linewidth}{!}{
\setlength{\tabcolsep}{1.15mm}
\begin{tabular}{l l l ccc}
\toprule
Paradigms & Model & Backbone & Base (\Sa) & {$\Delta$(\%, +)} & \cellcolor{boxgray}{$\Delta$(\%, ++)}\\
\midrule
\multicolumn{6}{l}{\textit{Convolutional detectors}}\\
\midrule
Enc.-dec.      & SINet-V2~\cite{fan2021concealed}     & Res2Net50 & 0.815 & 1.76 $\uparrow$ & \cellcolor{boxgray}\best{4.23} $\uparrow$ \\
Graph reason.  & MGL-R~\cite{zhai2021mutual}          & ResNet50  & 0.814 & 2.49 $\uparrow$ & \cellcolor{boxgray}\best{4.77} $\uparrow$ \\
Bio-inspired   & PreyNet~\cite{zhang2022preynet}      & ResNet50  & 0.813 & 2.54 $\uparrow$ & \cellcolor{boxgray}\best{5.10} $\uparrow$ \\
Edge$+$freq.   & FEDER~\cite{He2023Camouflaged}       & ResNet50  & 0.810 & 3.05 $\uparrow$ & \cellcolor{boxgray}\best{4.30} $\uparrow$ \\
\midrule
\multicolumn{6}{l}{\textit{Attention and transformer detectors}}\\
\midrule
Masked attn.   & CamoFormer~\cite{yin2024camoformer}  & PVT~V2 & 0.872 & 3.22 $\uparrow$ & \cellcolor{boxgray}\best{5.88} $\uparrow$ \\
Multi-scale    & ZoomNeXt~\cite{pang2024zoomnext}     & PVT~V2 & 0.887 & 2.07 $\uparrow$ & \cellcolor{boxgray}\best{4.14} $\uparrow$ \\
Freq.\ attn.   & FSEL~\cite{sun2024frequency}         & ResNet50 & 0.822 & 3.03 $\uparrow$ & \cellcolor{boxgray}\best{4.21} $\uparrow$ \\
Uncertainty    & UGTR~\cite{yang2021uncertainty}      & ResNet50 & 0.817 & 2.69  $\uparrow$ & \cellcolor{boxgray}\best{5.99} $\uparrow$\\
\midrule
\multicolumn{6}{l}{\textit{Generative and unfolding models}}\\
\midrule
Diffusion      & CamoDiff~\cite{sun2025conditional}   & PVT~V2 & 0.883 & 1.90 $\uparrow$ & \cellcolor{boxgray}\best{3.79} $\uparrow$ \\
Unfolding      & RUN~\cite{he2025run}                 & PVT~V2 & 0.878 & 4.78 $\uparrow$ & \cellcolor{boxgray}\best{6.46} $\uparrow$ \\
Nested unfold. & NUN~\cite{he2026nested}              & PVT~V2 & 0.890 & 1.70 $\uparrow$ & \cellcolor{boxgray}\best{3.80} $\uparrow$ \\
\midrule
\multicolumn{6}{l}{\textit{Foundation-model based}}\\
\midrule
Found.\ backb. & NUN~\cite{he2026nested} & DINO~V2-L & 0.905 & 2.14 $\uparrow$ & \cellcolor{boxgray}\best{4.28} $\uparrow$ \\
Found.\ adapt. & SAM-a~\cite{chen2023sam} & ViT-H & 0.883 & 3.29 $\uparrow$ & \cellcolor{boxgray}\best{4.70} $\uparrow$ \\
\bottomrule
\end{tabular}}\vspace{-5mm}
\end{table}

\begin{table}[t]
\setlength{\abovecaptionskip}{0cm} 
\centering
\caption{Generalization to broader dense prediction families. One primary metric per family. ``+'' and ``++'' denote the CurriSeg and PaceShift recipes on the listed baseline, and $\Delta$ is the relative gain of ``++'' over it. Only the training strategy changes, and splits follow each baseline.}
\label{tab:broaddense}
\resizebox{\linewidth}{!}{
\setlength{\tabcolsep}{0.6mm}
\begin{tabular}{l l l l ccc c}
\toprule
Tasks & Datasets & Baselines & Metrics & Base & {+} &\cellcolor{boxgray} {++} & $\Delta$(\%)\\
\midrule
\multicolumn{8}{l}{\textit{Tier 1: low-contrast detection with ambiguous annotation}}\\
\midrule
IRSTD & IRSTD-1k~\cite{zhang2022isnet} & IRSAM~\cite{zhang2024irsam} & IoU\up & 73.7 & 75.5 & \cellcolor{boxgray} \best{77.2} & 4.75 $\uparrow$\\
Shadow det.         & SBU~\cite{vicente2016large}    & Spider~\cite{zhao2024spider}             & BER\dn & 0.040 & 0.038 &\cellcolor{boxgray} \best{0.037} & 7.50 $\uparrow$\\
\midrule
\multicolumn{8}{l}{\textit{Tier 2: low-contrast clinical dense prediction}}\\
\midrule
Skin lesion    & ISIC~2018~\cite{codella2019skin}   & DconnNet~\cite{yang2023directional} & DSC\up & 90.4 & 90.7 &\cellcolor{boxgray} \best{92.8} & 2.65 $\uparrow$\\
Reti. OCT & RETOUCH~\cite{bogunovic2019retouch} & DconnNet~\cite{yang2023directional} & DSC\up & 87.7 & 87.5 &\cellcolor{boxgray} \best{89.6} & 2.17 $\uparrow$\\
\midrule
\multicolumn{8}{l}{\textit{Tier 3: standard dense prediction}}\\
\midrule
Dicho. seg.    & DIS-VD~\cite{qin2022highly}    & MVANet~\cite{yu2024multi}  & $F_\beta^{\omega}$\up & 0.855 & 0.871 &\cellcolor{boxgray} \best{0.887} & 3.74 $\uparrow$\\
SOD & DUTS-TE~\cite{wang2017learning} & VST-T++~\cite{liu2024vst++}   & \Fb\up  & 0.869 & 0.882 & \cellcolor{boxgray} \best{0.897} & 3.22 $\uparrow$\\ 
Semantic seg.      & ADE20K~\cite{zhou2017scene}     & CGRSeg~\cite{ni2024context} & mIoU\up & 45.5 & 46.8 &\cellcolor{boxgray} \best{47.8} & 5.05 $\uparrow$\\
Instance seg.      & COCO~\cite{lin2014microsoft}    & M2Form~\cite{cheng2022masked} & AP\up   & 38.0 & 38.6 &\cellcolor{boxgray} \best{39.4} & 3.68 $\uparrow$\\ 
\bottomrule
\end{tabular}}\vspace{-5mm}
\end{table}

\begin{table}[t]
\setlength{\abovecaptionskip}{0cm}
\centering
\caption{Online training strategies under full dataset access (RUN, ResNet50, COD10K) at a fixed training horizon. ``SA-RCS only'' uses the complete Phase-I recipe and removes only Phase~II.}
\label{tab:online}
\resizebox{\linewidth}{!}{
\setlength{\tabcolsep}{1.6mm}
\begin{tabular}{@{}l c cccc c}
\toprule
Strategies & Method source & $M$\dn & \Fb\up & \Ep\up & \Sa\up & $\Delta$(\%) \\
\midrule
Vanilla (RUN)~\cite{he2025run}      & ICML25     & 0.030 & 0.747 & 0.903 & 0.827  & --- \\
Loss curriculum                     & ---        & 0.031 & 0.742 & 0.898 & 0.824 & 1.23 $\downarrow$ \\
Co-teaching~\cite{han2018co}        & NeurIPS18  & 0.030 & 0.752 & 0.905 & 0.829  & 0.28 $\uparrow$ \\
Meta-Weight-Net~\cite{shu2019meta}  & NeurIPS19  & 0.029 & 0.751 & 0.908 & 0.830 & 1.20 $\uparrow$ \\
InfoBatch~\cite{qin2024infobatch}   & ICLR24     & 0.030 & 0.748 & 0.905 & 0.828  & 0.12 $\uparrow$ \\
SeTa~\cite{zhou2025scale}            & CVPR25     & 0.030 & 0.753 & 0.906 & 0.829 & 0.34 $\uparrow$ \\
IES~\cite{yuan2025instance}              & ICLR25     & 0.029 & 0.752 & 0.909 & 0.831 & 1.29 $\uparrow$ \\
Dyn-Unc (online)~\cite{he2024large} & CVPRW24    & 0.029 & 0.758 & 0.910 & 0.832 & 1.55 $\uparrow$ \\
TDDS (online)~\cite{zhang2024spanning}  & CVPR24     & 0.028 & 0.757 & 0.912 & 0.833 & 2.43 $\uparrow$ \\
\midrule
+CurriSeg~\cite{he2026refining}     & ICML26     & 0.029 & 0.765 & 0.917 & 0.836 & 2.10 $\uparrow$ \\
+ SA-RCS only (Ours)                  & ---        & 0.029 & 0.763 & 0.918 & 0.835 & 2.03 $\uparrow$ \\
\rowcolor{boxgray}
\textbf{+PaceShift (Ours)}          & ---        & \best{0.027} & \best{0.775} & \best{0.926} & \best{0.838} & 4.41 $\uparrow$ \\
\bottomrule
\end{tabular}}\vspace{-4mm}
\end{table}
\begin{figure}[t]
\centering
\includegraphics[width=0.95\linewidth]{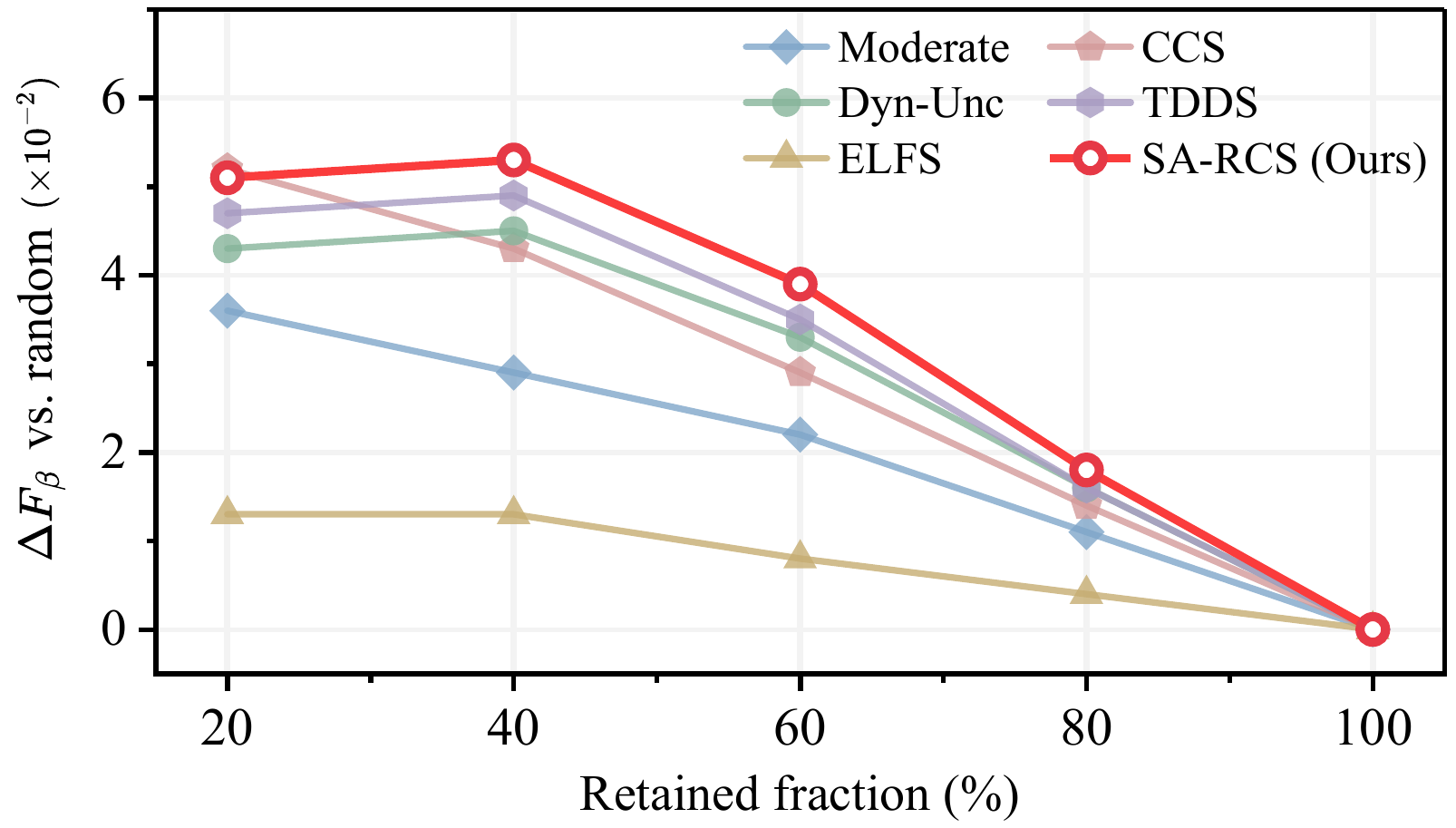}\vspace{-3mm}
\caption{One-shot coreset selection on COD10K (RUN, ResNet50), showing absolute \Fb\ differences ($\times10^{-2}$) from random selection at each retention rate. We compare the SA-RCS diagnostic with Moderate~\cite{xia2022moderate}, CCS~\cite{zheng2022coverage}, Dyn-Unc~\cite{he2024large}, TDDS~\cite{zhang2024spanning}, and label-free ELFS~\cite{zheng2025elfs}. All differences vanish at $100\%$ retention, where selection is inactive. Vanilla achieves \Fb=$0.747$.
}\vspace{-5mm}
\label{fig:retention}
\end{figure}

\begin{table}[t]
\setlength{\abovecaptionskip}{0cm} 
\centering
\caption{Weak supervision on COD10K (supervision: point, backbone: PVT~V2, foundation model: SAM3). ``$+$'' denotes CurriSeg adapted to this regime as described in the Supp., and ``$++$'' denotes PaceShift.}
\label{tab:weak}
\resizebox{\linewidth}{!}{
\setlength{\tabcolsep}{1.8mm}
\begin{tabular}{@{}lcccccc}
\toprule
Methods         & Method source   & $M$\dn & \Fb\up & \Ep\up & \Sa\up    & $\Delta$(\%) \\ \midrule
WS-SAM~\cite{he2023weaklysupervised}          & NeurIPS23 & 0.035 & 0.724 & 0.865 & 0.808 & ---      \\
SEE~\cite{he2025segment}             & TPAMI25   & 0.033 & 0.735 & 0.866 & 0.814 & ---      \\
SEE+~\cite{he2026refining}            & ICML26    & 0.031 & 0.746 & 0.878 & 0.821 & 2.45 \up    \\
\rowcolor{boxgray} SEE++ (Ours)    & ---       & 0.029 & 0.761 & 0.892 & 0.827 & 5.06 \up    \\
SCALER~\cite{he2025scaler}          & ECCV26    & 0.029 & 0.758 & 0.897 & 0.827 & ---      \\
SCALER+~\cite{he2026refining}         & ICML26    & 0.028 & 0.767 & 0.906 & 0.829 & 1.47 \up    \\
\rowcolor{boxgray} SCALER++ (Ours) & ---       & 0.026 & 0.780 & 0.915 & 0.834 & 4.03 \up    \\ \bottomrule
\end{tabular}}\vspace{-5mm}
\end{table}
\looseness=-1
\noindent\textbf{Polyp, transparent, and concealed defect.}
\cref{tab:pis,tab:tod,tab:cdd} cover the remaining three CECS tasks. On PIS and TOD, PaceShift improves both a task-specific detector, MEGANet~\cite{bui2024meganet} and RFENet~\cite{fan2023rfenet} respectively, and the RUN segmenter, and it consistently exceeds the conference recipe. On CDD, where a COD-trained model is transferred to CDS2K without adaptation, the relative gains are the largest of the four tasks, roughly $7.7\%$ to $7.9\%$, and the gain of PaceShift substantially exceeds that of CurriSeg. Fig.~\ref{fig:qual} gives qualitative comparisons on all four tasks.

\subsection{Generalization across Architectures}
\label{sec:exp:archgen}
PaceShift replaces a training schedule, not a module, so it should compose with any segmenter. Table~\ref{tab:archgen} attaches the identical recipe, with identical global constants, to thirteen configurations of twelve detectors spanning encoder-decoder, graph, bio-inspired, frequency, attention, uncertainty, diffusion, unfolding, and foundation-model designs. 
Each baseline keeps its own backbone, optimizer, and augmentation, and only the training recipe changes. Six rows reuse runs reported elsewhere: FEDER, FSEL, RUN, and the two NUN rows are the COD10K results of the canonical runs in Table~\ref{tab:cod}, and the SAM-Adapter row repeats Table~\ref{tab:foundation}. The remaining seven detectors are trained under their native epoch budgets with the phase boundaries scaled as in the Supp. Three observations follow. 
(1)~Every configuration improves, within a band of $3.8$ to $6.5$ percent, and PaceShift exceeds CurriSeg on all thirteen rows. 
(2)~It is complementary to architectural modeling of the very quantities it controls: UGTR already reasons about uncertainty inside the network and is nonetheless among the larger beneficiaries, and FSEL, which models frequency, still gains, because such a module changes \emph{how features are computed} while the schedule changes \emph{what training signal the optimizer receives}.
(3)~The gains persist under the adaptation regime that is replacing task-specific training: DINO~V2 and the frozen SAM adapter improve by $4.3$ and $4.7$ percent without any change to their adapters or frozen weights.

\subsection{Generalization to Broader Dense Prediction Families}
\label{sec:exp:moredenseprediction}
To assess PaceShift beyond CECS, we evaluate increasingly diverse dense-prediction tasks exhibiting high-frequency shortcuts, supervision-reliability variation, both, or neither prominently (Sec.~\ref{sec:dynamics}).
Tier~1 collects low-contrast detection with ambiguous annotation, Tier~2 low-contrast clinical dense prediction, and Tier~3 salient object detection and dichotomous segmentation with standard semantic and instance segmentation, the last using the multi-class instantiation of Sec.~\ref{sec:sarcs}. For each family, Table~\ref{tab:broaddense} keeps one strong published baseline's architecture, optimizer, augmentation, and epoch budget, changes only the training strategy with no constant retuned, and reports the family's primary metric.

\noindent\textbf{Tier 1: low-contrast detection with ambiguous annotation.}
Infrared small-target detection (IRSTD) and shadow detection are low-contrast but not texture-matched, and both carry substantial annotation ambiguity~\cite{vicente2016large}. Gains persist here although the texture shortcut is weaker, indicating that the full formulation is not limited to shortcut-dominated data.

\noindent\textbf{Tier 2: clinical dense prediction.}
Skin lesion and retinal OCT fluid segmentation combine low boundary contrast with costly, uncertain annotation, making supervision quality and boundary ambiguity particularly relevant to the reliability formulation. On RETOUCH, CurriSeg falls marginally below the baseline, whereas PaceShift recovers this loss and further improves performance under the same clinical setting.

\noindent\textbf{Tier 3: standard dense prediction.}
Dichotomous, salient, semantic, and instance segmentation are not defined by a texture-matched background. PaceShift nevertheless yields consistent gains under the native host budgets, with the same proportional stabilize-then-challenge schedule and without introducing task-specific constants. The multi-class and instance rows use the corresponding output-form instantiations of Sec.~\ref{sec:sarcs}. These results support transfer of the full training paradigm beyond context-entangled settings.

\subsection{Comparison with Data-Centric Training and Selection}
\label{sec:exp:datacentric}
Data-centric methods split into two families with different objectives: online strategies change which samples contribute during one run, whereas one-shot coreset scores select a subset before retraining and collapse to the vanilla result at full retention. We therefore compare the two families under separate protocols, all on a fixed RUN (ResNet50) backbone on COD10K with shared optimization rules.

\begin{table}[t]
\setlength{\abovecaptionskip}{0cm} 
\centering
\caption{Semi-supervision on COD10K ($1/16$ labeled training data, PVT~V2, SAM3). ``$+$'' denotes CurriSeg adapted to this regime as detailed in the Supp., and ``$++$'' denotes PaceShift.}
\label{tab:semi}
\resizebox{\linewidth}{!}{
\setlength{\tabcolsep}{1.6mm}
\begin{tabular}{@{}lcccccc}
\toprule
Methods         & Method source   & $M$\dn & \Fb\up & \Ep\up & \Sa\up    & $\Delta$(\%) \\ \midrule
CoSOD~\cite{chakraborty2024unsupervised}    & WACV24  & 0.053 & 0.674 & 0.807 & 0.766 & ---      \\
SEE~\cite{he2025segment}      & TPAMI25 & 0.043 & 0.713 & 0.825 & 0.771 & ---      \\
SEE+~\cite{he2026refining}     & ICML26  & 0.041 & 0.723 & 0.836 & 0.776 & 2.01 \up    \\
\rowcolor{boxgray} SEE++ (Ours)   & ---     & 0.038 & 0.740 & 0.848 & 0.783 & 4.94 \up    \\
SCALER~\cite{he2025scaler}   & ECCV26  & 0.034 & 0.733 & 0.849 & 0.790 & ---      \\
SCALER+~\cite{he2026refining}  & ICML26  & 0.033 & 0.739 & 0.855 & 0.793 & 1.21 \up    \\
\rowcolor{boxgray} SCALER++ (Ours) & ---     & 0.031 & 0.748 & 0.863 & 0.797 & 3.35 \up    \\ \bottomrule
\end{tabular}}\vspace{-3mm}
\end{table}
\begin{table}[t]
\setlength{\abovecaptionskip}{0cm}
\centering
\caption{Corrupted pseudo-label detection on COD10K under the $1/16$ semi-supervised setting. We corrupt $20\%$ of the EMA-teacher pseudo-masks, and the frozen SAM3 proposal is never corrupted. Each criterion flags its lowest-scoring $20\%$, matching the corruption rate, so flagging precision and recall coincide (Flag P/R, random $0.200$). ``Ret.\ PL \Sa'' is the mean quality of the retained pseudo-labels against held-out ground truth, used only for evaluation. AUROC and AUPRC are threshold-free (random $0.50$ and about $0.20$). ``Pseudo-mask--SAM3 agreement'' scores every evaluated pseudo-mask, corrupted or not, directly by its IoU with the frozen proposal, without the segmenter.}
\label{tab:triage}
\resizebox{\linewidth}{!}{
\setlength{\tabcolsep}{1.3mm}
\begin{tabular}{@{}l cccc}
\toprule
Criteria & Flag P/R\up & Ret.\ PL \Sa\up & AUROC\up & AUPRC\up\\
\midrule
Confidence filter            & 0.589 & 0.801 & 0.682 & 0.641\\
Entropy only                 & 0.631 & 0.807 & 0.708 & 0.669\\
Peer small-loss              & 0.643 & 0.810 & 0.723 & 0.687\\
Dyn-Unc~\cite{he2024large}  & 0.702 & 0.817 & 0.774 & 0.735\\
\midrule
Pseudo-mask--SAM3 agreement  & 0.735 & 0.822 & 0.806 & 0.774\\
Prediction--SAM3 agreement   & 0.681 & 0.814 & 0.759 & 0.719\\
$\rho_i$ w/o ext.\ proposal  & 0.724 & 0.820 & 0.792 & 0.761\\
\rowcolor{boxgray}
{Reliability $\rho_i$} & \best{0.769} & \best{0.827} & \best{0.827} & \best{0.801}\\
\bottomrule
\end{tabular}}\vspace{-4mm}
\end{table}
\noindent\textbf{Adaptation of scores designed for classification.}
Per-pixel quantities use a class-balanced spatial mean so background pixels do not dominate, trajectory scores share one checkpoint cache, and coreset methods read the same pooled encoder feature. For the online comparison, Dyn-Unc and TDDS are mapped to sample weights by the same normalization and floor $w_{\min}$ as \cref{eq:omega}, and the retention study keeps their one-shot form. Per-score definitions are in the Supp.

\noindent\textbf{Online strategies.}
Three observations follow from Table~\ref{tab:online}. A naive loss curriculum hurts under this schedule, consistent with Sec.~\ref{sec:dynamics} and with its null effect under the matched budget of Table~\ref{tab:ordering}: easy, clean samples first let the model commit to the texture shortcut. Second, the trajectory scores Dyn-Unc and TDDS are the strongest prior baselines, which supports our premise that the relevant signal is a trajectory property. Third, SA-RCS alone exceeds both on \Fb, \Ep, and \Sa\ while remaining comparable in $M$, and the full PaceShift widens the margin: the distinction is the supervision-adaptive reliability score, its online refresh, the pixel-level coupling, and the guarded post-training, none of which these efficiency-oriented scores designed for classification provide.

\noindent\textbf{Fixed-retention coreset study.}
Fig.~\ref{fig:retention} evaluates a representative one-shot score per family as the \Fb\ gain over random selection. Since PaceShift is online and emits no static coreset, its diagnostic averages the Phase-I weight, $s_i^{\mathrm{SA}}=\frac{1}{T_c-T_w}\sum_{t=T_w+1}^{T_c}\omega_i^{(t)}$, retains the top fraction, and retrains from scratch. Our diagnostic gives the largest gain at $40$--$80\%$ retention and a competitive one at $20\%$. PaceShift does not compete there, since it uses the full set and targets accuracy, robustness, and calibration instead of subset reduction.

\subsection{Learning under Incomplete Supervision}
\label{sec:exp:incomplete}
Because ``supervision-adaptive'' is central to our claim, weak and semi-supervision are treated as primary experiments: a point-annotated weak regime and a scarce $1/16$ semi-supervised regime, both driving curriculum selection by the reliability score $\rho_i$ of \cref{eq:rho} in place of a ground-truth IoU.

\noindent\textbf{Weak supervision.}
Table~\ref{tab:weak} layers PaceShift on two point-supervised COD segmenters, SEE~\cite{he2025segment} and SCALER~\cite{he2025scaler}. 
For a point cue the supervision-agreement term of \cref{eq:rho} averages point-cue satisfaction (foreground-disk coverage with a background-seed penalty) and agreement with the SAM3 mask proposal~\cite{carion2025sam3segmentconcepts}, while the augmentation-consistency, model-disagreement, and entropy cues are unchanged, so the curriculum discounts pseudo-masks that disagree with all available cues, not the model's own confident ones.
Using no additional dense label, PaceShift (``{++}'') improves both baselines on every metric, and its mean relative gain more than doubles that of CurriSeg (``{+}'') in each case, precisely where the ground-truth difficulty $d_i^{\mathrm{full}}$ of \cref{eq:dfull} is undefined.

\begin{table}[t]
\setlength{\abovecaptionskip}{0cm}
\centering
\caption{Foundation-model adapters on COD10K under full supervision. ``+'' and ``++'' denote CurriSeg and PaceShift recipes.}
\label{tab:foundation}
\resizebox{\linewidth}{!}{
\setlength{\tabcolsep}{2.5mm}
\begin{tabular}{@{}l ccccc}
\toprule
Methods & \MAE\dn & \Fb\up & \Ep\up & \Sa\up & $\Delta$(\%) \\
\midrule
SAM-Adapter~\cite{chen2023sam}            & 0.025 & 0.800 & 0.918 & 0.883 & ---\\
SAM-Adapter+~\cite{he2026refining} & 0.023 & 0.822 & 0.935 & 0.888 & 3.29 $\uparrow$ \\
\rowcolor{boxgray}
SAM-Adapter++ (Ours)         & 0.022 & 0.827 & 0.940 & 0.892 & 4.70 $\uparrow$\\
SAM2-Adapter~\cite{chen2024sam2}           & 0.018 & 0.848 & 0.950 & 0.899 & ---\\
SAM2-Adapter+~\cite{he2026refining} & 0.016 & 0.869 & 0.962 & 0.903 & 3.82 $\uparrow$ \\
\rowcolor{boxgray}
SAM2-Adapter++ (Ours)        & 0.015 & 0.876 & 0.969 & 0.906 & 5.69 $\uparrow$\\
SAM3-Adapter~\cite{chen2025sam3}           & 0.015 & 0.883 & 0.965 & 0.927& ---\\
SAM3-Adapter+~\cite{he2026refining} & 0.014 & 0.892 & 0.968 & 0.930 & 2.08 $\uparrow$  \\
\rowcolor{boxgray}
SAM3-Adapter++ (Ours)        & \textbf{0.013} & {\textbf{0.901}} &  \textbf{0.975} & \textbf{0.932} & 4.24 $\uparrow$ \\
\bottomrule
\end{tabular}}\vspace{-5mm}
\end{table}
\begin{table}[t]
\setlength{\abovecaptionskip}{0cm}
\centering
\caption{VLM-based weakly supervised semantic segmentation (mIoU\,\%). ``++'' adds our PaceShift to each CLIP-based baseline.}
\label{tab:vlm-ws}
\resizebox{\linewidth}{!}{
\setlength{\tabcolsep}{2.5mm}
\begin{tabular}{@{}lccccc}
\toprule
\multirow{2}{*}{Methods} & \multirow{2}{*}{Method source} & \multicolumn{2}{c}{VOC} & COCO & \multirow{2}{*}{$\Delta$(\%)} \\ \cmidrule(lr){3-4} \cmidrule(lr){5-5}
                         &                          & Val.       & Test       & Val. &                           \\ \midrule
DIAL~\cite{jang2024dial}          & ECCV24 & 74.5 & 74.9 & 44.4 & ---\\
WeCLIP~\cite{zhang2024frozen}     & CVPR24 & 76.4 & 77.2 & 47.1 & ---\\
\rowcolor{boxgray} WeCLIP++ (Ours)& ---    & 77.3 & 77.8 & 48.1 & 1.36 \up\\
ExCEL~\cite{yang2025exploring}    & CVPR25 & 78.4 & 78.5 & 50.3 & ---\\
\rowcolor{boxgray} ExCEL++ (Ours) & ---    & \textbf{79.2} & \textbf{79.2} & \textbf{50.9} & 1.04 \up\\ \bottomrule
\end{tabular}}\vspace{-3mm}
\end{table}
\noindent\textbf{Semi-supervision.}
Table~\ref{tab:semi} repeats the protocol in the particularly scarce $1/16$ regime, with CoSOD~\cite{chakraborty2024unsupervised} for context. The unlabeled samples are routed by $\rho_i$, which fuses agreement with an EMA teacher and the SAM3 proposal, augmentation consistency, model disagreement, and prediction entropy, so pseudo-mask reliability is assessed from several complementary cues, not from confidence alone. PaceShift again improves both segmenters on every metric despite the severely limited labeled supervision, and again more than doubles the conference gain.

\looseness=-1
\noindent\textbf{Pseudo-label triage.}
To test whether $\rho_i$ captures pseudo-label quality or acts only as a regularizer, we corrupt $20\%$ of the pseudo-labels in the $1/16$ regime with an equal mix of boundary displacement, false-positive patches, and morphological distortion. Each criterion flags its lowest-scoring $20\%$. Ground truth is used only for evaluation. The external SAM3 proposal remains uncorrupted, so this diagnostic assesses corruption detection with an available external comparison signal. Table~\ref{tab:triage} reports flagging precision/recall, which coincide by construction, retained pseudo-label quality, and threshold-free AUROC and AUPRC: First, $\rho_i$ outperforms the confidence, entropy, small-loss, and Dyn-Unc~\cite{he2024large} on both detection and retained quality, so no single temporal or confidence signal explains its advantage. Second, the control rows isolate the external cue. Scoring each pseudo-mask directly by its agreement with the uncorrupted proposal is a strong detector on its own (AUROC $0.806$), stronger than agreement between the model prediction and the proposal ($0.759$), yet $\rho_i$ improves on direct checking on all four measures, by $2.1$ AUROC points (paired bootstrap 95\% CI $1.0$ to $3.2$, Supp.) and $3.4$ flagging points, and $\rho_i$ without any proposal already beats every non-external baseline, so the internal cues add information beyond the proposal and the gain comes from combining the two.

\subsection{Foundation-Model Adaptation}
\label{sec:exp:foundation}
\looseness=-1
Modern dense prediction often adapts frozen foundation backbones using lightweight adapters. We evaluate SAM-, SAM2-, and SAM3-Adapters~\cite{chen2023sam,chen2024sam2,chen2025sam3} on COD under the canonical full-supervision protocol, comparing each baseline with CurriSeg and PaceShift. The prompt-ensembled frozen foundation model serves as teacher and the adapter-tuned model as student. Their disagreement provides reliability evidence, while the teacher mask is never treated as GT. The proposal informs the disagreement cue in $\rho_i$ and the consistency term $Q_{\mathrm{cons}}$ of \cref{eq:guard}, while $Q_{\mathrm{sup}}$ uses available mask supervision. Both phases share the same supervision-evidence construction without relying on teacher pseudo-labels. Table~\ref{tab:foundation} shows that PaceShift improves every adapter and consistently exceeds CurriSeg.

\begin{table}[t]
\setlength{\abovecaptionskip}{0cm}
\centering
\caption{VLM-based open-vocabulary camouflaged object segmentation on OVCamo. The $c \cdot $ measures are the open-vocabulary COS metrics and cIoU is the class-aware IoU. }
\label{tab:ovcos}
\resizebox{\linewidth}{!}{
\setlength{\tabcolsep}{1.1mm}
\begin{tabular}{lccccccc}
\toprule
Methods   & $cS_\alpha$\up   & $cF^w_\beta$\up & $cM$\dn  & $cF_\beta$\up   & $cE_\phi$\up    & cIoU\up  & $\Delta$(\%) \\ \midrule
FC-CLIP~\cite{yu2023convolutions}   & 0.080 & 0.076 & 0.872 & 0.090 & 0.191 & 0.072 & ---\\
OVCoser~\cite{pang2024open}   & 0.579 & 0.490 & 0.336 & 0.520 & 0.616 & 0.443 & ---\\
\rowcolor{boxgray} OVCoser++ (Ours) & 0.593 & 0.526 & 0.312 & 0.556 & 0.651 & 0.466 & 5.78 \up\\
COCUS~\cite{zhao2026open}      & 0.668 & 0.615 & 0.265 & 0.631 & 0.697 & 0.568 & ---\\
\rowcolor{boxgray} COCUS++ (Ours)   & \textbf{0.683} & \textbf{0.647} & \textbf{0.252} & \textbf{0.662} & \textbf{0.725} & \textbf{0.584} & 4.02 \up\\ \bottomrule
\end{tabular}}\vspace{-3mm}
\end{table}
\begin{table}[t]
\setlength{\abovecaptionskip}{0cm}
\centering
\caption{Reference-guided camouflaged object segmentation on R2C7K. }
\label{tab:refcos}
\resizebox{\linewidth}{!}{
\setlength{\tabcolsep}{2.9mm}
\begin{tabular}{@{}lccccc}
\toprule
Methods          & \MAE\dn & $F^w_\beta$\up & \Ep\up & \Sa\up    & $\Delta$(\%) \\ \midrule
SINetV2-Ref~\cite{fan2021concealed,zhang2025referring}      & 0.033 & 0.700 & 0.888 & 0.823 & ---\\
R2CNet~\cite{zhang2025referring}           & 0.036 & 0.669 & 0.879 & 0.805 & ---\\
\rowcolor{boxgray} R2CNet++ (Ours)  & 0.033 & 0.692 & 0.896 & 0.815 & 3.74 \up\\
RefOnce~\cite{wu2025refonce}          & 0.019 & 0.819 & 0.937 & 0.890 & ---\\
\rowcolor{boxgray} RefOnce++ (Ours) & \textbf{0.018} & \textbf{0.832} & \textbf{0.951} & \textbf{0.896} & 2.26 \up\\ \bottomrule
\end{tabular}}\vspace{-4mm}
\end{table}

\subsection{Results on VLM- and Reference-Guided Segmentation}
\label{sec:exp:crossmodal}
A growing family of segmenters uses vision-language or reference information. We apply PaceShift to three such settings for controlled comparison, each with the original baseline and a stronger recent method, preserving the architecture, input protocol, optimizer, training horizon, and trainable parameter count while changing only data scheduling and post-training. Each setting is evaluated on its native benchmark with the corresponding official metrics. The unreliability PaceShift addresses is instantiated per case. The conference recipe is compared in the CECS and incomplete-supervision experiments, and this section reports PaceShift against the host baselines.

\noindent\textbf{VLM-based weakly supervised segmentation.}
CLIP-based segmenters derive supervision from CAM pseudo-masks whose quality varies widely across images and classes. Since no dense GT exists at training time, the difficulty stays $d_i\!=\!1\!-\!\rho_i$, with class-balanced CAM agreement serving as the supervision signal $s_i^{\mathrm{sup}}$, vision-language evidence entering through the consistency and disagreement cues of Sec.~\ref{sec:sarcs}, and normalized multi-class entropy completing $\rho_i$, so pseudo-mask reliability is assessed from several decoupled cues instead of CAM confidence alone. Table~\ref{tab:vlm-ws} plugs PaceShift into WeCLIP~\cite{zhang2024frozen} and ExCEL~\cite{yang2025exploring} on PASCAL VOC~\cite{everingham2010pascal} and MS COCO~\cite{lin2014microsoft}, with DIAL~\cite{jang2024dial} for context, improving both models on every evaluated split.

\noindent\textbf{VLM-based open-vocabulary camouflaged segmentation.}
Open-vocabulary segmentation requires both localizing camouflaged objects and recognizing potentially unseen classes. We incorporate vision-language agreement into the existing disagreement and consistency cues, while $Q_{\mathrm{sup}}$ uses available mask supervision for each candidate. Table~\ref{tab:ovcos} shows that PaceShift consistently improves OVCoser~\cite{pang2024open} and COCUS~\cite{zhao2026open}, including class-aware IoU. The generic FC-CLIP~\cite{yu2023convolutions} performs worse on OVCamo, highlighting the difficulty of this task for general-purpose segmenters.

\noindent\textbf{Reference-guided camouflaged segmentation.}
Reference-target agreement provides condition-specific reliability evidence during training, computed in an independent feature space so it does not merely echo the segmenter's own matching branch, and enters through the same cues. Table~\ref{tab:refcos} plugs PaceShift into R2CNet~\cite{zhang2025referring}, which introduced the R2C7K benchmark, and the stronger RefOnce~\cite{wu2025refonce}, which distills references into a prototype memory and needs none at inference. Both improve consistently on every metric.

\subsection{Ablation Study and Hyperparameter Sensitivity}
\label{sec:exp:abl}
Unless otherwise stated, ablations use RUN with PVT~V2 on COD10K, and supervision-dependent ablations switch the regime, not the dataset. Single-run Phase-II controls share one checkpoint $\theta_0$, and the five-run Phase-II comparison uses one Phase-I endpoint per run.

\noindent\textbf{Curriculum components and their complementarity.} Table~\ref{tab:factorial} runs the complete factorial over the three Phase-I mechanisms without post-training, so the phase is isolated. Selection is the strongest standalone mechanism and temporal weighting adds a smaller, consistent gain, and pixel weighting is weak alone but improves all four metrics on top of RCS and TSSW, acting as complementary refinement.

\begin{table}[t]
\setlength{\abovecaptionskip}{2pt}
\centering
\caption{Full factorial over Phase-I without post-training. RCS: robust curriculum selection, the percentile selection of \cref{eq:select}, TSSW: temporal statistics-based sample weighting, PUE: pixel uncertainty weighting. The all-on configuration corresponds to SA-RCS.}
\label{tab:factorial}
\resizebox{\linewidth}{!}{
\setlength{\tabcolsep}{3.2mm}
\begin{tabular}{ccc cccc}
\toprule
RCS & TSSW & PUE & $M$\dn & \Fb\up & \Ep\up & \Sa\up\\
\midrule
--- & --- & --- & 0.021 & 0.810 & 0.941 & 0.878\\
\cmark & --- & --- & 0.020 & 0.817 & 0.946 & 0.879\\
--- & \cmark & --- & 0.021 & 0.814 & 0.943 & 0.879\\
--- & --- & \cmark & 0.021 & 0.813 & 0.942 & 0.878\\
\cmark & \cmark & --- & 0.020 & 0.821 & 0.949 & 0.880\\
\cmark & --- & \cmark & 0.020 & 0.819 & 0.947 & 0.879\\
--- & \cmark & \cmark & 0.020 & 0.816 & 0.945 & 0.879\\
\rowcolor{boxgray}
\cmark & \cmark & \cmark & \best{0.019} & \best{0.824} & \best{0.952} & \best{0.881}\\
\bottomrule
\end{tabular}}\vspace{-5mm}
\end{table}

\looseness=-1
\noindent\textbf{Curriculum ordering.}
The factorial shows that selection helps. Table~\ref{tab:ordering} asks what it should be based on. Every row keeps the same progressive schedule $p(t)$, subset sizes, optimizer-step counts, and weighting. All dynamic rankings are refreshed at the same cadence, Static $\rho_i$ is computed once after warm-up and frozen, and Random fixes one random order. The random-order row remains close to the no-selection configuration of Table~\ref{tab:factorial}, with only small gains in \Fb\ and \Ep, so the progressive schedule alone does not explain the full improvement.
Loss- and confidence-based rankings give inconsistent and limited gains over this control, 
in line with the weak link between instantaneous loss and useful updates in Sec.~\ref{sec:exp:mech},
whereas reliability-based rankings improve all four metrics. The online score keeps a modest, consistent margin over the RCS ranking, and freezing $\rho_i$ after warm-up removes that margin and becomes comparable to RCS, suggesting that online tracking of the learning dynamics contributes to the improvement.

\noindent\textbf{Inside the supervision-adaptive curriculum.}
Table~\ref{tab:sarcsparts}(a) isolates the reliability cues of $\rho_i$, the main Phase-I change from CurriSeg. Supervision evidence alone performs substantially below the full reliability score, and removing model disagreement causes the largest drop among the drop-one variants. These results support combining complementary cues instead of relying on supervision agreement alone, and the external proposal is isolated in Table~\ref{tab:triage}. In~(b) and~(c), the parameter-free mid-variance weight and the pixel weight with floor and decay match or exceed every compared variant on all metrics.

\begin{figure}[t]
\centering
\includegraphics[width=0.95\linewidth]{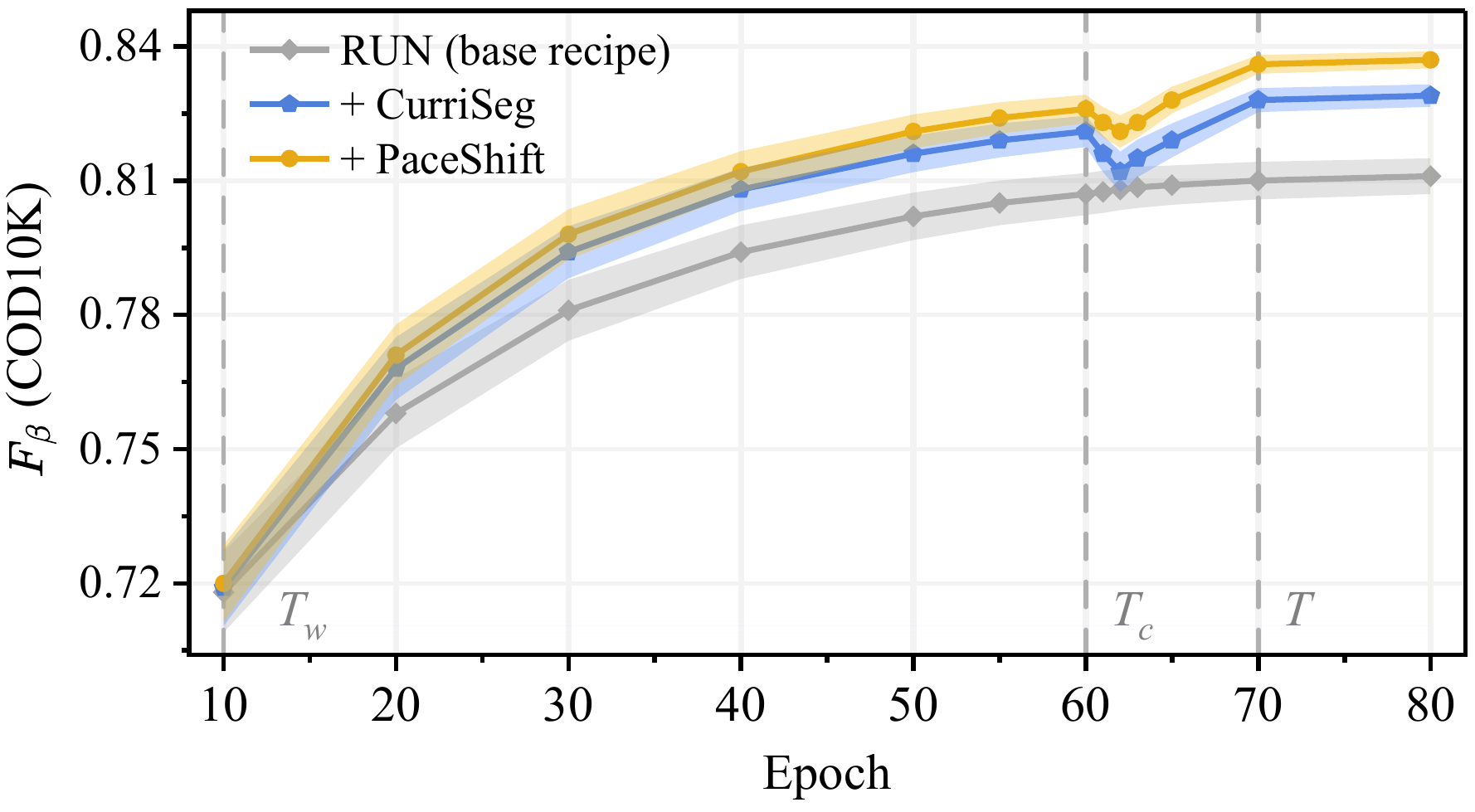}
\vspace{-3mm}
\caption{Training dynamics on COD10K (RUN with PVT~V2): \Fb\ on the test set at each checkpoint, mean over five runs with one-standard-deviation bands. The marks on the epoch axis indicate $T_w$, $T_c$, and $T$, and ten extra diagnostic epochs beyond $T$ are shown for reference.}\vspace{-3mm}
\label{fig:dynamics}
\end{figure}

\begin{figure}[t]
\centering
\includegraphics[width=0.95\linewidth]{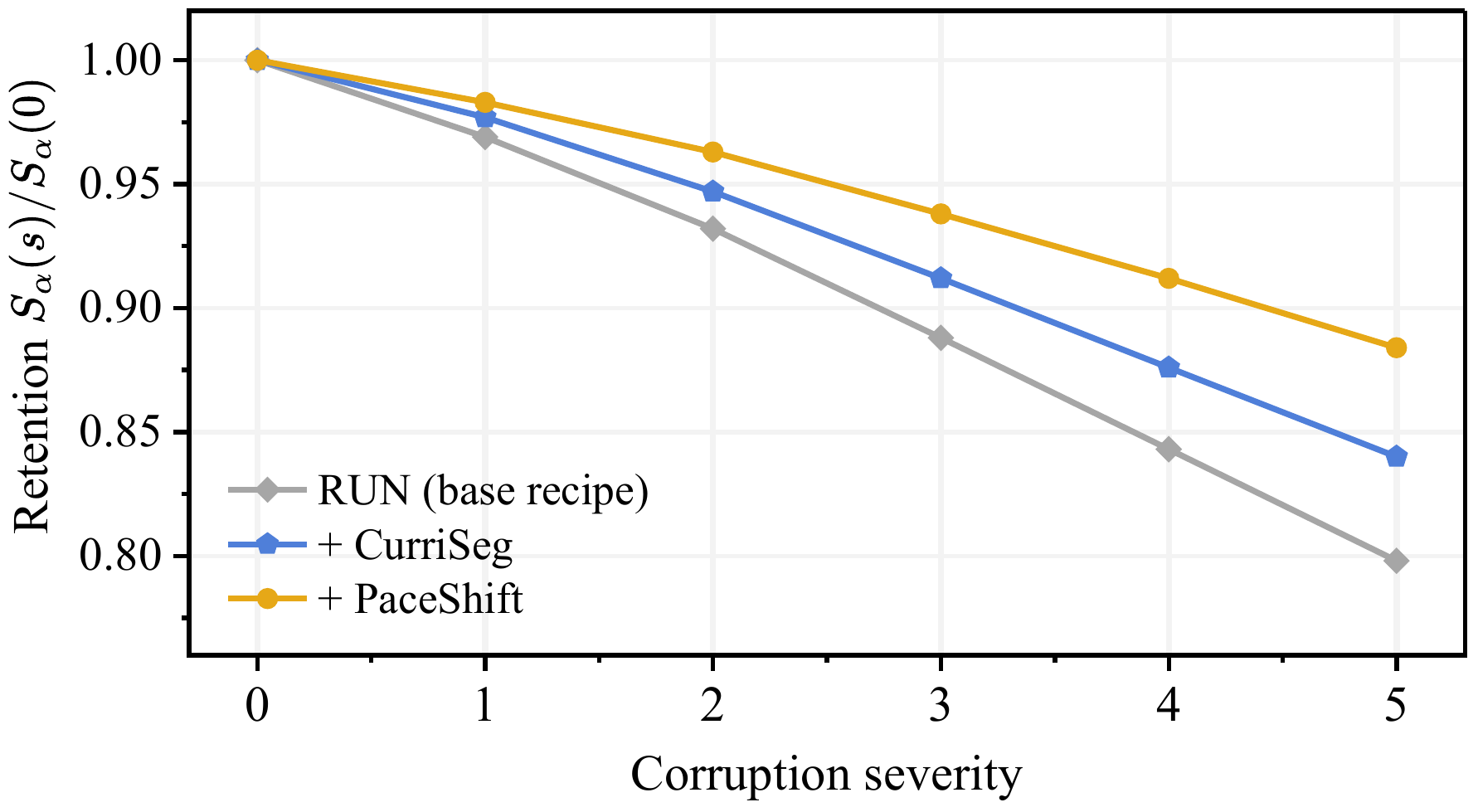}
\vspace{-3.5mm}
\caption{Robustness on COD10K as retention of each recipe's own clean $S_\alpha$. Severity $0$ is the clean test set. At each severity, retention is averaged within each corruption family and then equally over the four families.}\vspace{-5mm}
\label{fig:robust}
\end{figure}

\looseness=-1
\noindent\textbf{The guarded anti-curriculum.}
Beyond the four accuracy metrics, Table~\ref{tab:postabl} reports pixel-pooled ECE, the ground-truth-stratified confidence--accuracy gap ($\cag$), and \Sa$^{c}$,
the absolute \Sa\ averaged over the corruption suite of Fig.~\ref{fig:robust}, first within each family and then across the four families, over the five nonzero severities. The upper part uses the five runs of Fig.~\ref{fig:dynamics}. Every post-training variant starts from the Phase-I endpoint of its run with the same budget, and the rows compare alternative designs, not a cumulative path. The lower part reports single-run controls from the canonical run. 
Multi-radius supervision improves over the single-radius conference operator but stays below the full design. Guarded $m^{+}$ distillation, which differs from full SBFT-P only in the clean-view term, recovers most of its clean-accuracy improvement, with small differences in the five-run mean accuracy scores. Replacing distillation with the reference-relative pairwise objective improves \Sa$^{c}$ by $0.0058\pm0.0018$ and reduces ECE by $0.0010\pm0.0003$ and $\cag$ by $0.0027\pm0.0009$ in all five runs (mean paired difference $\pm$ $95\%$ CI half-width, Holm-adjusted paired $t$-test $p<0.01$).
In the canonical run, random preferences stay at the level of multi-radius supervision, confidence-ranked preferences recover some accuracy and robustness but widen $\cag$, and removing the anchor costs a margin similar to that of the pairwise objective. The guard drop-one ablations (Supp.) largely align with the intended roles: removing $Q_{\mathrm{rob}}$ mainly costs corruption robustness, $Q_{\mathrm{unc}}$ the confidence gap, and $Q_{\mathrm{sup}}$ clean accuracy. Swapping the frozen SA-RCS reference in $Q_{\mathrm{cons}}$ for the current prediction costs substantially more corruption robustness than clean accuracy, supporting a fixed consistency reference over self-agreement. The final row is a direct confidence-reward negative control (defined in the Supp.) that leaves candidate selection unchanged and instead minimizes prediction entropy during post-training. Combined with the confidence-preference row, it separates confidence-based candidate selection from directly optimizing confidence, and Table~\ref{tab:calib} reports its calibration effect.

\begin{table}[t]
\setlength{\abovecaptionskip}{2pt}
\centering
\caption{Curriculum ordering under a matched budget on COD10K.}
\label{tab:ordering}
\resizebox{\linewidth}{!}{
\setlength{\tabcolsep}{4mm}
\begin{tabular}{@{}l cccc}
\toprule
Ranking criteria & $M$\dn & \Fb\up & \Ep\up & \Sa\up \\
\midrule
Random               & 0.021 & 0.814 & 0.944 & 0.878 \\
Loss easy-first      & 0.021 & 0.812 & 0.943 & 0.877 \\
Loss hard-first      & 0.020 & 0.816 & 0.945 & 0.878 \\
Confidence easy-first& 0.021 & 0.813 & 0.943 & 0.877 \\
RCS ranking (CurriSeg) & 0.020 & 0.820 & 0.949 & 0.880 \\
Static $\rho_i$      & 0.020 & 0.818 & 0.947 & 0.879 \\
\rowcolor{boxgray}
Online SA-RCS        & \best{0.019} & \best{0.824} & \best{0.952} & \best{0.881} \\
\bottomrule
\end{tabular}}\vspace{-5mm}
\end{table}

\begin{table*}[t]
\setlength{\abovecaptionskip}{1pt}
\centering
\caption{Inside the supervision-adaptive curriculum. Each Phase-I variant is inserted into the otherwise unchanged full PaceShift pipeline, including Phase~II, so every panel reports final performance. 
(a)~Reliability cues under $1/16$ semi-supervision (SCALER, SAM3).
(b)~Temporal weighting shape under full supervision. 
(c)~Pixel uncertainty weighting under full supervision.
}
\label{tab:sarcsparts}
\resizebox{\linewidth}{!}{%
\setlength{\tabcolsep}{1mm}%
\begin{tabular}{@{}c@{\hspace{6mm}}c@{\hspace{6mm}}c@{}}
\smash{(a) Reliability cues} & \smash{(b) Temporal weighting} & \smash{(c) Pixel weighting}\\ 
\begin{tabular}[t]{l cccc}
\toprule
$\rho_i$ variants & $M$\dn & \Fb\up & \Ep\up & \Sa\up\\
\midrule
Supervision only & 0.035 & 0.730 & 0.848 & 0.788\\
w/o supervision  & 0.033 & 0.740 & 0.855 & 0.792\\
w/o consistency  & 0.032 & 0.742 & 0.857 & 0.793\\
w/o disagreement & 0.034 & 0.735 & 0.851 & 0.789\\
w/o entropy      & 0.032 & 0.744 & 0.859 & 0.795\\
\rowcolor{boxgray}
Full $\rho_i$    & \best{0.031} & \best{0.748} & \best{0.863} & \best{0.797}\\
\bottomrule
\end{tabular}
&
\begin{tabular}[t]{l cccc}
\toprule
$\omega_i$ variants & $M$\dn & \Fb\up & \Ep\up & \Sa\up\\
\midrule
CurriSeg TSSW        & 0.018 & 0.832 & 0.959 & 0.887\\
Mono.\ incr.\ $\sigma$ & 0.018 & 0.829 & 0.957 & 0.885\\
Mono.\ decr.\ $\sigma$ & 0.018 & 0.831 & 0.958 & 0.886\\
w/o $\omega^{\mathrm{out}}$ & \best{0.017} & 0.833 & 0.960 & 0.887\\
Unimodal mean        & \best{0.017} & 0.834 & 0.961 & 0.888\\
\rowcolor{boxgray}
Param.-free full     & \best{0.017} & \best{0.836} & \best{0.963} & \best{0.889}\\
\bottomrule
\end{tabular}
&
\begin{tabular}[t]{l cccc}
\toprule
PUE variants & $M$\dn & \Fb\up & \Ep\up & \Sa\up\\
\midrule
No PUE       & 0.018 & 0.830 & 0.958 & 0.886\\
Hard mask    & 0.018 & 0.829 & 0.956 & 0.884\\
Soft only    & 0.018 & 0.830 & 0.957 & 0.885\\
\,$+$\,decay only & \best{0.017} & 0.832 & 0.959 & 0.886\\
\,$+$\,floor only & \best{0.017} & 0.834 & 0.961 & 0.888\\
\rowcolor{boxgray}
\,$+$\,floor $+$ decay & \best{0.017} & \best{0.836} & \best{0.963} & \best{0.889}\\
\bottomrule
\end{tabular}
\end{tabular}}
\vspace{-4mm}
\end{table*}

\begin{table}[t]
\setlength{\abovecaptionskip}{2pt}
\centering
\caption{Phase-II designs on COD10K (RUN, PVT~V2). Upper part: mean over the five runs of Fig.~\ref{fig:dynamics}, with the sample standard deviation in the row beneath. Within each run, all post-training variants share the Phase-I endpoint and the optimization-step budget, and SA-RCS only evaluates that endpoint without post-training. Lower part: single-run controls from the canonical run (run~1). ECE is pixel-pooled, $\cag$ is a confidence diagnostic, and \Sa$^{c}$ is the corruption-averaged \Sa. Guard drop-one ablations are in the Supp.}
\label{tab:postabl}
\resizebox{\linewidth}{!}{
\setlength{\tabcolsep}{0.7mm}
\begin{tabular}{l ccccc cc}
\toprule
Variants & $M$\dn & \Fb\up & \Ep\up & \Sa\up & \Sa$^{c}$\up & ECE\dn & $\cag$\dn\\
\midrule
\multicolumn{8}{l}{\textit{Five runs: mean (first row), std (second row)}}\\
SA-RCS only (no post)             & 0.019 & 0.823 & 0.951 & 0.880 & 0.788 & 0.015 & 0.048\\[-2pt]
 & \sdv{.0003} & \sdv{.002} & \sdv{.002} & \sdv{.001} & \sdv{.003} & \sdv{.001} & \sdv{.001}\\
Single-$r$ SBFT (conf.)           & 0.018 & 0.827 & 0.955 & 0.883 & 0.802 & 0.014 & 0.044\\[-2pt]
 & \sdv{.0003} & \sdv{.002} & \sdv{.002} & \sdv{.001} & \sdv{.003} & \sdv{.001} & \sdv{.001}\\
Multi-$r$ supervised SBFT         & 0.018 & 0.829 & 0.957 & 0.884 & 0.811 & 0.013 & 0.040\\[-2pt]
 & \sdv{.0003} & \sdv{.002} & \sdv{.001} & \sdv{.001} & \sdv{.003} & \sdv{.001} & \sdv{.001}\\
SBFT $+$ guarded $m^{+}$ distill. & \best{0.017} & \best{0.835} & 0.961 & \best{0.888} & 0.825 & 0.012 & 0.038\\[-2pt]
 & \sdv{.0003} & \sdv{.003} & \sdv{.002} & \sdv{.001} & \sdv{.002} & \sdv{.001} & \sdv{.001}\\
\rowcolor{boxgray}
SBFT-P (full)                     & \best{0.017} & \best{0.835} & \best{0.962} & \best{0.888} & \best{0.831} & \best{0.011} & \best{0.036}\\[-2pt]
\rowcolor{boxgray}
 & \sdv{.0002} & \sdv{.002} & \sdv{.001} & \sdv{.001} & \sdv{.003} & \sdv{.001} & \sdv{.001}\\
\midrule
\multicolumn{8}{l}{\textit{Single run (run~1)}}\\
SBFT-P (full, run~1)              & 0.017 & 0.836 & 0.963 & 0.889 & 0.832 & 0.011 & 0.035\\
SBFT $+$ random pref.             & 0.018 & 0.829 & 0.957 & 0.884 & 0.809 & 0.013 & 0.041\\
SBFT $+$ confidence pref.         & 0.018 & 0.831 & 0.959 & 0.886 & 0.815 & 0.015 & 0.044\\
SBFT $+$ guarded, w/o anchor      & 0.017 & 0.833 & 0.960 & 0.887 & 0.824 & 0.012 & 0.039\\
Direct conf.\ reward              & 0.018 & 0.831 & 0.958 & 0.886 & 0.806 & 0.024 & 0.071\\
\bottomrule
\end{tabular}}\vspace{-4mm}
\end{table}

\begin{table}[t]
\setlength{\abovecaptionskip}{2pt}
\centering
\caption{Phase order on COD10K (single run). All schedules share the same warm-up, and for challenge-first variants the frozen warm-up checkpoint serves as the reference. 
}
\label{tab:phaseorder}
\resizebox{\linewidth}{!}{
\setlength{\tabcolsep}{0.6mm}
\begin{tabular}{l cccc cc}
\toprule
Training order & $M$\dn & \Fb\up & \Ep\up & \Sa\up & \Sa$^{c}$\up & $\cag$\dn\\
\midrule
Standard training              & 0.021 & 0.810 & 0.941 & 0.878 & 0.778 & 0.056\\
SBFT after warm-up             & 0.023 & 0.785 & 0.924 & 0.858 & 0.744 & 0.062\\
SBFT-P after warm-up           & 0.022 & 0.792 & 0.930 & 0.865 & 0.752 & 0.059\\
SBFT-P $\rightarrow$ SA-RCS    & 0.020 & 0.812 & 0.943 & 0.876 & 0.771 & 0.052\\
SA-RCS $\rightarrow$ SBFT (conf.) & 0.018 & 0.828 & 0.956 & 0.884 & 0.803 & 0.043\\
\rowcolor{boxgray}
SA-RCS $\rightarrow$ SBFT-P (ours) & \best{0.017} & \best{0.836} & \best{0.963} & \best{0.889} & \best{0.832} & \best{0.035}\\
Alternating SA-RCS / SBFT-P    & 0.018 & 0.830 & 0.957 & 0.885 & 0.816 & 0.039\\
Adaptive $T_c$ trigger         & \best{0.017} & \best{0.836} & 0.962 & 0.888 & 0.828 & 0.036\\
\bottomrule
\end{tabular}}\vspace{-5mm}
\end{table}

\begin{table}[t]
\setlength{\abovecaptionskip}{2pt}
\centering
\caption{Trust parameterization ($1/16$ semi-supervision, plus point supervision for the sparse-human preset). Uniform: equal cue coefficients, single $\trust$: one shared trust for all sources, ``Mismatched'': supervision receives an incorrect source-type preset (in semi the dense-human and model-derived presets are swapped, and in point the sparse-human source receives the model-derived preset). The $\pm20\%$ row reports mean$\pm$std over ten order-preserving random draws. Variation is across configurations, not seeds. Per-regime tuning selects $\trust$ on held-out training data only.}
\label{tab:trust}
\resizebox{\linewidth}{!}{
\setlength{\tabcolsep}{0.4mm}
\begin{tabular}{l cccc c}
\toprule
Trust design & $M$\dn & \Fb\up & \Ep\up & \Sa\up & \Sa\ (point)\up\\
\midrule
Uniform cue & 0.034 & 0.731 & 0.848 & 0.786 & 0.820\\
Single $\trust{=}0.5$    & 0.033 & 0.738 & 0.854 & 0.791 & 0.826\\
Mismatched trust & 0.036 & 0.719 & 0.838 & 0.777 & 0.812\\
Preset $\pm20\%$    & 0.032{\tiny$\pm$0.001} & 0.745{\tiny$\pm$0.002} & 0.861{\tiny$\pm$0.002} & 0.796{\tiny$\pm$0.001} & 0.831{\tiny$\pm$0.001}\\
\rowcolor{boxgray}
Source-type preset  & \best{0.031} & 0.748 & 0.863 & 0.797 & 0.834\\
Val.-tuned $\trust$ & \best{0.031} & \best{0.749} & \best{0.864} & \best{0.798} & \best{0.835}\\
\bottomrule
\end{tabular}}\vspace{-4mm}
\end{table}

\noindent\textbf{Phase ordering.}
Table~\ref{tab:phaseorder} examines whether the two phases are interchangeable. Challenge-first schedules, guarded or not, fall substantially below standard training, consistent with an early spectral challenge interfering with representation formation. Reversing the phases recovers much of the damage but stays below standard training in \Sa\ and \Sa$^{c}$, so later stabilization does not recover the loss from an early challenge in this setting. After stabilization, the conference operator realizes part of the gain and SBFT-P the rest, largest in \Sa$^{c}$ and $\cag$, the properties Phase~II targets. Alternating the two losses recovers part of the benefit but stays below the strict two-stage order, again with the clearest gaps in \Sa$^{c}$ and $\cag$. 
An adaptive trigger that starts Phase~II at the validation plateau performs on par with the fixed $T_c$, so the result is not sensitive to the transition epoch. 
Fig.~\ref{fig:dynamics} complements these end points with the training dynamics of all three recipes.

\noindent\textbf{Trust parameterization.}
The source-type presets are the only regime-level constants of the method, so Table~\ref{tab:trust} stresses them in two regimes that together cover all three supervision sources: semi-supervised $1/16$ (Table~\ref{tab:semi}), where dense human masks coexist with model-derived pseudo-labels, and point supervision (Table~\ref{tab:weak}), which exercises the sparse-human preset. Uniform cue weights and a source-agnostic $\trust{=}0.5$ both reduce performance, and mismatched source trust degrades the most, supporting source-dependent trust over a single scalar. The presets are also insensitive to their exact values: ten order-preserving $\pm20\%$ perturbations stay close to the default, and validation tuning improves \Sa\ by only $0.001$.

\noindent\textbf{Global constants.}
One-at-a-time sweeps under the fixed budget $T{=}70$ leave performance stable over broad ranges, with the largest changes in $\cag$ in the two Phase-II safety scales: 
$\varepsilon{=}0$ raises $\cag$ to $0.041$ and $\lambda_{\mathrm{anc}}{=}0$ raises it to $0.039$, reproducing the no-anchor configuration of Table~\ref{tab:postabl}.
Joint perturbation of the five non-transition constants keeps mean \Sa\ within $0.002$ of the default at $\pm20\%$ and produces no abrupt collapse at $\pm40\%$, and full sweeps are in the Supp.

\noindent\textbf{Training dynamics.}
Fig.~\ref{fig:dynamics} shows \Fb\ on the COD10K test set across training checkpoints, averaged over five runs with one-standard-deviation bands. All schedules, hyperparameters, and reported checkpoints are fixed before these retrospective evaluations, so no test result enters selection or tuning. The three trajectories are indistinguishable through the warm-up. Phase~I then separates them twice over: both curricula pull away from standard training, and PaceShift keeps a modest, consistent margin over CurriSeg. After $T_c$ the challenge briefly costs both recipes accuracy, but PaceShift drops less, recovers earlier, and ends higher with the narrowest band. Ten extra epochs change every curve by at most $0.001$, so the $T{=}70$ checkpoints are not cut short.


\subsection{Robustness and Calibration}
\label{sec:exp:robust}
This section reports how the trained model behaves when the input degrades and whether its confidence tracks its correctness. The mechanism behind both is analyzed in~\cref{sec:exp:mech}.

\noindent\textbf{Robustness under diverse corruptions.}
Complementing the controlled single-band interventions of Sec.~\ref{sec:exp:mech}, we evaluate robustness under a broad suite of common image corruptions, grouped into four families: noise (Gaussian, shot, impulse), blur (defocus, motion), weather and visibility (fog, frost), and digital or photometric (contrast, brightness, JPEG), each at five severities. 
Because the three recipes differ in clean accuracy, Fig.~\ref{fig:robust} reports the retention $S_\alpha(s)/S_\alpha(0)$, so that every curve starts at one and the separation reflects robustness relative to each model's own clean performance. At each severity, retention is averaged first over corruptions within each family and then equally over the four families. Both recipes degrade more gracefully than the base schedule, and PaceShift retains more than CurriSeg at every nonzero severity, with the margin widening as corruption strengthens.

\begin{table}[t]
\setlength{\abovecaptionskip}{0cm}
\centering
\caption{Calibration and confidence diagnostics. ECE: pixel-pooled expected calibration error, used to assess calibration. $\cag$: ground-truth-stratified confidence--accuracy gap, the mean of the gaps on ground-truth foreground and background pixels, $\cagb$: the same gap within a dilated boundary band, and WConf: mean confidence on incorrect pixels, all three being complementary diagnostics. TS~\cite{guo2017calibration}: post-hoc temperature scaling, a single scalar fitted on held-out validation data. (a)~Recipe-level comparison on COD10K (single run), where the final column is the direct confidence-reward control defined in the Supp. (b)~The guarded stage across tasks and supervision regimes (RUN with PVT~V2 for PIS and TOD, SCALER for the COD regimes).}
\label{tab:calib}
\resizebox{\linewidth}{!}{
\setlength{\tabcolsep}{3.2mm}
\begin{tabular}{@{}l ccccc}
\multicolumn{6}{c}{\footnotesize (a) Recipe-level comparison on COD10K}\\
\toprule
Metrics & RUN & $+$TS & $+$CurriSeg & \cellcolor{boxgray}$+$PaceShift & Direct conf.\ reward\\
\midrule
ECE $\downarrow$     & 0.019 & \best{0.010} & 0.016 & \cellcolor{boxgray}0.011 & 0.024\\
\midrule
$\cag$ $\downarrow$  & 0.056 & \best{0.034} & 0.046 & \cellcolor{boxgray}0.035 & 0.071\\
$\cagb$ $\downarrow$ & 0.121 & 0.102 & 0.096 & \cellcolor{boxgray}\best{0.074} & 0.148\\
WConf $\downarrow$   & 0.86  & 0.80  & 0.82  & \cellcolor{boxgray}\best{0.79}  & 0.94\\
\bottomrule
\end{tabular}}

\vspace{5pt}
\resizebox{\linewidth}{!}{
\setlength{\tabcolsep}{0.5mm}
\begin{tabular}{l cccc cccc}
\multicolumn{9}{c}{\footnotesize (b) Generality across tasks and regimes}\\
\toprule
\multirow{2}{*}{Tasks / regimes} & \multicolumn{4}{c}{SA-RCS only} & \multicolumn{4}{c}{\,$+$\,SBFT-P}\\
\cmidrule(lr){2-5}\cmidrule(lr){6-9}
 & ECE & $\cag$ & $\cagb$ & WConf & ECE & $\cag$ & $\cagb$ & WConf\\
\midrule
PIS              & 0.018 & 0.042 & 0.098 & 0.80 & \textbf{0.014} & \best{0.032} & \best{0.072} & \best{0.77}\\
TOD              & 0.018 & 0.038 & 0.089 & 0.78 & \textbf{0.013} & \best{0.030} & \best{0.067} & \best{0.75}\\
COD, point       & 0.019 & 0.054 & 0.127 & 0.86 & \textbf{0.015} & \best{0.040} & \best{0.091} & \best{0.81}\\
COD, semi $1/16$ & 0.017 & 0.051 & 0.119 & 0.85 & \textbf{0.014} & \best{0.037} & \best{0.085} & \best{0.80}\\
\bottomrule
\end{tabular}}\vspace{-4mm}
\end{table}

\noindent\textbf{Calibration.}
Table~\ref{tab:calib} reports pixel-pooled ECE for calibration, alongside three complementary confidence diagnostics. In panel~(a), PaceShift reduces ECE from $0.019$ to $0.011$ on COD10K, below CurriSeg ($0.016$), and improves all three diagnostics. Post-hoc temperature scaling~\cite{guo2017calibration}, which needs a separate labeled calibration split, reaches a slightly lower ECE and $\cag$, whereas PaceShift obtains its calibration from training alone and gives a smaller boundary gap and lower confidence on incorrect pixels. Direct confidence reward retains competitive clean accuracy (Table~\ref{tab:postabl}) but worsens ECE and the confidence diagnostics. Panel~(b) shows consistent improvements across various tasks and supervision regimes.

\begin{table}[t]
\setlength{\abovecaptionskip}{1pt}
\centering
\caption{Validation of PaceShift's internal training signals. (a)~Pseudo-label quality across $\rho_i$ quintiles on the unlabeled split of the semi-supervised $1/16$ regime. Ground truth is withheld from training and used only post hoc, and $r_s{=}0.64$ is computed over individual pseudo-labels before binning. (b)~Alignment of sample scores on $\mathcal{D}_s$ with the update-direction proxy $A^{\mathrm{opt}}$: Spearman and top-$20\%$ overlap (random: $20\%$), reported as mean$\pm$std over the four Phase-I refresh checkpoints with at least two history entries. Variation is across checkpoints, not training seeds.}
\label{tab:mechscores}
\resizebox{\linewidth}{!}{%
\setlength{\tabcolsep}{1.6mm}%
\begin{tabular}{@{}c@{\hspace{2mm}}c@{}}
\smash{(a) $\rho_i$ vs.\ pseudo-label quality} & \smash{(b) Alignment with $A^{\mathrm{opt}}$}\\
\begin{tabular}[t]{l ccc}
\toprule
$\rho_i$ quin. & IoU\up & bnd.\ F\up & pix.\ err.\dn\\
\midrule
Q5 (high) & \best{0.871} & \best{0.822} & \best{0.032}\\
Q4           & 0.804 & 0.741 & 0.058\\
Q3           & 0.742 & 0.668 & 0.083\\
Q2           & 0.668 & 0.585 & 0.121\\
Q1 (low)  & 0.549 & 0.468 & 0.192\\
\bottomrule
\end{tabular}
&
\begin{tabular}[t]{l cc}
\toprule
Scores & Spearman\up & Top-$20\%$\up\\
\midrule
Instant.\ loss & $-$0.25{\tiny$\pm$0.02} & 11{\tiny$\pm$1}\%\\
GraNd          & 0.20{\tiny$\pm$0.01} & 23{\tiny$\pm$1}\%\\
Forgetting     & 0.29{\tiny$\pm$0.01} & 25{\tiny$\pm$1}\%\\
$\rho_i$       & 0.32{\tiny$\pm$0.01} & 28{\tiny$\pm$1}\%\\
\rowcolor{boxgray}
$\omega_i$     & \best{0.53{\tiny$\pm$0.02}} & \best{41{\tiny$\pm$2}\%}\\
\bottomrule
\end{tabular}
\end{tabular}}
\vspace{-4mm}
\end{table}

\begin{table}[t]
\setlength{\abovecaptionskip}{1pt}
\centering
\caption{Guard selection against held-out GT, shown only where $Q_{\mathrm{sup}}$ is surrogate-based. Top-1 is the rate of selecting a candidate within $10^{-3}$ true IoU of the oracle candidate, and Regret is the true IoU left on the table. Random reports the exact expectation under uniform selection among the three candidates, including tolerance-induced oracle ties.}
\label{tab:guardsel}
\resizebox{\linewidth}{!}{%
\setlength{\tabcolsep}{3.2mm}%
\begin{tabular}{@{}l cc cc}
\toprule
\multirow{2}{*}{Selection rules} & \multicolumn{2}{c}{Semi $1/16$ (unlab.)} & \multicolumn{2}{c}{Point} \\
\cmidrule(lr){2-3}\cmidrule(lr){4-5}
 & Top-1 (\%)\up & Regret\dn & Top-1 (\%)\up & Regret\dn \\
\midrule
Random (expected)                        & 36.7 & 0.052 & 34.9 & 0.061 \\
Confidence only              & 47.1 & 0.038 & 44.8 & 0.046 \\
Consistency only             & 54.6 & 0.031 & 52.9 & 0.038 \\
Guard w/o $Q_{\mathrm{sup}}$ & 58.3 & 0.027 & 57.1 & 0.033 \\
\rowcolor{boxgray}
Full guard $Q(m)$            & \best{71.8} & \best{0.014} & \best{68.4} & \best{0.019} \\
\bottomrule
\end{tabular}}
\vspace{-5mm}
\end{table}
\begin{table}[t]
\setlength{\abovecaptionskip}{1pt}
\centering
\caption{Test-time frequency intervention on COD10K (RUN, PVT~V2). Each cell is the drop from that model's own clean \Sa. The naive easy-first row is a conventional easy-to-hard schedule, unlike the matched-budget control of Table~\ref{tab:ordering}.}
\label{tab:freqint}
\resizebox{\linewidth}{!}{%
\setlength{\tabcolsep}{1.4mm}%
\begin{tabular}{@{}l c ccc}
\toprule
\multirow{2}{*}{Training} & \multirow{2}{*}{Clean\up} & \multicolumn{3}{c}{drop under intervention}\\
\cmidrule(lr){3-5}
 & & HF rem.\dn & HF noise\dn & LF supp.\ (DC retained)\\
\midrule
Standard            & 0.878 & 0.086 & 0.177 & 0.276\\
Naive easy-first CL & 0.871 & 0.104 & 0.201 & 0.249\\
SA-RCS only         & 0.881 & 0.071 & 0.142 & 0.288\\
\rowcolor{boxgray}
PaceShift        & \best{0.889} & \best{0.049} & \best{0.098} & 0.309\\
\bottomrule
\end{tabular}}
\vspace{-5mm}
\end{table}

\subsection{Mechanism Analysis}
\label{sec:exp:mech}
We next assess whether $\rho_i$ tracks supervision quality, $\omega_i$ reflects update value, $Q(m)$ ranks better candidates, and spectral post-training reduces high-frequency reliance.

\subsubsection{Does reliability track supervision quality}
In the triage of Table~\ref{tab:triage}, the full $\rho_i$ gives the best corruption detection and retained quality. Removing the external proposal lowers AUROC by $3.5$ points, and direct pseudo-mask--proposal agreement reaches $0.806$, so $\rho_i$ adds $2.1$ points over direct external checking and the gain comes from fusing external and internal evidence, not from either alone. On naturally occurring pseudo-labels (Table~\ref{tab:mechscores}(a)), quality decreases monotonically across the five $\rho_i$ quintiles, with a $32.2$-point top-to-bottom IoU gap and a sample-level Spearman correlation of $0.64$ with true IoU. Thus, in the pseudo-label regime, $\rho_i$ ranks true supervision quality beyond model confidence.

\subsubsection{Does the temporal weight align with useful updates}
On a fixed diagnostic subset $\mathcal{D}_s$, we compare sample scores against an optimizer-conditioned first-order alignment proxy $A_i^{\mathrm{opt}}$, measured against a $200$-image probe set $\mathcal{D}_p$ held out from optimization and tuning. The proxy uses the unweighted loss and a frozen Adam preconditioner, so no compared weighting enters its own target. Correlations are computed independently over the four Phase-I refresh checkpoints with at least two history entries and then averaged. Full protocol details are in the Supp. Table~\ref{tab:mechscores}(b) shows that instantaneous loss is negatively correlated with update alignment, whereas $\rho_i$ is moderately positive and $\omega_i$ achieves the strongest correlation and top-$20\%$ overlap. The two therefore answer different questions: $\rho_i$ asks whether supervision can be trusted, and $\omega_i$ asks whether a sample is worth updating on now.

\subsubsection{Does the guard select the better mask, and does the challenge reduce high-frequency reliance}
\mbox{}\par\vspace{0\baselineskip}
\noindent\textbf{Guard selection.} We evaluate samples for which $Q_{\mathrm{sup}}$ is surrogate-based, namely semi-supervised unlabeled and point-supervised samples. Selection is scored by post-hoc Top-1 accuracy, the fraction for which $m^{+}$ is within $10^{-3}$ of the highest true IoU among the three candidates, and by regret $\max_r\mathrm{IoU}(m^r,y)-\mathrm{IoU}(m^{+},y)$. 
Table~\ref{tab:guardsel} shows that the full guard halves regret or better relative to the better single-cue rule, substantially exceeds the uniform-random Top-1 baseline, and improves over the guard without $Q_{\mathrm{sup}}$.
Confidence is the weaker single cue, and consistency performs better but can still favor stable errors. Selection is also not trivial: the clean candidate accounts for at most $42\%$ of selections.

\noindent\textbf{Frequency intervention.}
We probe frequency reliance at an unseen cutoff of $0.70$ of the Nyquist radius, outside the training radii $\{0.90,0.95,1.00\}$, using high-frequency removal, low-frequency suppression, and band-limited high-frequency noise, with full protocol details in the Supp. Table~\ref{tab:freqint} reports drops relative to each model's clean score. Relative to standard training, PaceShift reduces sensitivity to high-frequency removal and noise by $43\%$ and $45\%$, while becoming more sensitive to low-frequency suppression, consistent with reduced reliance on high-frequency content. Naive easy-first training is both less accurate and more sensitive to the high-frequency interventions, consistent with the shortcut hypothesis.

\section{Discussion}
\label{sec:discussion}

\subsection{Why the Paradigm Works}
The diagnostics of Sec.~\ref{sec:exp:mech} say more when considered together than separately. Reliability and update value are two different quantities: $\rho_i$ ranks held-out target quality more effectively than the evaluated confidence-based criteria (Tables~\ref{tab:mechscores}(a) and~\ref{tab:triage}), whereas $\omega_i$ is the score most positively associated with the optimizer-conditioned alignment proxy, and instantaneous loss is negatively associated with it (Table~\ref{tab:mechscores}(b)). The guard selects low-regret candidates that confidence alone would not (Table~\ref{tab:guardsel}), both phases reduce high-frequency sensitivity (Table~\ref{tab:freqint}), and the proposed order outperforms reversed and challenge-first schedules (Table~\ref{tab:phaseorder}). Together, these diagnostics support treating training order, supervision reliability, and post-training calibration as connected aspects of the training design.

\subsection{Scope and Limitations}
\looseness=-1
Spectral intervention is most useful when targets and backgrounds share low-level statistics, as in camouflaged, polyp, and transparent segmentation, and this mechanism is most directly supported by the CECS diagnostics. The broader experiments evaluate the full training recipe without isolating the contribution of each phase (Table~\ref{tab:broaddense}). When an object occupies only a few pixels, spectral attenuation may remove useful target evidence along with nuisance cues, and residual failures also include overlapping instances and heavy occlusion (see the Supp.). Reliability-aware scheduling may offer smaller gains when sample reliability varies little, and incomplete-supervision settings that use external teacher or proposal evidence depend on its quality. Finally, Phase~II adds modest training-time computation (Table~\ref{tab:overhead}) but no inference-time parameters or cost.

\section{Conclusion}
\label{sec:conclusion}
PaceShift turns CurriSeg's setting-specific training recipe into a unified supervision-adaptive formulation and replaces its unguarded single-radius post-training with guarded multi-radius preference learning.
Across CECS tasks, incomplete supervision, foundation-model adaptation, and broader dense prediction, it improves accuracy, and targeted evaluations show improved pooled calibration and corruption robustness in the evaluated settings and support the proposed mechanism.

\balance

\bibliographystyle{IEEEtran}
\bibliography{COD}

@IEEEtranBSTCTL{IEEEexample:BSTcontrol,
  CTLdash_repeated_names  = "no",
  CTLuse_forced_etal       = "yes",
  CTLmax_names_forced_etal = "6",   
  CTLnames_show_etal       = "1",   
  CTLuse_alt_spacing       = "yes",
  CTLalt_stretch_factor    = "4"
}

@inproceedings{zhao2024spider,
  title={Spider: a unified framework for context-dependent concept segmentation},
  author={Zhao, Xiaoqi and Pang, Youwei and Ji, Wei and Sheng, Baicheng and Zuo, Jiaming and Zhang, Lihe and Lu, Huchuan},
  booktitle={ICML},
  pages={60906--60926},
  year={2024}
}

@article{he2026hard,
  title={Hard, Yet Reducible: Controlled Forward Transfer for Synthetic Degradation Curation},
  author={He, Chunming and Zhou, Kailai and Zuo, Jiaming and Liu, Hanqi and Xiao, Fengyang and Pang, Youwei and Liu, Xiaofeng and Lin, Weisi and Zhao, Xiaoqi},
  journal={arXiv preprint arXiv:2610.09849},
  year={2026}
}

@inproceedings{bengio2009curriculum,
  title={Curriculum learning},
  author={Bengio, Yoshua and Louradour, J{\'e}r{\^o}me and Collobert, Ronan and Weston, Jason},
  booktitle={ICML},
  pages={41--48},
  year={2009}
}

@article{he2026unfoldcrf,
  title={UnfoldCRF: Structured Mask Refinement with Image-Conditioned Latent Regions},
  author={He, Chunming and Zhang, Rihan and Xu, Lei and Qin, Guanyi and Fang, Chengyu and Tang, Longxiang and Xiao, Fengyang and Farsiu, Sina},
  journal={arXiv preprint arXiv:2609.33996},
  year={2026}
}

@article{fan2021concealed,
  title={Concealed object detection},
  author={Fan, Deng-Ping and Ji, Ge-Peng and Cheng, Ming-Ming and Shao, Ling},
  journal={IEEE Trans. Pattern Anal. Mach. Intell.},
  volume={44},
  number={10},
  pages={6024--6042},
  year={2022},
  publisher={IEEE}
}

@inproceedings{kumar2010self,
  title={Self-paced learning for latent variable models},
  author={Kumar, M. Pawan and Packer, Benjamin and Koller, Daphne},
  booktitle={NeurIPS},
  volume={23},
  year={2010}
}

@article{soviany2020curriculum,
  title={Curriculum learning with diversity for supervised computer vision tasks},
  author={Soviany, Petru},
  journal={arXiv preprint arXiv:2009.10625},
  year={2020}
}

@inproceedings{pentina2015curriculum,
  title={Curriculum learning of multiple tasks},
  author={Pentina, Anastasia and Sharmanska, Viktoriia and Lampert, Christoph H},
  booktitle={CVPR},
  pages={5492--5500},
  year={2015}
}

@inproceedings{saglietti2022analytical,
  title={An analytical theory of curriculum learning in teacher-student networks},
  author={Saglietti, Luca and Mannelli, Stefano and Saxe, Andrew},
  booktitle={NeurIPS},
  volume={35},
  pages={21113--21127},
  year={2022}
}

@article{matiisen2019teacher,
  title={Teacher--student curriculum learning},
  author={Matiisen, Tambet and Oliver, Avital and Cohen, Taco and Schulman, John},
  journal={IEEE Trans. Neural Networks Learn. Syst.},
  volume={31},
  number={9},
  pages={3732--3740},
  year={2020},
  publisher={IEEE}
}

@article{zhang2025referring,
  title={Referring camouflaged object detection},
  author={Zhang, Xuying and Yin, Bowen and Lin, Zheng and Hou, Qibin and Fan, Deng-Ping and Cheng, Ming-Ming},
  journal={IEEE Trans. Pattern Anal. Mach. Intell.},
  volume={47},
  number={5},
  pages={3597--3610},
  year={2025},
  publisher={IEEE}
}

@inproceedings{he2025scaler,
  title={When specialists meet generalists: Segmenter-coordinated asymmetric learning for label-deficient concealed object segmentation},
  author={He, Chunming and Zhang, Dingming and Tang, Longxiang and Yang, Ziyun and Xiao, Fengyang and Farsiu, Sina},
  booktitle={ECCV},
  year={2026}
}

@inproceedings{toneva2018empirical,
  title={An empirical study of example forgetting during deep neural network learning},
  author={Toneva, Mariya and Sordoni, Alessandro and {Tachet des Combes}, Remi and Trischler, Adam and Bengio, Yoshua and Gordon, Geoffrey J},
  booktitle={ICLR},
  year={2019}
}

@inproceedings{paul2021deep,
  title={Deep learning on a data diet: Finding important examples early in training},
  author={Paul, Mansheej and Ganguli, Surya and Dziugaite, Gintare Karolina},
  booktitle={NeurIPS},
  volume={34},
  pages={20596--20607},
  year={2021}
}

@inproceedings{tishby2015deep,
  title={Deep learning and the information bottleneck principle},
  author={Tishby, Naftali and Zaslavsky, Noga},
  booktitle={ITW},
  pages={1--5},
  year={2015}
}

@article{zhi2020frequency,
  title={Frequency principle: {Fourier} analysis sheds light on deep neural networks},
  author={Xu, Zhi-Qin John and Zhang, Yaoyu and Luo, Tao and Xiao, Yanyang and Ma, Zheng},
  journal={Commun. Comput. Phys.},
  volume={28},
  number={5},
  pages={1746--1767},
  year={2020}
}

@inproceedings{rafailov2023direct,
  title={Direct preference optimization: Your language model is secretly a reward model},
  author={Rafailov, Rafael and Sharma, Archit and Mitchell, Eric and Manning, Christopher D and Ermon, Stefano and Finn, Chelsea},
  booktitle={NeurIPS},
  volume={36},
  pages={53728--53741},
  year={2023}
}

@inproceedings{hu2022lora,
  title={{LoRA}: Low-rank adaptation of large language models},
  author={Hu, Edward J and Shen, Yelong and Wallis, Phillip and Allen-Zhu, Zeyuan and Li, Yuanzhi and Wang, Shean and Wang, Lu and Chen, Weizhu},
  booktitle={ICLR},
  year={2022}
}

@inproceedings{swayamdipta2020dataset,
  title={Dataset cartography: Mapping and diagnosing datasets with training dynamics},
  author={Swayamdipta, Swabha and Schwartz, Roy and Lourie, Nicholas and Wang, Yizhong and Hajishirzi, Hannaneh and Smith, Noah A and Choi, Yejin},
  booktitle={EMNLP},
  pages={9275--9293},
  year={2020}
}

@inproceedings{rahaman2019spectral,
  title={On the spectral bias of neural networks},
  author={Rahaman, Nasim and Baratin, Aristide and Arpit, Devansh and Draxler, Felix and Lin, Min and Hamprecht, Fred and Bengio, Yoshua and Courville, Aaron},
  booktitle={ICML},
  pages={5301--5310},
  year={2019}
}

@inproceedings{shah2020pitfalls,
  title={The pitfalls of simplicity bias in neural networks},
  author={Shah, Harshay and Tamuly, Kaustav and Raghunathan, Aditi and Jain, Prateek and Netrapalli, Praneeth},
  booktitle={NeurIPS},
  volume={33},
  pages={9573--9585},
  year={2020}
}

@inproceedings{shu2019meta,
  title={{Meta-Weight-Net}: Learning an explicit mapping for sample weighting},
  author={Shu, Jun and Xie, Qi and Yi, Lixuan and Zhao, Qian and Zhou, Sanping and Xu, Zongben and Meng, Deyu},
  booktitle={NeurIPS},
  volume={32},
  year={2019}
}

@inproceedings{han2018co,
  title={Co-teaching: Robust training of deep neural networks with extremely noisy labels},
  author={Han, Bo and Yao, Quanming and Yu, Xingrui and Niu, Gang and Xu, Miao and Hu, Weihua and Tsang, Ivor and Sugiyama, Masashi},
  booktitle={NeurIPS},
  volume={31},
  year={2018}
}

@article{soviany2022curriculum,
  title={Curriculum learning: A survey},
  author={Soviany, Petru and Ionescu, Radu Tudor and Rota, Paolo and Sebe, Nicu},
  journal={Int. J. Comput. Vis.},
  number={6},
  pages={1526--1565},
  volume={130},
  year={2022}
}

@inproceedings{chen2023sam,
  title={{SAM-Adapter}: Adapting segment anything in underperformed scenes},
  author={Chen, Tianrun and Zhu, Lanyun and Deng, Chaotao and Cao, Runlong and Wang, Yan and Zhang, Shangzhan and Li, Zejian and Sun, Lingyun and Zang, Ying and Mao, Papa},
  booktitle={ICCVW},
  pages={3367--3375},
  year={2023}
}

@inproceedings{chakraborty2024unsupervised,
  title={Unsupervised and semi-supervised co-salient object detection via segmentation frequency statistics},
  author={Chakraborty, Souradeep and Naha, Shujon and Bastan, Muhammet and {K. C.}, Amit Kumar and Samaras, Dimitris},
  booktitle={WACV},
  pages={332--342},
  year={2024}
}

@inproceedings{ni2024context,
  title={Context-guided spatial feature reconstruction for efficient semantic segmentation},
  author={Ni, Zhenliang and Chen, Xinghao and Zhai, Yingjie and Tang, Yehui and Wang, Yunhe},
  booktitle={ECCV},
  pages={239--255},
  year={2024},
}

@article{chen2024sam2,
  title={{SAM2-Adapter}: Evaluating \& adapting segment anything 2 in downstream tasks: Camouflage, shadow, medical image segmentation, and more},
  author={Chen, Tianrun and Lu, Ankang and Zhu, Lanyun and Ding, Chaotao and Yu, Chunan and Ji, Deyi and Li, Zejian and Sun, Lingyun and Mao, Papa and Zang, Ying},
  journal={arXiv preprint arXiv:2408.04579},
  year={2024}
}

@article{chen2025sam3,
  title={{SAM3-Adapter}: Efficient Adaptation of Segment Anything 3 for Camouflage Object Segmentation, Shadow Detection, and Medical Image Segmentation},
  author={Chen, Tianrun and Cao, Runlong and Yu, Xinda and Zhu, Lanyun and Ding, Chaotao and Ji, Deyi and Chen, Cheng and Zhu, Qi and Xu, Chunyan and Mao, Papa and Zang, Ying},
  journal={arXiv preprint arXiv:2511.19425},
  year={2025}
}

@inproceedings{vicente2016large,
  title={Large-scale training of shadow detectors with noisily-annotated shadow examples},
  author={Vicente, Tom{\'a}s F Yago and Hou, Le and Yu, Chen-Ping and Hoai, Minh and Samaras, Dimitris},
  booktitle={ECCV},
  pages={816--832},
  year={2016}
}

@inproceedings{zhang2024irsam,
  title={{IRSAM}: Advancing segment anything model for infrared small target detection},
  author={Zhang, Mingjin and Wang, Yuchun and Guo, Jie and Li, Yunsong and Gao, Xinbo and Zhang, Jing},
  booktitle={ECCV},
  pages={233--249},
  year={2024},
}

@inproceedings{zhang2022isnet,
  title={{ISNet}: Shape matters for infrared small target detection},
  author={Zhang, Mingjin and Zhang, Rui and Yang, Yuxiang and Bai, Haichen and Zhang, Jing and Guo, Jie},
  booktitle={CVPR},
  pages={877--886},
  year={2022}
}

@inproceedings{cheng2022masked,
  title={Masked-attention mask transformer for universal image segmentation},
  author={Cheng, Bowen and Misra, Ishan and Schwing, Alexander G and Kirillov, Alexander and Girdhar, Rohit},
  booktitle={CVPR},
  pages={1290--1299},
  year={2022}
}

@inproceedings{lin2014microsoft,
  title={Microsoft {COCO}: Common objects in context},
  author={Lin, Tsung-Yi and Maire, Michael and Belongie, Serge and Hays, James and Perona, Pietro and Ramanan, Deva and Doll{\'a}r, Piotr and Zitnick, C Lawrence},
  booktitle={ECCV},
  pages={740--755},
  year={2014}
}

@inproceedings{zhou2017scene,
  title={Scene parsing through {ADE20K} dataset},
  author={Zhou, Bolei and Zhao, Hang and Puig, Xavier and Fidler, Sanja and Barriuso, Adela and Torralba, Antonio},
  booktitle={CVPR},
  pages={633--641},
  year={2017}
}

@inproceedings{zhai2021mutual,
  title={Mutual graph learning for camouflaged object detection},
  author={Zhai, Qiang and Li, Xin and Yang, Fan and Chen, Chenglizhao and Cheng, Hong and Fan, Deng-Ping},
  booktitle={CVPR},
  pages={2997--3007},
  year={2021}
}

@inproceedings{he2023weaklysupervised,
  title={Weakly-supervised concealed object segmentation with {SAM}-based pseudo labeling and multi-scale feature grouping},
  author={He, Chunming and Li, Kai and Zhang, Yachao and Xu, Guoxia and Tang, Longxiang and Zhang, Yulun and Guo, Zhenhua and Li, Xiu},
  booktitle={NeurIPS},
  volume={36},
  pages={30726--30737},
  year={2023}
}

@inproceedings{fan2020camouflaged,
  title={Camouflaged object detection},
  author={Fan, Deng-Ping and Ji, Ge-Peng and Sun, Guolei and Cheng, Ming-Ming and Shen, Jianbing and Shao, Ling},
  booktitle={CVPR},
  pages={2777--2787},
  year={2020}
}

@inproceedings{yang2023oaformer,
  title={{OAFormer}: Occlusion aware transformer for camouflaged object detection},
  author={Yang, Xin and Zhu, Hengliang and Mao, Guojun and Xing, Shuli},
  booktitle={ICME},
  pages={1421--1426},
  year={2023},
}

@article{fan2023advances,
  title={Advances in deep concealed scene understanding},
  author={Fan, Deng-Ping and Ji, Ge-Peng and Xu, Peng and Cheng, Ming-Ming and Sakaridis, Christos and Van Gool, Luc},
  journal={Visual Intell.},
  volume={1},
  number={1},
  note={Art. no. 16},
  year={2023},
  publisher={Springer}
}

@inproceedings{lv2021simultaneously,
  title={Simultaneously localize, segment and rank the camouflaged objects},
  author={Lv, Yunqiu and Zhang, Jing and Dai, Yuchao and Li, Aixuan and Liu, Bowen and Barnes, Nick and Fan, Deng-Ping},
  booktitle={CVPR},
  pages={11591--11601},
  year={2021}
}

@inproceedings{yang2021uncertainty,
  title={Uncertainty-guided transformer reasoning for camouflaged object detection},
  author={Yang, Fan and Zhai, Qiang and Li, Xin and Huang, Rui and Luo, Ao and Cheng, Hong and Fan, Deng-Ping},
  booktitle={ICCV},
  pages={4146--4155},
  year={2021}
}

@inproceedings{zhang2022preynet,
  title={{PreyNet}: Preying on camouflaged objects},
  author={Zhang, Miao and Xu, Shuang and Piao, Yongri and Shi, Dongxiang and Lin, Shusen and Lu, Huchuan},
  booktitle={ACM MM},
  pages={5323--5332},
  year={2022}
}

@inproceedings{He2023Camouflaged,
title={Camouflaged object detection with feature decomposition and edge reconstruction},
author={He, Chunming and Li, Kai and Zhang, Yachao and Tang, Longxiang and Zhang, Yulun and Guo, Zhenhua and Li, Xiu},
booktitle={CVPR},
pages={22046--22055},
year={2023}
}

@article{he2026ride,
  title={{RIDE}: {Retinex}-Informed Decoupling for Exposing Concealed Objects},
  author={He, Chunming and Zhang, Rihan and Zhang, Dingming and Fang, Chengyu and Tang, Longxiang and Feng, Jingjia and Xiao, Fengyang and Farsiu, Sina},
  journal={arXiv preprint arXiv:2605.15450},
  year={2026}
}

@inproceedings{he2023strategic,
 title={Strategic Preys Make Acute Predators: Enhancing Camouflaged Object Detectors by Generating Camouflaged Objects},
 author={He, Chunming and Li, Kai and Zhang, Yachao and Zhang, Yulun and You, Chenyu and Guo, Zhenhua and Li, Xiu and Danelljan, Martin and Yu, Fisher},
 booktitle={ICLR},
 year={2024}
}

@inproceedings{xiao2026beyond,
  title={Beyond Ground-Truth: Leveraging Image Quality Priors for Real-World Image Restoration},
  author={Xiao, Fengyang and Hu, Peng and Xu, Lei and Guo, XingE and Qin, Guanyi and Shen, Yuqi and Fang, Chengyu and Zhang, Rihan and He, Chunming and Farsiu, Sina},
  booktitle={CVPR},
  pages={29897--29908},
  year={2026}
}

@inproceedings{sun2024frequency,
  title={Frequency-spatial entanglement learning for camouflaged object detection},
  author={Sun, Yanguang and Xu, Chunyan and Yang, Jian and Xuan, Hanyu and Luo, Lei},
  booktitle={ECCV},
  pages={343--360},
  year={2024}
}

@inproceedings{wang2024lssnet,
  title={{LSSNet}: A Method for Colon Polyp Segmentation Based on Local Feature Supplementation and Shallow Feature Supplementation},
  author={Wang, Wei and Sun, Huiying and Wang, Xin},
  booktitle={MICCAI},
  pages={446--456},
  year={2024}
}

@article{dong2023polyp,
  title={{Polyp-PVT}: Polyp Segmentation with {Pyramid Vision Transformers}},
  author={Dong, Bo and Wang, Wenhai and Fan, Deng-Ping and Li, Jinpeng and Fu, Huazhu and Shao, Ling},
  journal={CAAI Artif. Intell. Res.},
  volume={2},
  note={Art. no. 9150015},
  year={2023}
}

@article{jain2023coinnet,
  title={{CoInNet}: A convolution-involution network with a novel statistical attention for automatic polyp segmentation},
  author={Jain, Samir and Atale, Rohan and Gupta, Anubhav and Mishra, Utkarsh and Seal, Ayan and Ojha, Aparajita and Jaworek-Korjakowska, Joanna and Krejcar, Ondrej},
  journal={IEEE Trans. Med. Imaging},
  volume={42},
  number={12},
  pages={3987--4000},
  year={2023},
  publisher={IEEE}
}

@inproceedings{he2021enhanced,
  title={Enhanced boundary learning for glass-like object segmentation},
  author={He, Hao and Li, Xiangtai and Cheng, Guangliang and Shi, Jianping and Tong, Yunhai and Meng, Gaofeng and Prinet, V{\'e}ronique and Weng, LuBin},
  booktitle={ICCV},
  pages={15859--15868},
  year={2021}
}

@article{liu2024vst++,
  title={{VST++}: Efficient and stronger visual saliency transformer},
  author={Liu, Nian and Luo, Ziyang and Zhang, Ni and Han, Junwei},
  journal={IEEE Trans. Pattern Anal. Mach. Intell.},
  volume={46},
  number={11},
  pages={7300--7316},
  year={2024},
  publisher={IEEE}
}

@inproceedings{fan2023rfenet,
  title={{RFENet}: towards reciprocal feature evolution for glass segmentation},
  author={Fan, Ke and Wang, Changan and Wang, Yabiao and Wang, Chengjie and Yi, Ran and Ma, Lizhuang},
  booktitle={IJCAI},
  pages={717--725},
  year={2023}
}

@inproceedings{luo2024vscode,
  title={{VSCode}: General Visual Salient and Camouflaged Object Detection with {2D} Prompt Learning},
  author={Luo, Ziyang and Liu, Nian and Zhao, Wangbo and Yang, Xuguang and Zhang, Dingwen and Fan, Deng-Ping and Khan, Fahad and Han, Junwei},
  booktitle={CVPR},
  pages={17169--17180},
  year={2024}
}

@article{pang2024zoomnext,
  title={{ZoomNeXt}: A unified collaborative pyramid network for camouflaged object detection},
  author={Pang, Youwei and Zhao, Xiaoqi and Xiang, Tian-Zhu and Zhang, Lihe and Lu, Huchuan},
  journal={IEEE Trans. Pattern Anal. Mach. Intell.},
  volume={46},
  number={12},
  pages={9205--9220},
  year={2024},
  publisher={IEEE}
}

@article{yin2024camoformer,
  title={{CamoFormer}: Masked separable attention for camouflaged object detection},
  author={Yin, Bowen and Zhang, Xuying and Fan, Deng-Ping and Jiao, Shaohui and Cheng, Ming-Ming and Van Gool, Luc and Hou, Qibin},
  journal={IEEE Trans. Pattern Anal. Mach. Intell.},
  volume={46},
  number={12},
  pages={10362--10374},
  year={2024},
  publisher={IEEE}
}

@inproceedings{kirillov2023segment,
  title={Segment anything},
  author={Kirillov, Alexander and Mintun, Eric and Ravi, Nikhila and Mao, Hanzi and Rolland, Chloe and Gustafson, Laura and Xiao, Tete and Whitehead, Spencer and Berg, Alexander C and Lo, Wan-Yen and others},
  booktitle={ICCV},
  pages={4015--4026},
  year={2023}
}

@article{jan2025c3net,
  title={{C3Net}: Context-Contrast Network for Camouflaged Object Detection},
  author={Jan, Baber and El-Maleh, Aiman H and Siddiqui, Abdul Jabbar and Bais, Abdul and Anwar, Saeed},
  journal={arXiv preprint arXiv:2511.12627},
  year={2025}
}

@article{bogunovic2019retouch,
  title={{RETOUCH}: The retinal {OCT} fluid detection and segmentation benchmark and challenge},
  author={Bogunovi{\'c}, Hrvoje and Venhuizen, Freerk and Klimscha, Sophie and Apostolopoulos, Stefanos and Bab-Hadiashar, Alireza and Bagci, Ulas and Beg, Mirza Faisal and Bekalo, Loza and Chen, Qiang and Ciller, Carlos and others},
  journal={IEEE Trans. Med. Imaging},
  volume={38},
  number={8},
  pages={1858--1874},
  year={2019},
  publisher={IEEE}
}

@inproceedings{yang2023directional,
  title={Directional connectivity-based segmentation of medical images},
  author={Yang, Ziyun and Farsiu, Sina},
  booktitle={CVPR},
  pages={11525--11535},
  year={2023}
}

@article{codella2019skin,
  title={Skin lesion analysis toward melanoma detection 2018: A challenge hosted by the {International Skin Imaging Collaboration} ({ISIC})},
  author={Codella, Noel and Rotemberg, Veronica and Tschandl, Philipp and Celebi, M Emre and Dusza, Stephen and Gutman, David and Helba, Brian and Kalloo, Aadi and Liopyris, Konstantinos and Marchetti, Michael and others},
  journal={arXiv preprint arXiv:1902.03368},
  year={2019}
}

@inproceedings{yu2024multi,
  title={Multi-view aggregation network for dichotomous image segmentation},
  author={Yu, Qian and Zhao, Xiaoqi and Pang, Youwei and Zhang, Lihe and Lu, Huchuan},
  booktitle={CVPR},
  pages={3921--3930},
  year={2024}
}

@inproceedings{qin2022highly,
  title={Highly accurate dichotomous image segmentation},
  author={Qin, Xuebin and Dai, Hang and Hu, Xiaobin and Fan, Deng-Ping and Shao, Ling and Van Gool, Luc},
  booktitle={ECCV},
  pages={38--56},
  year={2022}
}

@inproceedings{he2026nested,
  title={{NUN}: Nested unfolding network for real-world concealed object segmentation},
  author={He, Chunming and Zhang, Rihan and Tang, Longxiang and Zhang, Dingming and Zhang, Bojian and Xiao, Fengyang and Feng, Jingjia and Farsiu, Sina},
  booktitle={ECCV},
  year={2026}
}

@inproceedings{fan2018enhanced,
  title={Enhanced-alignment measure for binary foreground map evaluation},
  author={Fan, Deng-Ping and Gong, Cheng and Cao, Yang and Ren, Bo and Cheng, Ming-Ming and Borji, Ali},
  booktitle={IJCAI},
  pages={698--704},
  year={2018}
}

@inproceedings{achanta2009frequency,
  title={Frequency-tuned salient region detection},
  author={Achanta, Radhakrishna and Hemami, Sheila and Estrada, Francisco and S{\"u}sstrunk, Sabine},
  booktitle={CVPR},
  pages={1597--1604},
  year={2009}
}

@article{zhang2024evf,
  title={{EVF-SAM}: Early vision-language fusion for text-prompted segment anything model},
  author={Zhang, Yuxuan and Cheng, Tianheng and Zhu, Lianghui and Liu, Lei and Liu, Heng and Ran, Longjin and Chen, Xiaoxin and Liu, Wenyu and Wang, Xinggang},
  journal={Image Vis. Comput.},
  volume={173},
  note={Art. no. 106028},
  year={2026}
}

@inproceedings{cho2024cat,
  title={{CAT-Seg}: Cost aggregation for open-vocabulary semantic segmentation},
  author={Cho, Seokju and Shin, Heeseong and Hong, Sunghwan and Arnab, Anurag and Seo, Paul Hongsuck and Kim, Seungryong},
  booktitle={CVPR},
  pages={4113--4123},
  year={2024}
}

@inproceedings{luddecke2022image,
  title={Image segmentation using text and image prompts},
  author={L{\"u}ddecke, Timo and Ecker, Alexander},
  booktitle={CVPR},
  pages={7086--7096},
  year={2022}
}

@inproceedings{hermann2020origins,
  title={The origins and prevalence of texture bias in convolutional neural networks},
  author={Hermann, Katherine L. and Chen, Ting and Kornblith, Simon},
  booktitle={NeurIPS},
  volume={33},
  pages={19000--19015},
  year={2020}
}

@inproceedings{pezeshki2021gradient,
  title={Gradient starvation: A learning proclivity in neural networks},
  author={Pezeshki, Mohammad and Kaba, Oumar and Bengio, Yoshua and Courville, Aaron C and Precup, Doina and Lajoie, Guillaume},
  booktitle={NeurIPS},
  volume={34},
  pages={1256--1272},
  year={2021}
}

@inproceedings{arazo2020pseudo,
  title={Pseudo-labeling and confirmation bias in deep semi-supervised learning},
  author={Arazo, Eric and Ortego, Diego and Albert, Paul and O’Connor, Noel E and McGuinness, Kevin},
  booktitle={IJCNN},
  pages={1--8},
  year={2020}
}

@article{le2019anabranch,
  title={Anabranch network for camouflaged object segmentation},
  author={Le, Trung-Nghia and Nguyen, Tam V and Nie, Zhongliang and Tran, Minh-Triet and Sugimoto, Akihiro},
  journal={Comput. Vis. Image Underst.},
  volume={184},
  pages={45--56},
  year={2019},
  publisher={Elsevier}
}

@misc{skurowski2018animal,
  title={Animal camouflage analysis: {CHAMELEON} database},
  author={Skurowski, Przemys{\l}aw and Abdulameer, Hassan and B{\l}aszczyk, Jakub and Depta, Tomasz and Kornacki, Adam and Kozie{\l}, Przemys{\l}aw},
  howpublished={Unpublished manuscript},
  year={2018}
}

@article{bernal2012towards,
  title={Towards automatic polyp detection with a polyp appearance model},
  author={Bernal, Jorge and S{\'a}nchez, Javier and Vilari{\~n}o, Fernando},
  journal={Pattern Recognit.},
  volume={45},
  number={9},
  pages={3166--3182},
  year={2012},
  publisher={Elsevier}
}

@article{bernal2015wm,
  title={{WM-DOVA} maps for accurate polyp highlighting in colonoscopy: Validation vs. saliency maps from physicians},
  author={Bernal, Jorge and S{\'a}nchez, F Javier and Fern{\'a}ndez-Esparrach, Gloria and Gil, Debora and Rodr{\'\i}guez, Cristina and Vilari{\~n}o, Fernando},
  journal={Comput. Med. Imaging Graph.},
  volume={43},
  pages={99--111},
  year={2015}
}

@article{silva2014toward,
  title={Toward embedded detection of polyps in {WCE} images for early diagnosis of colorectal cancer},
  author={Silva, Juan and Histace, Aymeric and Romain, Olivier and Dray, Xavier and Granado, Bertrand},
  journal={Int. J. Comput. Assist. Radiol. Surg.},
  volume={9},
  pages={283--293},
  number={2},
  year={2014}
}

@article{xiao2024survey,
  title={A Survey of Camouflaged Object Detection and Beyond},
  author={Xiao, Fengyang and Hu, Sujie and Shen, Yuqi and Fang, Chengyu and Huang, Jinfa and Tang, Longxiang and Yang, Ziyun and Li, Xiu and He, Chunming},
  journal={CAAI Artif. Intell. Res.},
  volume={3},
  note={Art. no. 9150044},
  year={2024}
}

@inproceedings{jha2020kvasir,
  title={{Kvasir-SEG}: A segmented polyp dataset},
  author={Jha, Debesh and Smedsrud, Pia H and Riegler, Michael A and Halvorsen, P{\aa}l and de Lange, Thomas and Johansen, Dag and Johansen, H{\aa}vard D.},
  booktitle={MMM},
  pages={451--462},
  year={2020}
}

@inproceedings{mei2020don,
  title={Don't hit me! glass detection in real-world scenes},
  author={Mei, Haiyang and Yang, Xin and Wang, Yang and Liu, Yuanyuan and He, Shengfeng},
  booktitle={CVPR},
  pages={3687--3696},
  year={2020}
}

@inproceedings{lin2021rich,
  title={Rich context aggregation with reflection prior for glass surface detection},
  author={Lin, Jiaying and He, Zebang},
  booktitle={CVPR},
  pages={13415--13424},
  year={2021}
}

@article{yan2024ghostingnet,
  title={{GhostingNet}: A Novel Approach for Glass Surface Detection With Ghosting Cues},
  author={Yan, Tao and Gao, Jiahui and Xu, Ke and Zhu, Xiangjie and Huang, Hao and Li, Helong and Wah, Benjamin and Lau, Rynson W. H.},
  journal={IEEE Trans. Pattern Anal. Mach. Intell.},
  volume={47},
  number={1},
  pages={323--337},
  year={2025},
  publisher={IEEE}
}

@article{wang2022pvt,
  title={{PVT v2}: Improved baselines with {Pyramid Vision Transformer}},
  author={Wang, Wenhai and Xie, Enze and Li, Xiang and Fan, Deng-Ping and Song, Kaitao and Liang, Ding and Lu, Tong and Luo, Ping and Shao, Ling},
  journal={Comput. Vis. Media},
  volume={8},
  number={3},
  pages={415--424},
  year={2022},
  publisher={Springer}
}

@article{han2023internal,
  title={Internal-external boundary attention fusion for glass surface segmentation},
  author={Han, Dongshen and Yoon, Heechan and Kwon, Hyukmin and Kim, Hyun-Cheol and Choo, Hyon-Gon and Lee, Seungkyu and Zhang, Chaoning},
  journal={Neural Netw.},
  volume={195},
  note={Art. no. 108232},
  year={2026}
}

@inproceedings{guo2017calibration,
  title={On calibration of modern neural networks},
  author={Guo, Chuan and Pleiss, Geoff and Sun, Yu and Weinberger, Kilian Q},
  booktitle={ICML},
  pages={1321--1330},
  year={2017}
}

@article{everingham2010pascal,
  title={The {PASCAL} visual object classes ({VOC}) challenge},
  author={Everingham, Mark and Van Gool, Luc and Williams, Christopher K. I. and Winn, John and Zisserman, Andrew},
  journal={Int. J. Comput. Vis.},
  volume={88},
  pages={303--338},
  number={2},
  year={2010},
  publisher={Springer}
}

@inproceedings{bui2024meganet,
  title={{MEGANet}: Multi-scale edge-guided attention network for weak boundary polyp segmentation},
  author={Bui, Nhat-Tan and Hoang, Dinh-Hieu and Nguyen, Quang-Thuc and Tran, Minh-Triet and Le, Ngan},
  booktitle={WACV},
  pages={7985--7994},
  year={2024}
}

@article{he2025segment,
  title={Segment concealed objects with incomplete supervision},
  author={He, Chunming and Li, Kai and Zhang, Yachao and Yang, Ziyun and Pang, Youwei and Tang, Longxiang and Fang, Chengyu and Zhang, Yulun and Kong, Linghe and Li, Xiu and Farsiu, Sina},
  journal={IEEE Trans. Pattern Anal. Mach. Intell.},
  volume={47},
  number={9},
  pages={7832--7851},
  year={2025}
}

@inproceedings{wang2017learning,
  title={Learning to detect salient objects with image-level supervision},
  author={Wang, Lijun and Lu, Huchuan and Wang, Yifan and Feng, Mengyang and Wang, Dong and Yin, Baocai and Ruan, Xiang},
  booktitle={CVPR},
  pages={136--145},
  year={2017}
}

@inproceedings{pleiss2020identifying,
  title={Identifying mislabeled data using the area under the margin ranking},
  author={Pleiss, Geoff and Zhang, Tianyi and Elenberg, Ethan and Weinberger, Kilian Q},
  booktitle={NeurIPS},
  volume={33},
  pages={17044--17056},
  year={2020}
}

@article{wu2025refonce,
  title={{RefOnce}: Distilling References into a Prototype Memory for Referring Camouflaged Object Detection},
  author={Wu, Yu-Huan and Zhu, Zi-Xuan and Wang, Yan and Zhen, Liangli and Fan, Deng-Ping},
  journal={arXiv preprint arXiv:2511.20989},
  year={2025}
}

@article{zhao2026open,
  title={Open-vocabulary camouflaged object segmentation with cascaded vision language models},
  author={Zhao, Kai and Yuan, Wubang and Wang, Zheng and Li, Guanyi and Zhu, Xiaoqiang and Fan, Deng-Ping and Zeng, Dan},
  journal={Comput. Vis. Media},
  volume={12},
  number={2},
  pages={473--492},
  year={2026}
}

@inproceedings{pang2024open,
  title={Open-vocabulary camouflaged object segmentation},
  author={Pang, Youwei and Zhao, Xiaoqi and Zuo, Jiaming and Zhang, Lihe and Lu, Huchuan},
  booktitle={ECCV},
  pages={476--495},
  year={2024}
}

@inproceedings{yu2023convolutions,
  title={Convolutions die hard: Open-vocabulary segmentation with single frozen convolutional {CLIP}},
  author={Yu, Qihang and He, Ju and Deng, Xueqing and Shen, Xiaohui and Chen, Liang-Chieh},
  booktitle={NeurIPS},
  volume={36},
  pages={32215--32234},
  year={2023}
}

@inproceedings{jang2024dial,
  title={{DIAL}: Dense image-text alignment for weakly supervised semantic segmentation},
  author={Jang, Soojin and Yun, Jungmin and Kwon, Junehyoung and Lee, Eunju and Kim, Youngbin},
  booktitle={ECCV},
  pages={248--266},
  year={2024}
}

@inproceedings{yang2025exploring,
  title={Exploring {CLIP}'s Dense Knowledge for Weakly Supervised Semantic Segmentation},
  author={Yang, Zhiwei and Meng, Yucong and Fu, Kexue and Tang, Feilong and Wang, Shuo and Song, Zhijian},
  booktitle={CVPR},
  pages={20223--20232},
  year={2025}
}

@inproceedings{zhang2024frozen,
  title={Frozen {CLIP}: A strong backbone for weakly supervised semantic segmentation},
  author={Zhang, Bingfeng and Yu, Siyue and Wei, Yunchao and Zhao, Yao and Xiao, Jimin},
  booktitle={CVPR},
  pages={3796--3806},
  year={2024}
}

@inproceedings{zheng2025elfs,
  title={{ELFS}: Label-free coreset selection with proxy training dynamics},
  author={Zheng, Haizhong and Tsai, Elisa and Lu, Yifu and Sun, Jiachen and Bartoldson, Brian R. and Kailkhura, Bhavya and Prakash, Atul},
  booktitle={ICLR},
  year={2025}
}

@inproceedings{zheng2022coverage,
  title={Coverage-centric coreset selection for high pruning rates},
  author={Zheng, Haizhong and Liu, Rui and Lai, Fan and Prakash, Atul},
  booktitle={ICLR},
  year={2023}
}

@inproceedings{xia2022moderate,
  title={Moderate coreset: A universal method of data selection for real-world data-efficient deep learning},
  author={Xia, Xiaobo and Liu, Jiale and Yu, Jun and Shen, Xu and Han, Bo and Liu, Tongliang},
  booktitle={ICLR},
  year={2023}
}

@inproceedings{liu2022adaptive,
  title={Adaptive early-learning correction for segmentation from noisy annotations},
  author={Liu, Sheng and Liu, Kangning and Zhu, Weicheng and Shen, Yiqiu and Fernandez-Granda, Carlos},
  booktitle={CVPR},
  pages={2606--2616},
  year={2022}
}

@inproceedings{yuan2025instance,
  title={Instance-dependent early stopping},
  author={Yuan, Suqin and Lin, Runqi and Feng, Lei and Han, Bo and Liu, Tongliang},
  booktitle={ICLR},
  year={2025}
}

@inproceedings{zhou2025scale,
  title={Scale efficient training for large datasets},
  author={Zhou, Qing and Gao, Junyu and Wang, Qi},
  booktitle={CVPR},
  pages={20458--20467},
  year={2025}
}

@inproceedings{qin2024infobatch,
  title={{InfoBatch}: Lossless training speed up by unbiased dynamic data pruning},
  author={Qin, Ziheng and Wang, Kai and Zheng, Zangwei and Gu, Jianyang and Peng, Xiangyu and Xu, Zhaopan and Zhou, Daquan and Shang, Lei and Sun, Baigui and Xie, Xuansong and You, Yang},
  booktitle={ICLR},
  year={2024}
}

@inproceedings{zhang2024spanning,
  title={Spanning training progress: Temporal dual-depth scoring ({TDDS}) for enhanced dataset pruning},
  author={Zhang, Xin and Du, Jiawei and Li, Yunsong and Xie, Weiying and Zhou, Joey Tianyi},
  booktitle={CVPR},
  pages={26223--26232},
  year={2024}
}

@inproceedings{he2024large,
  title={Large-scale dataset pruning with dynamic uncertainty},
  author={He, Muyang and Yang, Shuo and Huang, Tiejun and Zhao, Bo},
  booktitle={CVPRW},
  pages={7713--7722},
  year={2024}
}

@article{carion2025sam3segmentconcepts,
      title={{SAM 3}: Segment Anything with Concepts}, 
      author={Carion, Nicolas and Gustafson, Laura and Hu, Yuan-Ting and Debnath, Shoubhik and Hu, Ronghang and others},
      journal={arXiv preprint arXiv:2511.16719},
      year={2025},
      eprint={2511.16719},
      archivePrefix={arXiv},
      primaryClass={cs.CV},
}

@article{sun2025conditional,
  title={Conditional diffusion models for camouflaged and salient object detection},
  author={Sun, Ke and Chen, Zhongxi and Lin, Xianming and Sun, Xiaoshuai and Liu, Hong and Ji, Rongrong},
  journal={IEEE Trans. Pattern Anal. Mach. Intell.},
  volume={47},
  number={4},
  pages={2833--2848},
  year={2025},
  publisher={IEEE}
}

@inproceedings{zhu2022can,
  title={I can find you! Boundary-guided separated attention network for camouflaged object detection},
  author={Zhu, Hongwei and Li, Peng and Xie, Haoran and Yan, Xuefeng and Liang, Dong and Chen, Dapeng and Wei, Mingqiang and Qin, Jing},
  booktitle={AAAI},
  volume={36},
  pages={3608--3616},
  year={2022}
}

@inproceedings{fan2017structure,
  title={Structure-measure: A new way to evaluate foreground maps},
  author={Fan, Deng-Ping and Cheng, Ming-Ming and Liu, Yun and Li, Tao and Borji, Ali},
  booktitle={ICCV},
  pages={4548--4557},
  year={2017}
}

@inproceedings{hu2022high,
  title={High-resolution iterative feedback network for camouflaged object detection},
  author={Hu, Xiaobin and Wang, Shuo and Qin, Xuebin and Dai, Hang and Ren, Wenqi and Luo, Donghao and Tai, Ying and Shao, Ling},
  booktitle={AAAI},
  volume={37},
  number={1},
  pages={881--889},
  year={2023}
}

@inproceedings{he2025run,
  title={{RUN}: Reversible Unfolding Network for Concealed Object Segmentation},
  author={He, Chunming and Zhang, Rihan and Xiao, Fengyang and Fang, Chengyu and Tang, Longxiang and Zhang, Yulun and Kong, Linghe and Fan, Deng-Ping and Li, Kai and Farsiu, Sina},
  booktitle={ICML},
  pages={22853--22864},
  year={2025}
}

@inproceedings{sohn2020fixmatch,
  title={{FixMatch}: Simplifying semi-supervised learning with consistency and confidence},
  author={Sohn, Kihyuk and Berthelot, David and Carlini, Nicholas and Zhang, Zizhao and Zhang, Han and Raffel, Colin A and Cubuk, Ekin Dogus and Kurakin, Alexey and Li, Chun-Liang},
  booktitle={NeurIPS},
  volume={33},
  pages={596--608},
  year={2020}
}

@inproceedings{tarvainen2017mean,
  title={Mean teachers are better role models: Weight-averaged consistency targets improve semi-supervised deep learning results},
  author={Tarvainen, Antti and Valpola, Harri},
  booktitle={NeurIPS},
  volume={30},
  year={2017}
}

@inproceedings{he2026refining,
  title={Refining Context-Entangled Content Segmentation via Curriculum Selection and Anti-Curriculum Promotion},
  author={He, Chunming and Zhang, Rihan and Xiao, Fengyang and Zhang, Dingming and Cao, Zhiwen and Farsiu, Sina},
  booktitle={ICML},
  pages={42267--42279},
  year={2026}
}

{\fontsize{8pt}{\baselineskip}\selectfont \noindent\textbf{Chunming He} received the B.S. degree from Nanjing University of Posts and Telecommunications, China, and the M.E. degree from Tsinghua University, China. He is currently a Ph.D. student at Duke University, Durham, USA. His research interests include computer vision and biomedical image analysis.}

\vspace{0.6\baselineskip}

{\fontsize{8pt}{10pt}\selectfont
\noindent\textbf{Dingming Zhang} received the B.S. degree from Zhejiang Sci-Tech University, China. He is currently pursuing the M.S. degree at Duke University, Durham, USA. His research interests include computer vision.}

\vspace{0.6\baselineskip}

{\fontsize{8pt}{10pt}\selectfont
\noindent\textbf{Fengyang Xiao} 
received the B.S. degree from Nanjing University of Posts and Telecommunications, China, and the M.S. degree from Sun Yat-sen University, China. She is currently a Ph.D. student at Duke University, Durham, USA. Her research interests include differential equations and computer vision.}

\vspace{0.6\baselineskip}

{\fontsize{8pt}{10pt}\selectfont
\noindent\textbf{Zhiwen Cao} received the Ph.D. degree from Purdue University, USA. He is currently a Gen AI Researcher at Adobe. His research interests include image, video and audio generation and editing.}

\vspace{0.6\baselineskip}

{\fontsize{8pt}{10pt}\selectfont
\noindent\textbf{Sina Farsiu} (Fellow, IEEE) received his Ph.D. degree in electrical engineering from the University of California, Santa Cruz in 2005. He is now the Anderson-Rupp Professor of BME, with secondary appointments at the Departments of ECE and CS at Duke University. He has served as the Senior Area Editor of IEEE Transactions on Image Processing, Deputy Editor of the Biomedical Optics Express, and the Associate Editor of SIAM Journal on Imaging Sciences. He received the “Outstanding Member of the Editorial Board Award” from the IEEE Signal Processing Society in 2018. He is a Fellow of IEEE, Optica, SPIE, ARVO, and AIMBE.}

\end{document}